\documentclass[english]{article}

\usepackage{algorithmic}
\usepackage{algorithm}
\usepackage{amsfonts, amsmath, amssymb, amsthm}
\usepackage{array}
\usepackage{arydshln}
\usepackage{bm}
\usepackage{color}
\usepackage{dsfont}
\usepackage{enumitem}
\usepackage{float}
\usepackage[T1]{fontenc}
\usepackage{geometry}
\usepackage{graphicx,multirow}
\usepackage{grffile}
\usepackage{hyperref}\hypersetup{colorlinks,linkcolor=red,anchorcolor=blue,citecolor=blue}
\usepackage[utf8]{inputenc}
\usepackage{letltxmacro}
\usepackage{mathrsfs}
\usepackage{mathtools,natbib}
\usepackage{multirow,xspace,booktabs}
\usepackage{nicefrac}
\usepackage{tablefootnote}
\usepackage{tikz}
\usepackage{verbatim}
\usepackage{subcaption}

\newcommand{\R}{\mathbb R}
\newcommand{\E}{\mathbb E}

\DeclareMathOperator{\supp}{supp}

\newcommand{\reals}{\mathbb{R}}

\newcommand{\cL}{\mathcal{L}}

\newcommand{\cN}{\mathcal{N}}

\newcommand{\PP}{\mathbb{P}}

\newcommand{\rmd}{\mathrm{d}}

\DeclareMathOperator{\diag}{\mathrm{diag}}

\DeclareMathOperator{\tr}{\mathrm{tr}}

\newcommand{\TV}{\mathsf{TV}}

\newcommand{\1}{\mathbf 1}

\newcommand{\Var}{\operatorname{Var}}
\newcommand{\Cov}{\operatorname{Cov}}
\newcommand{\osc}{\operatorname{osc}}
\newcommand{\Hel}{\mathsf H^2}

\newcommand{\norm}[1]{\lVert#1\rVert}
\newcommand{\xs}{x^\star}
\newcommand{\Xs}{X^\star}
\newcommand{\Xh}{\widehat X}
\newcommand{\sh}{\widehat s}
\newcommand{\Esc}{\mathcal E_{\sf sc}}
\newcommand{\Gauss}{\mathcal N}
\newcommand{\lat}{\mathcal P}
\newcommand{\Kraw}{\widehat K}

\allowdisplaybreaks
\makeatletter

\let\epsilon\varepsilon
\newcommand{\matTol}{\varepsilon_{\rm mat}}

\theoremstyle{plain}\newtheorem{lemma}{\textbf{Lemma}}
\newtheorem{proposition}{\textbf{Proposition}}
\newtheorem{theorem}{\textbf{Theorem}}

\theoremstyle{definition}\newtheorem{definition}{\textbf{Definition}}

\theoremstyle{remark}\newtheorem{remark}{\textbf{Remark}}

\newcolumntype{M}[1]{>{\centering\arraybackslash}m{#1}}

\begin{document}
\title{Diffusion Models are Provably Efficient Priors for Sparse Recovery }

 \author
 {
 	Ivan Lee\thanks{Department of Statistics and Data Science, Yale University; Email: \texttt{\{ivan.lee, yuejie.chi\}@yale.edu}.} \\
    Yale
		\and
 	 Xingyu Xu\thanks{Department of Electrical and Computer Engineering, Carnegie Mellon University; Email:
 	 	\texttt{xingyuxu@andrew.cmu.edu}.}  \\
		CMU \\
		\and
		Yuejie Chi\footnotemark[1]
		\thanks{This work is supported in part
        by the Air Force Office of Scientific Research under award number FA9550-25-1-0060, and by National Science Foundation under ECCS-2126634. Any opinions, findings, and conclusions or recommendations expressed in this material are those of the author(s) and do not necessarily reflect the views of the United States Air Force. }  \\
		Yale
 }

\date{November 2026}

\setcounter{tocdepth}{2}
\maketitle

\begin{abstract}

Diffusion models have transformed generative modeling through their ability to capture and generate samples from complex data distributions. This success has spurred substantial interest in using pretrained diffusion models as expressive priors for highly ill-posed inverse problems in imaging and sensing. Despite strong empirical performance, theoretical understanding of this approach remains limited: existing guarantees are often asymptotic or entail prohibitive computational costs in the worst case. In this paper, we address this gap by studying sparse recovery from a small number of linear measurements, a classical problem in compressed sensing. We show that, given an accurate diffusion model capturing a sparsity-promoting Bernoulli-Gaussian prior, a likelihood-guided annealed Langevin sampler provably recovers the sparse signal in polynomial time with near-optimal sample complexity for a broad family of random and deterministic sensing matrices. When the signal is indeed drawn from the postulated prior, we further establish total variation guarantees for posterior sampling after a final denoising step. These results lay a theoretical foundation for statistically and computationally efficient inverse solvers with diffusion priors.

\end{abstract}

\noindent\textbf{Keywords:} diffusion models, posterior sampling, sparse recovery, compressed sensing

\tableofcontents{}

\section{Introduction}
\label{sec:intro}

Diffusion models \citep{sohl2015deep,song2019generative,song2020score,ho2020denoising,song2020denoising,dhariwal2021diffusion}, as the backbone of modern image and video generation engines \citep{peebles2023scalable}, are one of the driving forces behind the recent remarkable advances of generative artificial intelligence.
Since their debut, diffusion models have been adopted across many scientific and engineering domains, such as biological structure prediction \citep{abramson2024accurate,trippe2022diffusion}, robotics \citep{chi2023diffusionpolicy,janner2022planning}, materials \citep{dong2026multimodal}, and many others.

A central idea behind score-based diffusion models is to perturb the data with progressively stronger Gaussian noise and learn the associated score functions, namely the gradients of the log densities at different noise levels, for instance, via score matching \citep{hyvarinen2005estimation}. These scores then guide an iterative denoising process that transforms noise into samples resembling those from the data distribution \citep{song2019generative,song2020score}. Complementing this empirical success, a flurry of recent theory has established polynomial bounds on the number of sampling steps required when the score functions are accurate, without imposing any smoothness or log-concavity assumptions on the data distribution \citep[see, e.g.,][]{chen2023sampling,benton2024nearly,lee2022convergence,li2023towards}.

\subsection{Diffusion models as priors}

The expressive power of diffusion models makes them attractive priors for ill-posed inverse problems, where the goal is to recover high-dimensional signals from a small number of noisy measurements. Of particular interest is the plug-and-play framework \citep{buzzard2018plug}, in which the same pretrained diffusion model is reused to solve different instances of inverse problems. This flexibility is especially valuable in domains where the same class of signals is observed through different sensing modalities and under varying noise conditions, allowing the substantial cost of training a diffusion model to be amortized across reconstruction tasks.

Posterior sampling provides a natural approach to use a diffusion prior for inverse problems, where one can generate multiple samples from the resulting posterior as candidate reconstructions. However, turning an unconditional diffusion model into a posterior sampler introduces additional difficulties. Running the reverse diffusion directly for posterior sampling requires the scores of the measurement-conditioned distributions at intermediate noise levels, which are not supplied by the pretrained diffusion model. Many empirical methods are thus developed to circumvent this limitation, and have achieved qualitatively impressive performance \citep[see, e.g.,][]{jalal2021robust,aali2023solving,song2022pseudoinverse,song2023loss}.

Despite these empirical advances, the theoretical understanding of diffusion-based inverse solvers remains severely limited. Existing algorithms often come only with asymptotic consistency for sampling from the true posterior \citep{dou2024diffusion,cardoso2023monte,wu2023practical,xu2024provably,jiao2026provable}. When it comes to non-asymptotic guarantees, they can require exponentially many sampling steps in the worst case \citep{xu2024provably}, rely on strong assumptions on the algorithm's initialization \citep{leong2025recovery,xun2025posterior}, or constrain the data distribution to strict log-concavity \citep{xun2025posterior} or small susceptibility \citep{bruna2024posterior} to allow more efficient sampling guarantees. Indeed, under standard computational hardness assumptions, \citet{gupta2024diffusion,bruna2024posterior} showed that posterior sampling can be intractable even for linear inverse problems whose unconditional diffusion models admit efficient sampling. These computational barriers rule out a general polynomial-time guarantee for diffusion-based inverse solvers, and motivate the search for structured inverse problems that admit efficient inference and are more representative of where such solvers succeed in practice.

\subsection{Our contribution}

To understand when diffusion priors admit efficient inference, we adopt sparse recovery from noisy linear measurements, a classical problem in compressed sensing \citep{donoho2006compressed,candes2006near,candes2006stable}, as our testbed. Suppose we have access to a collection of measurements given as,
$$ y =  A \xs + \sigma\zeta, \qquad   \quad  \zeta \sim \cN(0, I_m),  $$
where $A\in\mathbb{R}^{m\times n}$ is the measurement or sensing matrix, $x^{\star}\in\mathbb{R}^n$ is the ground truth signal of interest, $\sigma\zeta\in\mathbb{R}^m$ is the additive noise with noise level $\sigma>0$.
We assume access to a pretrained diffusion model that captures a sparsity-promoting Bernoulli--Gaussian prior, under which each coordinate is either zero or drawn from a Gaussian distribution. This setting retains the key challenge associated with a non-log-concave data prior, while remaining tractable enough to yield practically meaningful insight into the statistical and computational efficiency of diffusion-based inverse solvers beyond the worst case.

Concretely, we analyze a simple likelihood-guided annealed Langevin algorithm \citep{jalal2021robust} that takes Langevin steps along a tilted target that smooths both the prior and the likelihood term in the posterior, with the smoothing strength annealed at every step. We show the algorithm succeeds within a polynomial number of iterations at near-optimal sample complexities for a broad family of sensing matrix designs, including sub-Gaussian random matrices, randomly subsampled incoherent orthonormal bases, and deterministic incoherent tight frames. Our main results are as follows.

\begin{itemize}
\item For fixed $k$-sparse signals, under suitable noise and signal strength conditions, we show that within $\widetilde O\bigl(n^6/(m k^2\sigma^2\varepsilon^3)\bigr)$ iterations, the algorithm with a postulated Bernoulli--Gaussian prior at the expected sparsity level $k$ yields a near-optimal $\ell_2$ recovery error of order
\[
 \sigma\left(\sqrt{k}+\log\frac{en}{\varepsilon}\right),
\]
with probability at least $1-\varepsilon-C\Esc(\xs)$, where $\Esc(\xs)$ is the score estimation error (Theorem~\ref{thm:application}).
\item For sparse signals drawn from the Bernoulli--Gaussian prior with expected sparsity level $k$ and noise level $0<\sigma<1$, we further establish that within $\widetilde O\bigl(n^6/(m k^2\sigma^2\varepsilon^6)\bigr)$ iterations, the algorithm using the same prior  achieves an expected total variation error with respect to the true posterior of order
\[
 \varepsilon+\sqrt{\Esc}
\]
after a final denoising step, where $\varepsilon$ is the target accuracy, $\Esc$ is the averaged score estimation error (Theorem~\ref{thm:sampling}).
\end{itemize}
Both guarantees degrade gracefully as the score estimation error grows. To the best of our knowledge, our result provides the first polynomial-time posterior sampling algorithm at the near-optimal sample complexity linear in the expected sparsity level using random sensing ensembles, answering an open problem raised in \citet{KSTZ26}. Together, these results identify a concrete and benign setting in which a simple diffusion-based solver provably achieves accurate recovery and posterior sampling under a non-log-concave prior, with guarantees that are efficient both statistically and computationally.

\subsection{Related works}
\label{sec:related}

Several recent works have sought non-asymptotic convergence guarantees for diffusion-based posterior sampling in general inverse problems. The convergence rate in \citet{xu2024provably} depends critically on the spectral gap of a Markov chain tied to the dynamics of their underlying algorithm, which can become exponentially small in the problem dimension in the worst case. \citet{leong2025recovery} developed a deterministic recovery theory leveraging the connection between diffusion and denoising operators \citep{kadkhodaie2021stochastic}. However, their theory does not provide a convergence guarantee with polynomial complexity, since it hinges on an unspecified initial burn-in phase.
A notable recent development is \citet{xun2025posterior}, which developed a provably efficient posterior sampling algorithm based on annealed Langevin dynamics with polynomial iteration complexities, assuming a log-concave prior. They subsequently went beyond log-concavity by assuming access to direct measurements of the signal of interest at sufficiently low noise levels, which are generally unavailable in practice. Our work removes these restrictions in the setting of sparse recovery with few measurements. Last but not least, \citet{xu2026fantastic} showed that diffusion-based posterior sampling is efficient for sparse signal denoising, corresponding to the special case of compressed sensing with an identity sensing matrix. The present paper builds on \citet{xu2026fantastic}, but significantly goes beyond it by handling general sensing matrices.

 We also note that Bayesian approaches have long been applied to high-dimensional sparse linear regression, e.g., \citet{george1993variable,ishwaran2005spike,yang2016computational}. Due to the challenge of sampling from multi-modal distributions, recent efforts have focused on developing efficient sampling algorithms tailored to sparsity-inducing distributions, such as the spike-and-slab prior \citep{montanari2025provably,kumar2025spike,jiang2023estimation,KSTZ26}. Although the goal of our work is not to compete with existing sparse recovery algorithms, it offers a complementary--and competitive--algorithmic perspective with efficient sampling guarantees, and highlights the benefit of annealing provided naturally by diffusion priors.

\paragraph{Notation.} Before proceeding, we introduce several useful notation used throughout the paper. Define the sub-Gaussian norm of a real random variable \(R\) as
\[
 \norm R_{\psi_2}\coloneqq\inf\{v>0:\E e^{R^2/v^2}\le2\}.
\]
Throughout the paper, we use \(c\) and \(C\) to denote positive universal constants whose values may change between occurrences, where $c$ is typically taken to be small enough and $C$ to be large enough. For a matrix \(B\), we let $B_{ij}$ be its $(i,j)$-th entry, $B_j$ be its $j$-th column, and for index sets \(J\) and \(K\),  let \(B_{JK}\) be the
submatrix with rows in \(J\) and columns in \(K\), and $B_J$ be the submatrix with columns in $J$. The diagonal part of a matrix $B$ is denoted by \(\diag B\), while \(\diag(c_i)\) is the diagonal matrix with entries \(c_i\).  We write \([n]\coloneqq\{1,\ldots,n\}\) and \(\supp x\coloneqq\{i:x_i\ne0\}\). Let $\|x\|$ be the $\ell_2$ norm of a vector $x$, and $\|A\|_{\rm F}$, $\|A\|$ be the Frobenius norm and the spectral norm of a matrix $A$, respectively. The law of a random variable \(X\) is denoted by \(\mathcal L(X)\). The product \(\mu\otimes\nu\) is the joint law of independent draws with laws \(\mu,\nu\); the law of their sum has convolution law \(\mu*\nu\).

\section{Problem formulation}
\label{sec:background}

We aim at solving an underdetermined linear inverse problem of the form
\begin{equation}
\label{eqn:inv}
y =  A \xs + \sigma\zeta, \qquad   \quad  \zeta \sim \cN(0, I_m),
\end{equation}
where $A \in \reals^{m \times n}$ is the sensing or measurement matrix, $\xs$ is the signal of interest, and $\sigma\zeta \in\reals^{m}$ is the additive Gaussian noise independent of the signal, with a {\em known} noise level $\sigma>0$.

\paragraph{Sparse signal prior.} Our canonical model of the signal prior is the \emph{Bernoulli--Gaussian model}, which provides a simple and analytically convenient description of sparsity.

\begin{definition}[Bernoulli--Gaussian model]
Let $s, n \in \mathbb{N}$, with $s < n$. A random vector
\[
X = (X_1,\dots,X_n) \in \mathbb{R}^n
\]
is said to follow the \emph{Bernoulli--Gaussian model} with sparsity level $s$ and dimension $n$, denoted as $\mathsf{BG}(s, n)$, if its coordinates are independent and identically distributed according to
\[
X_i =B_i Z_i  \sim \left(1-\frac sn\right)\,\delta_0 + \frac sn\,\mathcal{N}(0,1),
\qquad i=1,\dots,n,
\]
where $\delta_0$ is a point mass at zero, $B_i \sim \mathrm{Bernoulli}(\frac sn)$ and $Z_i \sim \mathcal{N}(0, 1)$ are all mutually independent.
\end{definition}
Each coordinate of \(X\) is thus zero with probability \(1-s/n\) and standard Gaussian otherwise. The expected support size of $X$ is $\mathbb{E}\bigl[|\operatorname{supp}(X)|\bigr] = s$, and therefore, we refer to $s$ as the sparsity level.

\paragraph{Posterior sampling with diffusion models}
We consider a Bayesian approach to signal recovery, even when the underlying ground truth signal $\xs$ may not be generated from a probabilistic model.
We postulate the ground truth signal $\xs$ follows a prior distribution $p_\star $, for example, $\mathsf{BG}(s, n)$ for some postulated sparsity level $s$. We attempt to solve the inverse problem in \eqref{eqn:inv} by sampling from the posterior distribution of $\xs$ given $y$, using Bayes' formula:
\begin{equation} \label{eq:posterior}
p_{\star}(x | y) \propto p_\star(x) \, p(y | \xs = x) = p_\star(x) \, \exp\left(\cL(x; y)\right),
\end{equation}
where the log-likelihood function of the measurement model \eqref{eqn:inv} is given by
\begin{equation}
\label{eqn:likelihood}
\cL(x; y) \coloneqq -\frac{1}{2\sigma^2} \| y -  Ax  \|^2 + C_\sigma.
\end{equation}
Here, $C_\sigma$ is a constant depending only on $\sigma$.

Rather than developing a posterior sampler tailored to the sparse prior, we instead use sparse recovery as a testbed for a broader question: can a diffusion model \citep{song2020score}, which describes \(p_\star\) only implicitly, enable provably efficient posterior sampling? Concretely, consider the smoothed prior
\begin{equation}\label{eq:smoothed_prior}
p_t\coloneqq    p_\star*\cN(0,tI_n), \qquad t>0,
\end{equation}
which is the law of $X+\sqrt t\,\xi$, where $X\sim p_\star$ and $\xi\sim\cN(0,I_n)$ are independent. We assume access to an estimate \(\sh_t(x)\) of the \emph{score function}
$$ s_t(x) \coloneqq \nabla\log p_t(x)$$
of the smoothed prior $p_t$, which is well-defined since $p_t$ is smooth and positive. In practice, the score functions can be learned, for instance, via score matching \citep{hyvarinen2005estimation}.  Given accurate scores, many diffusion samplers provably generate samples whose law is close to \(p_\star\) \citep[see, e.g.,][]{benton2024nearly,chen2023sampling,lee2022convergence,li2023towards}, and hence, the diffusion model (as specified by \(\sh_t(x)\)) encodes the prior information. Nonetheless, we cannot directly access an analytical form of $p_\star$.

\paragraph{Goal.} Our goal is to recover the signal of interest $\xs$ from the measurement $y$, assuming access to a diffusion model (as specified by \(\sh_t(x)\)), in a provably efficient manner, both statistically and computationally.

\smallskip

\begin{remark}[Basis information in the score]
When signals are sparse in a fixed orthonormal basis \(\Psi\), such as a wavelet or discrete cosine basis, classical \(\ell_1\) methods \citep{donoho2006compressed,candes2006near} use \(\Psi\) to form the penalty \(\|\Psi^\top x\|_1\). A diffusion sampler can use scores learned directly from these signals without taking \(\Psi\) as an input. Remark~\ref{rem:unknown-basis} gives conditions for recovery in this setting.
\end{remark}

\section{Diffusion-based inverse solver}
 \label{sec:algorithm}

Many diffusion-based posterior samplers add a guidance term to the reverse-time dynamics of the diffusion model, which steers the iterates to be consistent with the measurements \citep{chung2023diffusion}. For tractability, we analyze one of the simplest schemes of this type, annealed Langevin dynamics with a likelihood guidance term, which follows the form originally proposed in \citet{jalal2021robust} but takes carefully chosen annealing schedules and parameters. While this may not be the fastest horse in the race, its simplicity makes a precise analysis possible for deeper mathematical insights.

\paragraph{Tilted targets.} To begin with, consider a family of tilted targets. For each level \(t>0\), let
\begin{equation}\label{eq:target}
 \Pi_t(x)\propto p_t(x)\exp\left(-\frac{\norm{Ax-y}^2}{2a_t}\right),\qquad a_t\coloneqq\sigma^2+2t,
\end{equation}
where \(p_t\) is the smoothed prior in \eqref{eq:smoothed_prior}. It is easy to check that \(\Pi_t\) is the conditional law of \(X\) given the prior \(X \sim p_t\) and \(y=AX+w\), where \(w\sim\cN(0,a_tI_m)\) is independent of \(X\), with the variance \(a_t\) adding a smoothing term \(2t\) to the noise variance \(\sigma^2\). Decreasing the level \(t\) moves the tilted target $ \Pi_t(x)$ from a nearly Gaussian law (since \(p_t\) is close to \(\cN(0,tI_n)\) when \(t\) is large), toward the posterior of the signal given \(y\) (since \(p_t\to p_\star\) and \(a_t\to\sigma^2\) as \(t\to0\)).

At a fixed level \(t\), the target \(\Pi_t\) is the stationary law of the Langevin diffusion
\begin{equation}\label{eqn:ps}
 \rmd X_u=\left(\nabla\log p_t(X_u)+\frac{A^\top(y-AX_u)}{a_t} \right)\rmd u+\sqrt2\,\rmd B_u,\qquad u\ge0,
\end{equation}
where \((B_u)_{u\ge0}\) is a standard Brownian motion in \(\reals^n\), and the drift is the score of the tilted target
\begin{equation}\label{eq:target-drift}
 \nabla\log\Pi_t(x)=\nabla\log p_t(x)+\frac{A^\top(y-Ax)}{a_t}.
\end{equation}

\paragraph{Score-based annealed Langevin sampler.} Our sampler discretizes \eqref{eqn:ps} while decreasing the level along a schedule \(T=t_0>t_1>\dots>t_N=\tau\) specified below for a total of $N$ steps, with the prior score \(\nabla\log p_t(\cdot)\) replaced by the estimate \(\sh_t(\cdot)\). Starting from an initial iterate \(\Xh_0\), it takes one Euler--Maruyama step of size \(h_j\) at level \(t_j\),
\begin{equation}\label{eq:algorithm}
 \Xh_{j+1}=\Xh_j+
 h_j\left[\sh_{t_j}(\Xh_j)+\frac{A^\top(y-A\Xh_j)}{a_{t_j}}\right]
 +\sqrt{2h_j}\,\xi_j,\qquad 0\le j<N,
\end{equation}
where \(\xi_0,\dots,\xi_{N-1}\) are i.i.d.\ \(\cN(0,I_n)\) vectors.  At high levels, both the smoothed prior and the tilted target are close to broad Gaussian laws, so the updates are essentially Langevin steps toward a Gaussian target and the iterates move freely. As the level decreases, the two terms in the drift acquire distinct roles. The prior score pulls the iterates toward sparse signals, while the guidance pulls them toward signals consistent with the observation. Together, these two forces localize the final iterate near the true signal.

Fix an iteration budget \(N\) and a terminal level \(0<\tau\le1\). We use the following schedule, step sizes, and initialization.
\begin{itemize}
    \item \emph{Geometric diffusion schedule.} Starting at \(T=n/\tau\), decrease the diffusion level geometrically to \(\tau\),
    \begin{equation}\label{eqn: exp-schedule}
      t_j=T(\tau/T)^{j/N},\qquad 0\le j\le N .
    \end{equation}
    Such schedules are standard for variance-exploding diffusion models \citep{song2019generative,song2020score}.
    \item  \emph{Level-dependent step sizes.} Take
    \begin{equation}\label{eq:h}
       h_j=\gamma t_j,\qquad
       \gamma=\left[\frac{m}{nN^2}
       \left(1+\frac{n}{s\sigma\max\{1,\sigma\}}\right)\right]^{1/3}.
    \end{equation}
The steps size $h_j$ is selected to be linear in $t_j$, and as shall be seen, the choice of \(\gamma\) balances the accumulated Euler error with the change in targets along the discretization grid.
    \item \emph{Initialization.} The initial iterate \(\Xh_0\) is drawn, given the data, from the Gaussian law with density proportional to
\begin{equation}\label{eq:init}
 \exp\left(-\frac{\norm x^2}{2T}-\frac{\norm{Ax-y}^2}{2a_T}\right),
\end{equation}
 that is, from \(\cN(\mu_0,\Sigma_0)\) with \(\Sigma_0\coloneqq(I_n/T+A^\top A/a_T)^{-1}\) and \(\mu_0\coloneqq\Sigma_0A^\top y/a_T\), which can be sampled exactly. This law is the first target \(\Pi_T\) with \(p_T\) replaced by \(\cN(0,TI_n)\), and a large initial level \(T=n/\tau\) makes the two close.
\end{itemize}

\section{Main theorems}
\label{sec:main_theorem}

We give recovery guarantees for fixed sparse signals in Theorem~\ref{thm:application} and posterior sampling guarantees in Theorem~\ref{thm:sampling}.

\paragraph{Sensing matrix ensemble.}
The following definition collects the sensing matrix ensembles covered by our theory, including both random and deterministic sensing matrix designs.
\begin{definition}[Matrix families]\label{def:families}
Let $A=[A_{ij}]\in  \reals^{m \times n}$ with \(m\le n\), which is independent of the signal and the noise.
\begin{enumerate}
\item \emph{Independent sub-Gaussian entries.}
The variables \(\sqrt m A_{ij}\) are independent and identically distributed, centered, unit-variance, and of
\(\psi_2\) norm at most \(\kappa\geq 1\).
\item \emph{Randomly subsampled bases.}
Let \(U\in\reals^{n\times n}\) be a fixed orthonormal matrix with \(|U_{ij}|\le \kappa_b/\sqrt n\), where \(\kappa_b\ge1\). Let \(A=\sqrt{n/m}\,R_{\mathcal I}U\), where \(\mathcal I\) is a uniformly random subset of \([n]\) of size \(m\) and \(R_{\mathcal I}\)
extracts the rows of \(U\) indexed by \(\mathcal I\).
\item \emph{Deterministic frames.}
Let \(A\) be a fixed incoherent tight frame satisfying
\[
 \norm{A_i}=1,\qquad AA^\top=(n/m)I_m,\qquad
\mu(A) := \max_{i\ne j}|\langle A_i,A_j\rangle|\le\kappa_f/\sqrt m.
\]
Here \(\kappa_f\ge1\) is a fixed constant, \(A_i\) is the \(i\)-th column of \(A\), and \(\mu(A)\) is the \emph{coherence} of \(A\).
\end{enumerate}
\end{definition}

Definition~\ref{def:families} encompasses several popular models considered in the literature \citep{donoho2006compressed,candes2006near,donoho2005stable,bajwa2012two}, for example, Gaussian and Rademacher matrices, subsampled Hadamard and DCT matrices, and incoherent frames such as Gabor and equiangular tight frames.

\paragraph{Score estimation error.}
We measure the accuracy of the score estimate by its \(L^2\) error along the tilted targets \eqref{eq:target}, summed over the levels used by the sampler.
For a fixed matrix \(A\) and signal \(\xs\), define
\begin{equation}\label{eq:score-error-def}
 \Esc(A,\xs)\coloneqq
 \E_\zeta\sum_{j<N}h_j\int
 \norm{\sh_{t_j}(x)-\nabla\log p_{t_j}(x)}^2\,\Pi_{t_j}(dx),\qquad y=A\xs+\sigma\zeta .
\end{equation}
Here, \(h_j\) is the step size in \eqref{eq:h}, \(\Pi_{t_j}\) is the tilted target for the data \((A,y)\), and the expectation is over the measurement noise. Averaging over the measurement matrix gives the score error \(\Esc(\xs) \coloneqq\mathbb{E}_{A}  \Esc(A,\xs) \) at a fixed signal \(\xs\), and averaging further over a random signal \(\Xs\sim p_\star\) gives the score error \(\Esc\coloneqq\E_{\Xs}\,\Esc(\Xs)\).

Assumptions of this type are standard in the analysis of diffusion samplers, where the score error is usually measured under the marginals \(p_t\) \citep[see, e.g.,][]{benton2024nearly}. In an inverse problem, the tilted distribution \(\Pi_t\) instead reflects the information carried by the observation \(y\), and is used in collecting the score estimation error \citep{bruna2024posterior}. Due to annealing, the assumption captures the intuition that, as the inverse problem solver approaches the true signal, score accuracy in a neighborhood of \(\xs\) becomes increasingly important.

\subsection{Recovery guarantees for fixed signals}
We first consider recovery of a fixed sparse signal, without assuming that it follows the prior.

\begin{theorem}[Recovery of a fixed signal]\label{thm:application}
Suppose \(\xs\in\R^n\) is a $k$-sparse signal with support \(S\), observed through \eqref{eqn:inv} with a measurement matrix from one of the families in Definition~\ref{def:families}. Run the Langevin sampler \eqref{eq:algorithm}--\eqref{eq:init} with prior \(p_\star=\mathsf{BG}(s,n)\), where \(1\le s\le n/2\). Fix \(0<\varepsilon<1\). There are universal constants \(c,C>0\) such that the following holds. Suppose that
\begin{equation} \label{eq:param}
 \tau\le c\min\Bigl\{\frac{\sigma^2}{n},\varepsilon\Bigr\},\qquad s \leq \frac{c}{\sigma \sqrt{\log(en/\varepsilon)}},
\end{equation}
and
\begin{equation}\label{eq:signal-admission}
\min_{i \in S} |x_i^\star|\ge C\sigma\sqrt{\log\!\frac{en}{\varepsilon\sigma}},\qquad
 \sigma\|\xs\|\le c\Bigl(\sqrt{k}+\log\frac{en}{\varepsilon}\Bigr).
\end{equation}
As long as the number of measurements satisfies
\begin{equation}\label{eq:sample-size}
 m \geq C \cdot    \begin{cases}
 \kappa^8 (1+s+k+\|\xs\|^2) \log(en/(\varepsilon\sigma))^4,&\text{for i.i.d. sub-Gaussian matrices},\\
 \kappa_b^2(1+s+k+\|\xs\|^2) \log(en/(\varepsilon\sigma))^7,&\text{for randomly subsampled bases},\\
 \kappa_f^2(1+s+k+\|\xs\|^2)^2 \log(en/(\varepsilon\sigma))^4,&\text{for deterministic frames},
 \end{cases}
\end{equation}
and
\begin{equation}\label{eq:ald-recovery-count}
 N\ge C\,\frac{n^6}{m s^2\sigma^2\varepsilon^3}
 \log^6\!\left(\frac{en}{\tau}\right),
\end{equation}
then with probability at least \(1-\varepsilon-C\Esc(\xs)\) over the measurement matrix, the noise, and the algorithmic randomness, we have
\begin{equation}\label{eq:radii}
 \left\|\Xh_N-\xs \right\|\le C\sigma\Bigl(\sqrt{k}+\log\frac{en}\varepsilon\Bigr).
\end{equation}
\end{theorem}

We first comment on the parameter choices in \eqref{eq:param}. The terminal level \(\tau\) must be small relative to the noise level, so that the smoothing at the final level adds noise comparable to the measurement noise. On the other hand, the prior sparsity level $s$ needs to scale inverse proportionally to the noise level,  in order to prevent spurious contributions from the noise into the sparse support.
Turning to the condition \eqref{eq:signal-admission}, it prescribes two competing forces on the ground truth signal. The first condition says that the minimum signal strength needs to be not too small with respect to the noise level, which allows reliable detection of its support from noise. The second condition controls the shrinkage bias of the signal from the unit-variance Gaussian slab for nonzero entries to be within the recovery radius.

Under these suitable signal and noise conditions, Theorem~\ref{thm:application} establishes that, a simple likelihood-guided annealed Langevin sampler can recover fixed sparse signals at the near minimax-optimal rate (ignoring logarithmic factors) \citep{zhang2010nearly,candes2013well}
$$  \left\|\Xh_N-\xs\right\|\lesssim \sigma \sqrt{k} $$
from a small number of measurements using polynomially many score evaluations, despite the lack of log-concavity of its prior. In particular, the sparse signals do not necessarily need to agree with typical signals drawn from the prior, e.g., by allowing the sparsity level $k$ of the true signal to be different from the sparsity level $s$ postulated by the prior. A few discussions are in order.
\begin{itemize}
\item \textbf{Sample efficiency.} As long as the sparsity level of the Bernoulli-Gaussian prior satisfies $s  =O( k )$ and the signal strength satisfies \(\|\xs\|^2=O(k)\), then the proposed algorithm succeeds when the sample complexity $m$ is on the order of $O(k)$ for random matrix ensembles (including sub-Gaussian sensing matrices or randomly subsampled bases), and on the order of $O(k^2)$ for deterministic frames, up to logarithmic factors. For the random ensembles, this sample complexity is near optimal with respect to the sparsity level \citep{donoho2006compressed,candes2006near}, and for deterministic frames it matches the recovery guarantees based on the worst-case coherence \citep{donoho2005stable}.

\item \textbf{Computational efficiency.} Each iteration of the sampler takes one score function evaluation, and the required number of iterations \eqref{eq:ald-recovery-count} is  $\widetilde{O}\left(\frac{n^6}{m s^2\sigma^2\varepsilon^3}\right)$, which scales polynomially in $n$,  $1/\sigma$ and $1/\varepsilon$. This certifies diffusion models as computationally efficient prior models in sparse recovery {\em for the first time}, without requiring a warm start or unknown burn-in periods \citep{xun2025posterior,leong2025recovery}.

\item \textbf{Score estimation error.}
With exact scores, the success probability in Theorem~\ref{thm:application} is at least \(1-\varepsilon\). For approximate scores, the additional failure probability is controlled by \(\Esc(\xs)\), which degrades gracefully with the increase of the score estimation error.
\end{itemize}

\begin{remark}[Recovery in an unknown orthonormal basis]\label{rem:unknown-basis}
Suppose \(\xs=\Psi\theta^\star\), where \(\theta^\star\) is sparse and \(\Psi\) is orthonormal.
Since \(\cN(0,tI_n)\) is invariant under orthonormal transformations, the smoothed prior of the diffusion model is \(x\mapsto p_t(\theta)\), where $\theta= \Psi^\top x$, and its score is \(\Psi (\nabla\log p_t)(\Psi^\top x)\). Multiplying \eqref{eq:algorithm} by \(\Psi^\top\) shows that the iterates \(\widehat\theta_j\coloneqq \Psi^\top\Xh_j\) satisfy
\[
 \widehat\theta_{j+1}=\widehat\theta_j+h_j\left[\sh_\theta(t_j,\widehat\theta_j)+\frac{(A\Psi)^\top(y-A\Psi \widehat\theta_j)}{a_{t_j}}\right]+\sqrt{2h_j}\,\Psi^\top\xi_j,
\]
where \(\sh_\theta(t,\theta) \) estimates \(\nabla\log p_t\), and the vectors \(\Psi^\top\xi_j\) are again i.i.d.\ \(\cN(0,I_n)\). This is the sampler for the prior \(\mathsf{BG}(s,n)\) with the measurement matrix \(A\Psi\), and the initialization \eqref{eq:init} and the score error \eqref{eq:score-error-def} transform in the same way.
When \(A\) has i.i.d.\ Gaussian entries, \(A\Psi\) has the same law as \(A\), and Theorem~\ref{thm:application} applied to \(\theta^\star\) yields the bound \eqref{eq:radii} for \(\|\Xh_N-\xs\|=\|\widehat\theta_N-\theta^\star\|\), with the signal conditions \eqref{eq:signal-admission} imposed on \(\theta^\star\). More generally, the conclusion holds whenever \(A\Psi\) belongs to one of the families in Definition~\ref{def:families}, which for randomly subsampled bases requires \(U\Psi\) to satisfy the incoherence condition.
\end{remark}

\subsection{Posterior sampling guarantees}\label{sec:sampling}

We now turn to the case that the ground truth signal indeed follows the prior, \(\Xs\sim p_\star =\mathsf{BG}(s,n)\), independently of the measurement matrix. In this case, we are interested in obtaining posterior sampling guarantees, which compare the law of the sampler output to the posterior \(p_\star(\cdot\mid y)\) in \eqref{eq:posterior} in total variation.

For the sparse recovery task with Bernoulli-Gaussian priors, the vanilla annealed Langevin sampler does not readily yield a posterior sampling guarantee. This is because the Gaussian perturbation in the final update \eqref{eq:algorithm} makes \(\Xh_N\) nonzero with probability one for all coordinates, whereas the posterior in \eqref{eq:posterior} assigns positive mass to vectors with zero coordinates. To obtain a total-variation guarantee, we follow the Langevin updates with an endpoint draw that is defined as follows.

\begin{definition}[Endpoint kernel and endpoint draw]\label{def:endpoint}
For \(t>0\) and \(x\in\R^n\), the \emph{endpoint kernel} \(\mathcal D_t(x,\cdot)\) is the conditional law of \(Z\) given \(Z+\sqrt t\,\xi=x\), where \(Z\sim p_\star\) and \(\xi\sim\Gauss(0,I_n)\) are independent. The \emph{endpoint draw} is
\(\widehat Z\sim\mathcal D_\tau(\Xh_N,\cdot)\), generated using fresh randomness independent of the data and the Langevin chain.
\end{definition}

In other words, the endpoint draw applies an additional posterior denoising step on $\Xh_N$, assuming additive Gaussian noise at the noise level $\tau$.
For the Bernoulli--Gaussian prior $\mathsf{BG}(s,n)$, the endpoint draw admits a simple closed-form expression. Each coordinate of \(z\sim\mathcal D_t(x,\cdot)\) is independently generated by
\begin{equation}\label{eq:endpoint}
 z_i= \begin{cases}
 0\qquad  \text{with probability }1-\varsigma_t(x_i), \\
 \Gauss\Bigl(\frac{x_i}{1+t},\frac t{1+t}\Bigr)
 \ \text{otherwise},
 \end{cases}
\end{equation}
where
\[
 \varsigma_t(v)\coloneqq\Bigl[1+\frac{n-s}s\sqrt{1+1/t}\;e^{-v^2/[2t(1+t)]}\Bigr]^{-1}
\]
is the conditional probability that a coordinate is nonzero given its noisy value \(v\).
In general, the endpoint draw can be implemented efficiently using the same diffusion model for unconditional generation, as discussed in \citet{xu2024provably}. For simplicity, we assume the endpoint draw is implemented exactly using \eqref{eq:endpoint}.
Armed with this endpoint draw, we are ready to present the posterior sampling guarantee in total variation.

\begin{theorem}[Posterior sampling]\label{thm:sampling}
Suppose \(\Xs\sim p_\star=\mathsf{BG}(s,n)\), with \(1\le s \le n/2\), is observed through \eqref{eqn:inv} with a measurement matrix from one of the families in Definition~\ref{def:families}  independent of \(\Xs\). Fix \(0<\varepsilon<1\). There are universal constants \(c,C>0\) such that the following holds. Suppose that
\begin{equation}  \label{eq:sampling-resources}
 \tau\le c\min\left\{\varepsilon^2,\frac{\varepsilon\sigma^2}{n^2(1+s/\sigma^2)}\right\},
\end{equation}
and
\begin{equation}\label{eq:ald-sampling-count}
 N\ge C\,\frac{n^4}{m\varepsilon^6}
 \left(1+\frac{n}{s\sigma\max\{1,\sigma\}}\right)^2
 \log^6\!\left(\frac{en}{\tau}\right).
\end{equation}
As long as the number of measurements satisfies
\begin{equation}\label{eq:sampling-sample-size}
 m\ge C\cdot\begin{cases}
 \kappa^8\bigl(s+\log(1/\varepsilon)\bigr)\log(en/(\varepsilon\min\{1,\sigma\}))^4,&\text{for i.i.d. sub-Gaussian matrices},\\
 \kappa_b^2\bigl(s+\log(1/\varepsilon)\bigr)\log(en/(\varepsilon\min\{1,\sigma\}))^7,&\text{for randomly subsampled bases},\\
 \kappa_f^2\bigl(s+\log(1/\varepsilon)\bigr)^2\log(en/(\varepsilon\min\{1,\sigma\}))^4,&\text{for deterministic frames},
 \end{cases}
\end{equation}
 the annealed Langevin sampler \eqref{eq:algorithm}--\eqref{eq:init} followed by the endpoint draw \(\widehat Z\sim\mathcal D_\tau(\Xh_N,\cdot)\) satisfies
\[
 \E\,\TV\bigl(\mathcal L(\widehat Z\mid A,y),p_\star(\cdot\mid y)\bigr)\lesssim \varepsilon+ \sqrt{\Esc},
\]
where \(p_\star(\cdot\mid y)\) is the true posterior in \eqref{eq:posterior}. Here, \(\Esc\) is the averaged score error defined in \eqref{eq:score-error-def}.
\end{theorem}

Theorem~\ref{thm:sampling} establishes that, when the signal follows the Bernoulli--Gaussian prior, the likelihood-guided annealed Langevin sampler, followed by an endpoint draw, approximates the posterior in total variation up to an order of
$$\varepsilon +\sqrt{\Esc}$$
using $\widetilde{O}\left( \frac{n^6}{ms^2 \sigma^2 \varepsilon^6} \right)$ iterations when $0\leq \sigma<1$,  as soon as
the sample complexity $m$ is on the order of $\widetilde{O}\left( s+\log (1/\varepsilon) \right)$ for random matrix ensembles (including sub-Gaussian sensing matrices or randomly subsampled bases), and on the order of $\widetilde{O}\left( s+\log (1/\varepsilon) \right)^2$ for deterministic frames, up to logarithmic factors.
Furthermore, the performance guarantee degenerates gracefully as the score estimation error increases, analogous to that in unconditional diffusion guarantees \citep[Theorem~2]{chen2023sampling}.

To the best of our knowledge, Theorem~\ref{thm:sampling} provides the first polynomial-time posterior sampling algorithm at the near-optimal sample complexity $\widetilde{O}(s)$ using random sensing ensembles,   improving upon the sample complexities of several existing posterior sampling algorithms \citep{kumar2025spike,KSTZ26} that are not based on diffusion models.

\begin{remark}
Compared to Theorem~\ref{thm:application}, Theorem~\ref{thm:sampling} holds for every \(\sigma>0\), without the prior sparsity restriction \eqref{eq:param} or the signal conditions \eqref{eq:signal-admission}. The early stopping level in \eqref{eq:sampling-resources} must be small enough to control both the sampling error relative to \(\Pi_\tau\) and the difference between the endpoint distribution and the desired posterior.
\end{remark}

\section{Proof outline}\label{sec:proof-outline}

We provide an outline of our analysis. The central machinery is an {\em annealed transfer inequality}, detailed in Section~\ref{sec:ov-transfer}, which bounds the distance between the laws of the output of the sampler and the terminal target \(\Pi_\tau\) in \eqref{eq:target} under certain weak variance inequality that can be certified for the sensing ensembles of interest. We then move onto establishing Theorem~\ref{thm:application} in Section~\ref{sec:ov-recovery} and Theorem~\ref{thm:sampling} in Section~\ref{sec:ov-sampling}, by studying the relations of the terminal target \(\Pi_\tau\) with signal reconstruction and posterior sampling.

We also introduce some notation used throughout the proof. For a probability measure \(\mu\), write \(\E_\mu f=\mu f\coloneqq\int f\,d\mu\).
For a real square-integrable function \(f\) and a bounded function \(g\), respectively, define
\[
 \Var_\mu f\coloneqq\mu(f-\mu f)^2=\mu f^2-(\mu f)^2,
 \qquad \osc(g)\coloneqq\sup g-\inf g.
\]
A Markov kernel \(K(x,\cdot)\) is a probability measure for each \(x\), measurable in \(x\) on every measurable set. We let
\[
 Kf(x)\coloneqq\int f(z)K(x,dz),\qquad
 (\mu K)(\cdot)\coloneqq\int K(x,\cdot)\mu(dx).
\]
For probability measures
\(\mu_1,\mu_2\) with densities \({p},{q}\) with respect to a common dominating
measure \({\lambda}\), the \emph{squared Hellinger distance} is
\begin{equation}\label{eq:Helliger}
 \Hel(\mu_1,\mu_2)\coloneqq\int\bigl(\sqrt{{p}}-\sqrt{{q}}\bigr)^2\,d{\lambda}
 =2-2\int\sqrt{{p}\,{q}}\,d{\lambda} .
\end{equation}

\paragraph{Latent representation of the tilted target.}\label{sec:latent}\label{sec:evidence}
To compare consecutive tilted targets, we express each \(\Pi_t\) in terms of a latent sparse vector and a Gaussian conditional law. For \(t>0\), consider the auxiliary model
\[
 Z\sim p_\star,\qquad X_t=Z+\sqrt t\,\xi,\qquad
 \widetilde Y_t=AX_t+\sqrt{a_t}\,\xi',
\]
where \(Z\), \(\xi\sim\Gauss(0,I_n)\), and \(\xi'\sim\Gauss(0,I_m)\) are independent. Since \(X_t\) has density \(p_t\), it follows that \(\mathcal L(X_t\mid\widetilde Y_t=y)=\Pi_t\) in \eqref{eq:target}. In terms of \(Z\), the same observation takes the form
\[
 \widetilde Y_t=AZ+w_t,\qquad
\]
where \(w_t\sim\Gauss(0,\Sigma_t)\) is independent of \(Z\) with noise covariance
$$  \Sigma_t\coloneqq a_tI_m+tAA^\top. $$
Define the \emph{latent law}
\begin{equation}\label{eq:latent-def}
 \lat_t^{A,y}\coloneqq\mathcal L\bigl(Z\bigm|\widetilde Y_t=y\bigr),
\end{equation}
and we drop $A$ and $y$ and write \(\lat_t\) when there is no ambiguity. The \emph{forward kernel} is defined as the conditional law that maps the latent vector \(Z\) to \(X_t\),
\[
 \mathcal F_t^{A,y}(z,\cdot)\coloneqq\mathcal L\bigl(X_t\bigm|Z=z,\ \widetilde Y_t=y\bigr),
\]
which we similarly write as \(\mathcal F_t\). Conditioning on \(Z\) gives the mixture identity
\begin{equation}\label{eq:endpoint-identity}
 \lat_t\mathcal F_t=\Pi_t.
\end{equation}

\subsection{The annealed transfer inequality}\label{sec:ov-transfer}

Given a fixed instance \((A,y)\) following \eqref{eqn:inv}, we define the conditional law of the output of the sampler \eqref{eq:algorithm} as
\[
\rho_N\coloneqq\mathcal L(\Xh_N\mid A,y),
\]
and let \(\mu_{\sf init}\) be the law of
\(\Xh_0\) in \eqref{eq:init} given the data pair \((A,y)\).
By \eqref{eq:target-drift}, the drift of the update
\eqref{eq:algorithm} at \(x\) is
\[ \nabla\log\Pi_{t_j}(x)+\Delta_j(x) , \qquad \mbox{where}\quad  \Delta_j(x)\coloneqq\sh_{t_j}(x)-\nabla\log p_{t_j}(x)\]
is the error of the score estimate. Since the vectors \(\xi_j\) are independent of \((A,y)\) and of \(\Xh_0\), each update, conditioned on \((A,y)\), is a draw from
\begin{equation}\label{eq:ov-kernel}
 \Kraw_j(x,\cdot)\coloneqq\Gauss\left(x+h_j\left[\nabla\log\Pi_{t_j}(x)+\Delta_j(x) \right],\, 2h_jI_n \right),
\end{equation}
a Langevin step toward \(\Pi_{t_j}\) with the additional drift \(\Delta_j\). We can then write the conditional law of the sampler output as
\begin{equation}\label{eq:sampler_law}
\rho_N=\mu_{\sf init}\Kraw_0\cdots\Kraw_{N-1}.
\end{equation}

The annealed transfer inequality, which bounds the difference between \(\rho_N\) and \(\Pi_\tau\) at fixed instances by a sum of error terms, requires certain weak variance inequality for  the tilted targets, stated as follows.

\begin{definition}[Weak variance inequality]\label{def:ov-wvi}
A probability measure \(\gamma\) on \(\R^n\) satisfies a \emph{weak variance inequality} with
constant \(V\ge0\) and defect \(\delta\ge0\) if every bounded smooth \(g\) with bounded gradient
satisfies
\begin{equation}\label{eq:ov-wvi}
 \Var_\gamma g\le V\,\gamma\norm{\nabla g}^2+\delta\osc(g)^2.
\end{equation}
\end{definition}
The weak variance inequality \eqref{eq:ov-wvi} has the form of a weak Poincaré inequality
\citep{rockner2001weak} at a fixed defect level. We now state the annealed transfer inequality for a fixed instance, assuming the scores are smooth and the tilted targets satisfy the weak variance inequality.

\begin{proposition}[Annealed transfer]\label{prop:ov-abstract}
At each level \(\{t_j\}_{(j<N}\), suppose that \(\nabla\log p_{t_j}\) is \(L_j\)-Lipschitz and that \(\Pi_{t_j}\) satisfies \eqref{eq:ov-wvi} with constant \(V_j\) and defect \(\delta_j\). Define the movement of the latent law and of the Gaussian conditionals by
\[
 u_j=\Hel(\lat_{t_j},\lat_{t_{j+1}}),\qquad
 b_j=\int\Hel\bigl(\mathcal F_{t_j}(z,\cdot),\mathcal F_{t_{j+1}}(z,\cdot)\bigr)\,\lat_{t_j}(dz).
\]
For every measurable \(g:\R^n\to[0,1]\) and \(0<\lambda\le1\),
\begin{equation}\label{eq:ov-transfer}
 \rho_Ng\le(1+\lambda)\Pi_\tau g+C\mathcal E/\lambda+C\mathcal E_{\sf def},
\end{equation}
where
\begin{subequations}\label{eq:ov-transfer-error}
\begin{align}
 \mathcal E={}&\Hel(\Pi_T,\mu_{\sf init})+
 \sum_{j<N}h_j\Pi_{t_j}\|\Delta_j\|^2+
 \sum_{j<N}h_j^2\left[nL_j^2+\frac{\|A^\top A\|_{\rm F}^2}{a_{t_j}^2}\right]
 +\sum_{j<N}\left[\left(1+\frac{V_j}{h_j}\right)u_j+
                       \left(1+\frac{1}{\gamma}\right)b_j\right],\\
 \mathcal E_{\sf def}={}&\sum_{j<N}\sqrt{\delta_j u_j}.
\end{align}
\end{subequations}
\end{proposition}

Appendix~\ref{sec:finite-abstract} proves Proposition~\ref{prop:ov-abstract} by comparing each Euler step with a kernel that preserves the current target. The error terms in \eqref{eq:ov-transfer-error} have the following interpretations.
\begin{itemize}
\item \emph{Initialization error.} The squared Hellinger distance \(\Hel(\Pi_T,\mu_{\sf init})\) measures the mismatch between  the first target $\Pi_T$ and the initial law $\mu_{\sf init}$, which in turn measures the mismatch between \(p_T\) and \(\cN(0,TI_n)\), a term that diminishes quickly as $T$ becomes large.
\item \emph{Score estimation error.} The term \(\sum_{j<N}h_j\,\Pi_{t_j}\norm{\Delta_j}^2\) accumulates the mean squared error of the estimated prior score under each tilted target, weighted by the step size at that level.
\item \emph{Discretization error.} Each Euler--Maruyama step contributes \(h_j^2\bigl[nL_j^2+\|A^\top A\|_{\rm F}^2/a_{t_j}^2\bigr]\) to the discretization error, where the two terms reflect the Lipschitz constant of the prior score and the squared Frobenius norm of the log-likelihood Hessian, respectively.
\item \emph{Movement of the latent law.} The term \(\sum_{j<N}(1+V_j/h_j)u_j\) accounts for the change in the law of the sparse vector \(Z\) between consecutive levels.
\item \emph{Movement of the Gaussian conditionals.} The term \(\sum_{j<N}(1+1/\gamma)b_j\) accounts for the change in the conditional law at fixed \(Z=z\), averaged over \(z\sim\lat_{t_j}\).
\item \emph{Weak variance defect.} The term \(\mathcal E_{\sf def}=\sum_{j<N}\sqrt{\delta_j u_j}\) couples the defect at each level to the movement of the latent law alone.
\end{itemize}

It then boils down to 1) establishing the desired properties of the tilted targets $\Pi_{t_j}$ to apply Proposition~\ref{prop:ov-abstract}, and 2) controlling the error terms in \eqref{eq:ov-transfer-error} when averaged over the measurement noise.

\paragraph{The matrix condition.} To apply Proposition~\ref{prop:ov-abstract}, we use a condition on \(A\) to establish the weak variance inequality and to bound the discretization and target movement errors. For \(t>0\), set
\begin{equation}\label{eq:Sigma_Bt}
 B_t\coloneqq I_n+G_t=I_n+A^\top\Sigma_t^{-1}A, \qquad \mbox{where}\quad G_t = A^\top\Sigma_t^{-1}A.
\end{equation}

We define the following matrix condition.
\begin{definition}[Matrix condition]\label{def:truth}
Fix \(1\le K\le 2m\) as some universal constant. The \emph{matrix event} \(\mathcal{A}:=\mathcal{A}(q)\) is the set of matrices \(A\) that satisfy
\begin{subequations} \label{eq:matrix-event}
\begin{align}
\min_i\|A_i\|^2 & \ge\tfrac12,   \\
\|A\|^2 &  \le K\frac{n}{m}, \label{eq:matrix-spectral}\\
 \left\|(B_t-\diag B_t)_{JJ}\right\| & \le\matTol\min_{i\in J}B_{t,ii}
 \qquad\text{for every }\tau\le t\le T\text{ and }J\subseteq[n]\text{ with }1\le|J|\le2q. \label{eq:block-condition}
 \end{align}
\end{subequations}
Here, \(A_i\) denotes the \(i\)-th column of \(A\).
\end{definition}
The first two conditions are controlling the size of the columns of $A$ to be not too small, and the spectral norm of $A$ to be not too large.
The last condition \eqref{eq:block-condition}, which we refer to as the block condition \eqref{eq:block-condition}, requires \(B_t\) to be close to its diagonal on every set of at most \(2q\) coordinates, uniformly over \(t\in[\tau,T]\). This is reminiscent of the celebrated restricted isometry property \citep{candes2005decoding,tropp2008conditioning}, but requires the matrix $B_t$ to behave like a well-conditioned diagonal matrix along the diffusion path $t\in [\tau, T]$.

For transfer accuracy \(0<\eta\le1\), set \(L_\eta=\log(en/(\eta\min\{1,\sigma\}))\). For a fixed signal \(x^\star\) with support \(S\), choose the support {\em cap} \(q\) and the matrix tolerance \(\matTol\) as
\begin{equation}\label{eq:proof-parameters}
 q\coloneqq\left\lceil14\bigl(1+s+|S|+\|x^\star\|^2\bigr)L_\eta\right\rceil,\qquad
 \matTol\coloneqq\frac{c_{\rm mat}}{L_\eta},
\end{equation}
where \(c_{\rm mat}\in(0,1/16)\) is a fixed universal constant. Since \(L_\eta\ge1\), these choices give
\begin{equation}\label{eq:cap-facts}
 \matTol\le c_{\rm mat}<\tfrac{1}{16},\qquad q\ge|S|+\|x^\star\|^2\ge|S|.
\end{equation}
Whether a measurement matrix $A$ satisfies the matrix condition depends on the sensing matrix ensemble. Luckily, this event holds with high probability under the sample size of the main theorems for the sensing ensembles of interest, and we defer the verification to Appendix~\ref{sec:matrix-proof}.

\paragraph{The averaged annealed transfer inequality.} The matrix condition allows us to certify the desired properties of the tilted targets $\{\Pi_{t_j}\}_{j <N}$ to apply Proposition~\ref{prop:ov-abstract}. Averaging over the observation noise \(\zeta\), we can bound the error terms and obtain the next lemma summarizing the resulting error in terms of the terminal level \(\tau\) and the score error \(\Esc(A,x^\star)\).

\begin{lemma}[Averaged annealed transfer inequality]\label{lem:ov-transfer}
Let \(0<\tau\le\eta\le1\). Use the cap and matrix tolerance at accuracy \(\eta\) from \eqref{eq:proof-parameters}, and suppose \(q\le m\) and \(A\in\mathcal A(q)\). Suppose
\begin{equation}\label{eq:ald-transfer-count}
 N\ge C\,\frac{n^4}{m\eta^3}
 \left(1+\frac{n}{s\sigma\max\{1,\sigma\}}\right)^2\log^6(en/\tau),
\end{equation}
where \(C\) is a universal constant.
Then for every measurable \(g:\R^n\to[0,1]\) and \(0<\lambda\le1\),
\begin{subequations}
\begin{align}
 \E_\zeta\rho_Ng & \le(1+\lambda)\,\E_\zeta\Pi_\tau g+\frac C\lambda\bigl(\eta+\Esc(A,x^\star)\bigr),  \label{eq:ov-averaged-eta} \\
 \E_\zeta\TV(\rho_N,\Pi_\tau) & \le C\bigl(\eta+\Esc(A,x^\star)\bigr)^{1/2}.  \label{eq:ov-averaged-tv}
\end{align}
\end{subequations}
\end{lemma}
The proof of this lemma can be found in Appendix~\ref{sec:transfer-overview}. Our main theorems will be established by making use of
\eqref{eq:ov-averaged-eta} and \eqref{eq:ov-averaged-tv}, respectively.

\subsection{Proof of Theorem~\ref{thm:application}}\label{sec:ov-recovery}
For the fixed signal \(x^\star\) with support \(S\), \(|S|\le k\), define the failure set
\begin{equation}\label{eq:ov-failure-set}
 \mathcal B(x^\star)\coloneqq\bigl\{x:\ \|x-x^\star\|>C_{\sf rad}\,\sigma\bigl(\sqrt{k}+\log(en/\varepsilon)\bigr)\bigr\},
\end{equation}
where \(C_{\sf rad}\) is some universal constant. We abbreviate this set as \(\mathcal B\) when \(\xs\) is fixed.

Take \(\eta=\varepsilon\) in \eqref{eq:proof-parameters} and write \(\mathcal A=\mathcal A(q)\). Since \(s\ge1\) and \(c\le1\), \eqref{eq:param} gives
$ s\sigma\le\frac{c}{\sqrt{\log(en/\varepsilon)}}\le1$.
Consequently, \(\sigma\le1\) and \(n/(s\sigma)\ge1\), so
\[
 \left(1+\frac{n}{s\sigma\max\{1,\sigma\}}\right)^2
 =\left(1+\frac{n}{s\sigma}\right)^2
 \le\frac{4n^2}{s^2\sigma^2}.
\]
Thus \eqref{eq:ald-recovery-count} implies the iteration bound in Lemma~\ref{lem:ov-transfer}. Also, Proposition~\ref{prop:ov-sensing} and \eqref{eq:sample-size} give \(q\le m\) and
\[
 \PP(\mathcal A^c)\le\varepsilon.
\]
For every fixed \(A\in\mathcal A\), apply Lemma~\ref{lem:ov-transfer} with \(g=\1_{\mathcal B}\) and \(\lambda=1\) to obtain
\begin{equation}\label{eq:ov-failure-assembly}
 \E_\zeta\rho_N(\mathcal B)
 \le2\,\E_\zeta\Pi_\tau(\mathcal B)+C\bigl(\varepsilon+\Esc(A,\xs)\bigr).
\end{equation}
Since \(\mathcal A\) depends only on \(A\), we can decompose the error and average
\eqref{eq:ov-failure-assembly} over the sensing matrix as
\begin{align}\label{eq:ov-failure-split}
 \PP(\Xh_N\in\mathcal B)
 &\le\PP(\mathcal A^c)+\E_A\bigl[\1_{\mathcal A}\,\E_\zeta\rho_N(\mathcal B)\bigr] \nonumber \\
 &\le\PP(\mathcal A^c)
 +2\,\E_A\bigl[\1_{\mathcal A}\,\E_\zeta\Pi_\tau(\mathcal B)\bigr]
 +C\bigl(\varepsilon+\Esc(\xs)\bigr),
\end{align}
where \(\E_A\Esc(A,\xs)=\Esc(\xs)\). The second term in \eqref{eq:ov-failure-split} is controlled by the following lemma, whose proof is given in Appendix~\ref{sec:recovery-overview}.
\begin{lemma}[Terminal accuracy]\label{prop:ov-terminal}
Let \(0<\varepsilon<1\) and \(1\le s\le n/2\), and suppose \eqref{eq:param} and \eqref{eq:signal-admission} hold for a fixed signal \(x^\star\) with support \(S\), \(|S|\le k\). Choose \(q\) and \(\matTol\) by \eqref{eq:proof-parameters} with \(\eta=\varepsilon\). Then, for every fixed \(A\in\mathcal A(q)\),
 \[
\E_\zeta\Pi_\tau(\mathcal B)\le C\varepsilon.
\]
Here \(\mathcal B=\mathcal B(x^\star)\) is defined in \eqref{eq:ov-failure-set} with a sufficiently large universal \(C_{\sf rad}\), and the expectation is over the observation noise \(\zeta\).
\end{lemma}

Finally, substituting these bounds into
\eqref{eq:ov-failure-split} gives, for some universal constant \(C_1\ge1\),
\begin{equation}\label{eq:ov-failure-claim}
 \PP(\Xh_N\in\mathcal B)\le C_1\bigl(\varepsilon+\Esc(\xs)\bigr).
\end{equation}
Apply the entire argument above with \(\varepsilon'=\varepsilon/C_1\) in place of \(\varepsilon\) gives the desired theorem.

\subsection{Proof of Theorem~\ref{thm:sampling}}\label{sec:ov-sampling}

We begin by noting that the endpoint draw has conditional law
\[
\mathcal L(\widehat Z\mid A,y)=\rho_N\mathcal D_\tau, \qquad \mbox{where}\quad \rho_N=\mathcal L(\Xh_N\mid A,y).
\]
Denoting the true posterior \(p_\star^{A,y}\coloneqq p_\star(\cdot\mid y)\), we can bound that, for every \(A\) and \(y\),
\begin{align}
 \TV(\rho_N\mathcal D_\tau,\, p_\star^{A,y})
 &\le\TV(\rho_N\mathcal D_\tau,\,  \Pi_\tau\mathcal D_\tau)
 +\TV(\Pi_\tau\mathcal D_\tau,\,  p_\star^{A,y}) \nonumber\\
 &\le\TV(\rho_N,\, \Pi_\tau)+\TV(\Pi_\tau\mathcal D_\tau, \, p_\star^{A,y}).\label{eq:ov-sampling-triangle}
\end{align}
Here, the first line follows the triangle inequality, and the second line follows the contraction of total variation under \(\mathcal D_\tau\).
The two terms in the resulting bound \eqref{eq:ov-sampling-triangle} measure the sampler error relative to \(\Pi_\tau\), and the discrepancy between
the denoised terminal target \(\Pi_\tau\mathcal D_\tau\) and the true posterior $p_\star^{A,y}$, respectively.

Define the event
\begin{equation}\label{eq:ov-G-event}
\mathcal G\coloneqq\bigl\{ q \le m, \; A\in\mathcal A(q) \bigr\}.
\end{equation}
Since total variation is at most one, we can split \eqref{eq:ov-sampling-triangle} according to \(\mathcal G\) and bound
\begin{equation}\label{eq:ov-sampling-total}
 \E\,\TV(\rho_N\mathcal D_\tau,\, p_\star^{A,y})
\le\PP(\mathcal G^c) +\E\bigl[\1_{\mathcal G}\,\TV(\rho_N,\, \Pi_\tau)\bigr] +\E\bigl[\1_{\mathcal G}\,\TV(\Pi_\tau\mathcal D_\tau, \, p_\star^{A,y})\bigr].
\end{equation}
Setting \(\eta=\varepsilon^2\), the stopping rule \eqref{eq:sampling-resources} ensures \(\tau\le\eta\), and \eqref{eq:ald-sampling-count} is precisely the iteration condition \eqref{eq:ald-transfer-count}.
Applying Lemma~\ref{lem:ov-transfer} on \(\mathcal G\)  and then Jensen's inequality yields
\[
 \E\bigl[\1_{\mathcal G}\TV(\rho_N,\Pi_\tau)\bigr]
 \le C\,\E_{A,\Xs}\!\left[\1_{\mathcal G}\sqrt{\varepsilon^2+\Esc(A,\Xs)}\right]
 \le C\bigl(\varepsilon+\sqrt{\Esc}\bigr).
\]
 Additionally, under the measurement bound \eqref{eq:sampling-sample-size}, Proposition~\ref{prop:ov-sensing}  gives
\[
 \PP(\mathcal G^c)\le2\varepsilon.
\]

Finally, we bound the last term in \eqref{eq:ov-sampling-total}, which is controlled by the lemma below.

\begin{lemma}[Comparison of terminal target]\label{lem:ov-covariance}
Let \(\Xs\sim\mathsf{BG}(s,n)\), let \(A\in\R^{m\times n}\), with \(m\le n\), be independent
of \(\Xs\), and let \(y=A\Xs+\sigma\zeta\). Then, for every \(\tau>0\),
\[
 \E\bigl[\1_{\{\|A\|^2\le2n\}}\,\TV(\Pi_\tau\mathcal D_\tau,\, p_\star^{A,y})\bigr]
 \le\frac{4n^2\tau}{\sigma^2}\Bigl(1+\frac s{\sigma^2}\Bigr).
\]
\end{lemma}

The proof of this lemma can be found in Appendix~\ref{sec:covariance-proof}.
On \(\mathcal G\), \eqref{eq:matrix-spectral} gives \(\|A\|^2\le Kn/m\le2n\). Hence
 \begin{align*}
 \E\bigl[\1_{\mathcal G}\,\TV(\Pi_\tau\mathcal D_\tau,p_\star^{A,y})\bigr]
 &\le\E\bigl[\1_{\{\|A\|^2\le2n\}}\,\TV(\Pi_\tau\mathcal D_\tau,p_\star^{A,y})\bigr]\le\frac{4n^2\tau}{\sigma^2}\Bigl(1+\frac s{\sigma^2}\Bigr)\le\varepsilon,
 \end{align*}
where the last inequality follows from \eqref{eq:sampling-resources} for sufficiently small \(c\le1/4\).
Substituting the three bounds into \eqref{eq:ov-sampling-total} gives
\begin{equation}\label{eq:ov-sampling-claim}
 \E\,\TV\bigl(\mathcal L(\widehat Z\mid A,y),p_\star^{A,y}\bigr)
 \lesssim\varepsilon+\sqrt{\Esc},
\end{equation}
which proves Theorem~\ref{thm:sampling}.

\section{Discussion}
\label{sec:discussion}
We establish that the likelihood-guided annealed Langevin sampler \eqref{eq:algorithm} provides statistically and computationally efficient sparse recovery guarantees under a Bernoulli--Gaussian prior, when the prior information is provided implicitly via a diffusion model. This provides a first step toward a benign mathematical characterization of diffusion-based inverse solvers, when the prior model exhibits non-log-concavity.  The analysis highlights several directions for future work.
\begin{itemize}
    \item {\em Improving the computational efficiency.} This work provides a polynomial-time computational complexity whose order is rather pessimistic and not optimized. It will be of great interest to further sharpen the computational guarantee and possibly simplify the analysis framework.

    \item {\em Beyond sparse priors.}
    The Bernoulli-Gaussian prior provides a fruitful test ground for establishing the finite-time performance of diffusion-based inverse solvers. However, many applications involve richer priors involving low-dimensional structures. Extending the result to broader classes of low-dimensional priors would help clarify the extent to which diffusion-based solvers adapt to intrinsic low-dimensional structures inherent in real-world data.

    \item {\em Other types of solvers.}
    In practice, diffusion-based inverse problems are often solved using a wide range of tools, including predictor-corrector methods, proximal or projection-based updates, and plug-and-play samplers. Developing a unified theory that covers such algorithms would be extremely valuable.

\end{itemize}

\appendix

\section{Preliminaries}\label{sec:preliminaries}
This appendix collects the general tools and the latent representation of the tilted targets used in the proofs.

\subsection{Technical background}\label{sec:general-preliminaries}

\paragraph{Probability distances.}\label{par:probability-distances}
For the squared Hellinger distance defined in \eqref{eq:Helliger}, the next lemma collects some standard comparisons, whose proofs can be found in \citet{polyanskiy2024information}.

\begin{lemma}\citep{polyanskiy2024information}\label{lem:c-hellinger}
Let \(\nu\) be a \(\sigma\)-finite measure, and let \({p}(x,z),{q}(x,z)\ge0\) be functions such that
\[
 {p}(x)\coloneqq\int{p}(x,z)\,\nu(dz),\qquad
 {q}(x)\coloneqq\int{q}(x,z)\,\nu(dz)
\]
are probability densities.
\begin{enumerate}
\item The mixtures satisfy
\begin{equation}\label{eq:hel-mix}
 \Hel({p},{q})\le\iint\bigl(\sqrt{{p}(x,z)}-\sqrt{{q}(x,z)}\bigr)^2\,\nu(dz)\,dx.
\end{equation}
\item If \({p}(x,z)={w}(z){p}_z(x)\) and \({q}(x,z)={w}(z){q}_z(x)\) for probability densities \({p}_z,{q}_z\) and a common mixing density \({w}\), then the right side of \eqref{eq:hel-mix} equals \(\int\Hel({p}_z,{q}_z)\,{w}(z)\,\nu(dz)\).
\item For \(\Sigma\succ0\) and vectors \({m}_1,{m}_2\), with \(d_\Sigma\coloneqq({m}_1-{m}_2)^\top\Sigma^{-1}({m}_1-{m}_2)\),
\begin{equation}\label{eq:c-gauss-hel}
 \Hel\bigl(\Gauss({m}_1,\Sigma),\Gauss({m}_2,\Sigma)\bigr)=2-2e^{-d_\Sigma/8}\le d_\Sigma/4.
\end{equation}
\end{enumerate}
\end{lemma}

For \(\mu\ll\nu\), the relative entropy is
\[
 \mathsf{KL}(\mu\|\nu)\coloneqq\int\log\frac{d\mu}{d\nu}\,d\mu.
\]
It is \(+\infty\) when \(\mu\) is not absolutely continuous with respect to \(\nu\). For probability measures \(\mu_1,\mu_2\) with densities \(p,q\) with respect to a common dominating measure \(\lambda\), the total-variation distance is defined as
\[
\TV(\mu_1,\mu_2)\coloneqq\sup_E|\mu_1(E)-\mu_2(E)|=\frac12\int|{p}-{q}|\,d{\lambda}.
\]
We have $ \TV^2(\mu_1,\mu_2) \leq \Hel(\mu_1,\mu_2)$. For a Markov kernel \(K\), let the conditional variance after one transition be
\[
\Var^K g=K(g^2)-(Kg)^2.
\]
A Markov kernel does not increase total variation: \(\TV(\mu K,\nu K)\le\TV(\mu,\nu)\).

\paragraph{Poincar\'e constants.} A probability \(\mu\) on \(\R^{d}\) has \emph{Poincar\'e constant} at most \({K}\) if
\[
\Var_\mu f\le{K}\mu\norm{\nabla f}^2
\]
for every smooth \(f\) with compact support. By standard approximation using cutoff and mollification, the above inequality also holds for every \(C^1\) function \(f\) with \(\mu f^2<\infty\) and \(\mu\norm{\nabla f}^2<\infty\), in particular for bounded \(C^1\) functions with bounded gradient.

The following lemma collects some useful standard properties of Poincar\'e inequalities; see \citet{chafai2004entropies} for details. In particular, Item~5 is the strongly log-concave case of the Brascamp--Lieb inequality \citep{brascamp1976extensions}, in the form of \citet[Proposition~2.1]{bobkov2000brunn}.

\begin{lemma}[Properties for Poincar\'e constants]\label{lem:poincare-rules}
The following statements are true.
\begin{enumerate}
\item A product measure has Poincar\'e constant at most the maximum of those of its factors.
\item The image of a law under an affine map \(x\mapsto Tx+x_0\) has Poincar\'e constant at most \(\norm{T}^2\) times that of the law.
\item The Gaussian \(\Gauss(m,\Sigma)\) has Poincar\'e constant at most \(\norm\Sigma\), including for singular \(\Sigma\).
\item The law of a sum of independent random vectors has Poincar\'e constant at most the sum of theirs.
\item If \(\mu(dx)\propto e^{-\Psi(x)}dx\) for a smooth \(\Psi\) with \(\nabla^2\Psi\succeq aI\) for some \(a>0\), then \(\mu\) has Poincar\'e constant at most \(1/a\).
\end{enumerate}
\end{lemma}

\paragraph{Weak interaction.} For a probability \(\mu\) on \(\R^{d}\) whose Laplace transform \(\mathcal M(\eta)\coloneqq\int e^{\eta^\top w}\mu(dw)\) is finite for every \(\eta\in\R^{d}\), define its normalized linear tilt by
\begin{equation} \label{eq:linear_tilt}
\mu^{\eta}(dw)\coloneqq\mathcal M(\eta)^{-1}e^{\eta^\top w}\mu(dw).
\end{equation}
The next lemma controls the Poincaré constant of a quadratic tilt when the interaction is weak relative to the covariance bound. Its proof adapts the strategy of Bauerschmidt and Bodineau \citep[Section~1]{bauerschmidt2019very} to a variance inequality, using a representation as a mixture of linear tilts over a strongly log-concave field.

\begin{lemma}[Weak interaction]\label{lem:b-interaction}
Let \(\mu\) be a probability on \(\R^{d}\) with Laplace transform $\mathcal M(\eta)$ finite everywhere. Suppose that every linear tilt \(\mu^\eta\) has covariance at most \(v_0I\) and Poincaré constant at most \(K_0\). Let \(H\) be symmetric with \(H\succeq0\) and \(\bar v\coloneqq v_0\norm H<1\). Then
\begin{enumerate}
\item \(\int e^{w^\top Hw/2}\mu(dw)<\infty\), so that \(\mu_H(dw)\propto e^{w^\top Hw/2}\mu(dw)\) is a probability,
\item \(\mu_H\) has Poincaré constant at most \(K_0/(1-\bar v)\).
\end{enumerate}
\end{lemma}

\begin{proof}[Proof of Lemma~\ref{lem:b-interaction}]\label{sec:proof-b-interaction}
We represent \(\mu_H\) as a mixture of linear tilts of \(\mu\). The covariance hypothesis controls the mixing field, while the Poincaré hypothesis controls the conditional laws.

\paragraph{Proof of item 1.} Define
\[
 \Psi(g)\coloneqq\frac{\norm g^2}{2}-\log\mathcal M(H^{1/2}g).
\]
Since finiteness of \(\mathcal M\) everywhere permits differentiation under the integral and gives
\(\nabla^2\log\mathcal M(\eta)=\Cov(\mu^\eta)\), it leads to
\[
 \nabla^2\Psi(g)=I-H^{1/2}\Cov(\mu^{H^{1/2}g})H^{1/2}
 \succeq(1-\bar v)I.
 \]
Since \(\bar v<1\),
\[
 \Psi(g)\ge\Psi(0)+\nabla\Psi(0)^\top g+\frac{1-\bar v}{2}\norm g^2,
\]
so \(e^{-\Psi}\) is integrable. Tonelli's theorem and the Gaussian integral in \(g\) now give
\[
 \int e^{-\Psi(g)}\,dg
 =\iint e^{-\norm g^2/2+g^\top H^{1/2}w}\,dg\,\mu(dw)
 =(2\pi)^{d/2}\int e^{w^\top Hw/2}\mu(dw)<\infty,
\]
proving item~1.

\paragraph{Proof of item 2.} Let \((g,w)\) follow the law of the normalized joint measure \(e^{-\norm g^2/2+g^\top H^{1/2}w}\,dg\,\mu(dw)\). Then
\[
 w\sim\mu_H,\qquad
 g\sim\frac{e^{-\Psi(g)}\,dg}{\int e^{-\Psi}},\qquad
 w\mid g\sim\mu^\eta,\quad \mbox{with}\; \eta=H^{1/2}g.
\]
By Lemma~\ref{lem:poincare-rules}(5), the law of \(g\) has Poincaré constant at most \((1-\bar v)^{-1}\).

Let \(f\) be smooth with compact support. Given \(g\), the law of \(w\) is \(\mu^\eta\), so
\[
 \Var(f(w)\mid g)\le K_0\E[\norm{\nabla f(w)}^2\mid g].
\]
For the conditional mean \(\bar f(g)\coloneqq\E[f(w)\mid g]\), differentiation of the linear tilt gives
\[
 \nabla\bar f(g)=H^{1/2}\Cov(w,f(w)\mid g).
\]
For every unit vector \(e\), Cauchy--Schwarz and the covariance hypothesis imply
\[
 |e^\top\nabla\bar f(g)|^2
 \le\Var(f(w)\mid g)\,
 e^\top H^{1/2}\Cov(w\mid g)H^{1/2}e
 \le\bar v\Var(f(w)\mid g).
\]
Taking the supremum over unit vectors gives
\[
 \norm{\nabla\bar f(g)}^2\le\bar v\Var(f(w)\mid g).
\]
In particular, \(\bar f\) is bounded and smooth with bounded gradient, so the Poincaré inequality for \(g\) applies. The conditional variance decomposition therefore yields
\[
 \begin{split}
 \Var f(w)
 &=\E\Var(f(w)\mid g)+\Var\bar f(g)\\
 &\le\E\Var(f(w)\mid g)+\frac{\E\norm{\nabla\bar f(g)}^2}{1-\bar v}\\
 &\le\frac{\E\Var(f(w)\mid g)}{1-\bar v}
 \le\frac{K_0}{1-\bar v}\E\norm{\nabla f(w)}^2.
 \end{split}
\]
Since \(w\sim\mu_H\), this proves item~2.
\end{proof}
\subsection{Gaussian mixture representation of the tilted targets} \label{sec:direct}

We first provide the closed-form expression of the forward kernel in Section~\ref{sec:proof-outline}. To identify the conditional mean and covariance, let
\[
 G_t\coloneqq A^\top\Sigma_t^{-1}A,\qquad
 b_t\coloneqq A^\top\Sigma_t^{-1}y,\qquad M_t\coloneqq I_n-tG_t.
\]
Using \(M_t=(I_n+(t/a_t)A^\top A)^{-1}\) and \(M_tA^\top=a_tA^\top\Sigma_t^{-1}\), it follows that
\begin{equation}\label{eq:forward-kernel}
 \mathcal F_t^{A,y}(z,\cdot)=\Gauss\bigl(M_tz+tb_t,\,tM_t\bigr).
\end{equation}
In particular, \(0\prec M_t\preceq I_n\), so the conditional covariance is bounded by \(tI_n\).

We next decompose the latent law \(\Pi_t\) according to the support of \(Z\). For a fixed level $t$, it follows from Bayes' formula that
\begin{equation}\label{eq:latent}
 \lat_t(dz)\propto e^{-(y-Az)^\top\Sigma_t^{-1}(y-Az)/2}p_\star(dz)
 \propto e^{(b_t)^\top z-z^\top G_tz/2}p_\star(dz).
\end{equation}
Define the \emph{latent support} \(\mathsf S_t\coloneqq\supp z\) under \(\lat_t\). For \(W\subseteq[n]\), write \(\lat_t^W\coloneqq\lat_t(\cdot\mid \mathsf S_t=W)\). Combining \eqref{eq:endpoint-identity} with the support decomposition gives
\begin{equation}\label{eq:target-components}
 \Pi_t=\sum_{W\subseteq[n]}\lat_t\{\mathsf S_t=W\}\,\lat_t^W\mathcal F_t.
\end{equation}
Let \(B_{WW}\coloneqq(B_t)_{WW}\), \(G_{WW}\coloneqq(G_t)_{WW}\) and \(b_W\coloneqq(b_t)_W\), and \(\omega\coloneqq s/(n-s)\) be the prior odds of a coordinate, then by some calculation, we have
\(\lat_t\{\mathsf S_t=W\}=w_W/\sum_{V\subseteq[n]}w_V>0\) with
\begin{equation}\label{eq:weights}
 w_W\coloneqq\omega^{|W|}\det(B_{WW})^{-1/2}
 e^{(b_W)^\top B_{WW}^{-1}b_W/2},\qquad w_\varnothing=1,
\end{equation}
and, under \(\lat_t^W\),
\begin{equation}\label{eq:latent-support-law}
 z_W\sim\Gauss\bigl(B_{WW}^{-1}b_W,B_{WW}^{-1}\bigr),
 \qquad z_{W^c}=0.
\end{equation}
Therefore, each component \(\lat_t^W\mathcal F_t\) in \eqref{eq:target-components} is Gaussian by \eqref{eq:latent-support-law} and \eqref{eq:forward-kernel}. This representation permits separate control of the support probabilities and of the distributions within the components.

\section{Annealed transfer inequalities}\label{sec:finite}\label{sec:analysis}\label{sec:overview}
This appendix first proves Proposition~\ref{prop:ov-abstract} in Section~\ref{sec:finite-abstract}, then derives the weak variance inequality from the matrix condition in Section~\ref{sec:mixture}, and finally proves Lemma~\ref{lem:ov-transfer} in Section~\ref{sec:finite-model}.

\subsection{Proof of Proposition~\ref{prop:ov-abstract}}\label{sec:finite-abstract}

We start by introducing an invariant kernel that preserves the current target.

\paragraph{Step 1: comparison kernel and variance removal.}
Fix \(j<N\) and let \(\varphi_{h_j}\) be the density of \(\Gauss(0,h_j I_n)\). Define
\[
 r_{u}(dx')\coloneqq
 \frac{\varphi_{h_j}(u-x')\,\Pi_{t_j}(dx')}{\int\varphi_{h_j}(u-x'')\,\Pi_{t_j}(dx'')}.
\]
The comparison kernel \(P_j(x,\cdot)\) is defined as the law of \(X'\) obtained by drawing \(\widetilde X\sim\Gauss(x,h_j I_n)\) and then \(X'\sim r_{\widetilde X}\). This is one step of the proximal sampler \citep{lee2021structured,chen2022proximal}. To establish invariance, let \(X\sim\Pi_{t_j}\), and \(\widetilde X\coloneqq X+\sqrt{h_j}\,\xi\) with \(\xi\) standard Gaussian and independent of \(X\), and draw \(X'\sim r_{\widetilde X}\) given \((X,\widetilde X)\). By Bayes' formula, \(r_{\widetilde X}\) is the conditional law of \(X\) given \(\widetilde X\). Thus \(X\) and \(X'\) are conditionally independent and identically distributed given \(\widetilde X\), and in particular have the same marginal law \(\Pi_{t_j}\). Since \(X'\) given \(X\) has law \(P_j(X,\cdot)\), we obtain
\[
 \Pi_{t_j}P_j=\mathcal L(X')=\Pi_{t_j}.
\]

The weak variance inequality for the full target, following the Poincar\'e-based analysis of \citet[Theorem~4]{chen2022proximal}, and the Gaussian Poincar\'e inequality for each forward conditional give the following two bounds. The proof is in Appendix~\ref{sec:proof-c-return}.

\begin{lemma}[Variance removed by the comparison kernel]\label{lem:c-return}
Under the hypotheses of Proposition~\ref{prop:ov-abstract}, fix \(j<N\) and a measurable \(g:\R^n\to[0,1]\). We have
\begin{align*}
 \Var_{\Pi_{t_j}}g &\le (1+V_j/h_j)\,\Pi_{t_j}\Var^{P_j}g+\delta_j,\\
 \lat_{t_j}\Var^{\mathcal F_{t_j}}g &\le 2(1+t_j/h_j)\,\Pi_{t_j}\Var^{P_j}g.
\end{align*}
\end{lemma}

\paragraph{Step 2: telescope the target expectations.} Let \(0<\lambda\le1\),
and define the quadratic transform \(\Phi_\lambda(u)=u+\lambda u^2\). The following Hellinger bounds compare expectations while retaining a negative variance term. The proof is given in Appendix~\ref{sec:proof-c-sqrt}.

\begin{lemma}[Square-root inequalities]\label{lem:c-sqrt}
Let \(\mu_1,\mu_2\) be probability measures on \(\R^n\) and \(g\) a bounded function. Then
\begin{equation}\label{eq:sqrt-test}
 |\mu_2g-\mu_1g|\le2\sqrt{\Var_{\mu_1}g\cdot\Hel(\mu_1,\mu_2)}+\osc(g)\Hel(\mu_1,\mu_2).
\end{equation}
If moreover \(0\le g\le1\) and \(0<\lambda\le1\), then
\begin{align}
 \Phi_\lambda(\mu_2g)& \le\mu_1\Phi_\lambda(g)-\tfrac\lambda2\Var_{\mu_1}g
 +\frac{21}\lambda\Hel(\mu_1,\mu_2),
 \label{eq:Phi}\\
 \mu_2g& \le\mu_1\Phi_\lambda(g)+\frac2\lambda\Hel(\mu_1,\mu_2).\label{eq:initphi}
\end{align}
\end{lemma}

Fix a measurable \(g:\R^n\to[0,1]\). Starting from \(g_N=g\), define \(g_j=\Kraw_jg_{j+1}\) for \(j=N-1,\ldots,0\). Then \(0\le g_j\le1\) and \(\rho_Ng=\mu_{\sf init}g_0\). For each \(x\), applying \eqref{eq:Phi} of Lemma~\ref{lem:c-sqrt} with \(\mu_1=P_j(x,\cdot)\), \(\mu_2=\Kraw_j(x,\cdot)\), and test function \(g_{j+1}\), integrating over \(x\sim\Pi_{t_j}\) and applying \(\Pi_{t_j}P_j=\Pi_{t_j}\) gives
\begin{equation}\label{eq:kernel-comparison}
 \Pi_{t_j}\Phi_\lambda(g_j)
 \le \Pi_{t_j}\Phi_\lambda(g_{j+1}) -\frac{\lambda}{2}\Pi_{t_j}\Var^{P_j}g_{j+1}+\frac{21}{\lambda}e_j,
\end{equation}
where
\[
 e_j\coloneqq\int\Hel\bigl(P_j(x,\cdot),\Kraw_j(x,\cdot)\bigr)\,\Pi_{t_j}(dx).
\]

Telescoping over \(j<N\) for \eqref{eq:kernel-comparison} gives
\begin{equation}\label{eq:transfer-telescope}
 \Pi_T\Phi_\lambda(g_0)
 \le \Pi_\tau\Phi_\lambda(g)
 +\sum_{j<N}\bigl|(\Pi_{t_{j+1}}-\Pi_{t_j})\Phi_\lambda(g_{j+1})\bigr|-\frac{\lambda}{2}\sum_{j<N}\Pi_{t_j}\Var^{P_j}g_{j+1}
 +\frac{21}{\lambda}\sum_{j<N}e_j,
\end{equation}
where we use \(t_0=T\), \(t_N=\tau\), and \(g_N=g\).

\paragraph{Step 3: bound the kernel discrepancy and target movement.} It therefore remains to bound \(e_j\) and \(\bigl|(\Pi_{t_{j+1}}-\Pi_{t_j})\Phi_\lambda(g_{j+1})\bigr|\).
The following lemma bounds \(e_j\) by the score estimation and discretization errors. Its proof is given in Appendix~\ref{sec:proof-c-euler}.

\begin{lemma}[Comparison with the update]\label{lem:c-euler}
Under the hypotheses of Proposition~\ref{prop:ov-abstract}, for each \(j<N\),
\begin{equation}\label{eq:euler-distance}
 e_j\le h_j\Pi_{t_j}\|\Delta_j\|^2
 +4h_j^2\left[nL_j^2+\frac{\|A^\top A\|_{\rm F}^2}{a_{t_j}^2}\right].
\end{equation}
\end{lemma}

We now move to bound \(\bigl|(\Pi_{t_{j+1}}-\Pi_{t_j})\Phi_\lambda(g_{j+1})\bigr|\). Fix \(j<N\) and let \(f=\Phi_\lambda(g_{j+1})\). To compare \(\Pi_{t_j}f\) with \(\Pi_{t_{j+1}}f\), we use the mixture identity \(\Pi_t=\lat_t\mathcal F_t\) from \eqref{eq:endpoint-identity} to bound the movement of the latent law and of the Gaussian conditionals separately. Recall that these movements are measured by
\[
 u_j=\Hel(\lat_{t_j},\lat_{t_{j+1}}),\qquad
 b_j=\int\Hel\bigl(\mathcal F_{t_j}(z,\cdot),\mathcal F_{t_{j+1}}(z,\cdot)\bigr)\,\lat_{t_j}(dz).
\]
Adding and subtracting \(\lat_{t_j}\mathcal F_{t_{j+1}}f\) gives
\[
 \bigl|(\Pi_{t_{j+1}}-\Pi_{t_j})f\bigr|
 \le \bigl|(\lat_{t_{j+1}}-\lat_{t_j})\mathcal F_{t_{j+1}}f\bigr|
      +\bigl|\lat_{t_j}(\mathcal F_{t_{j+1}}-\mathcal F_{t_j})f\bigr|.
\]
We bound these two terms separately. Since \(0\le g_{j+1}\le1\) and \(0<\lambda\le1\), we have \(\osc(f)\le2\), and \(\Phi_\lambda\) is \(3\)-Lipschitz on \([0,1]\).
\begin{itemize}
\item \emph{Movement of the latent law.}
Apply \eqref{eq:sqrt-test} of Lemma~\ref{lem:c-sqrt} with \(\mu_1=\lat_{t_j}\), \(\mu_2=\lat_{t_{j+1}}\), and test function \(z\mapsto\mathcal F_{t_{j+1}}f(z)\). Since Markov kernels do not increase oscillation, we have
\[
 \bigl|(\lat_{t_{j+1}}-\lat_{t_j})\mathcal F_{t_{j+1}}f\bigr|
 \le 2\sqrt{u_j\Var_{\lat_{t_j}}(\mathcal F_{t_{j+1}}f)}+2u_j.
\]
To control this variance, first replace \(\mathcal F_{t_{j+1}}\) by \(\mathcal F_{t_j}\). The inequality \(\TV^2\le\Hel\) and \(\osc(f)\le2\) give
\[
 \lat_{t_j}\bigl|(\mathcal F_{t_{j+1}}-\mathcal F_{t_j})f\bigr|^2\le4b_j.
\]
Using the law of total variance for \(\Pi_{t_j}=\lat_{t_j}\mathcal F_{t_j}\) and the \(3\)-Lipschitz property of \(\Phi_\lambda\), we obtain
\begin{align*}
 \Var_{\lat_{t_j}}(\mathcal F_{t_{j+1}}f)
 &\le 2\Var_{\lat_{t_j}}(\mathcal F_{t_j}f)
      +2\lat_{t_j}\bigl|(\mathcal F_{t_{j+1}}-\mathcal F_{t_j})f\bigr|^2\\
 &\le 2\Var_{\Pi_{t_j}}f+8b_j
 \le 18\Var_{\Pi_{t_j}}g_{j+1}+8b_j.
\end{align*}
Substituting the first inequality of Lemma~\ref{lem:c-return} and separating the square roots therefore yields
\begin{align*}
 \bigl|(\lat_{t_{j+1}}-\lat_{t_j})\mathcal F_{t_{j+1}}f\bigr|
 &\le 6\sqrt{2}\sqrt{(1+V_j/h_j)u_j\,\Pi_{t_j}\Var^{P_j}g_{j+1}}+6\sqrt{2}\sqrt{\delta_j u_j}+4\sqrt{2}\sqrt{u_jb_j}+2u_j.
\end{align*}

\item \emph{Movement of the Gaussian conditionals.}
The \(3\)-Lipschitz property of \(\Phi_\lambda\) and the second inequality of Lemma~\ref{lem:c-return} give
\[
 \lat_{t_j}\Var^{\mathcal F_{t_j}}f
 \le9\lat_{t_j}\Var^{\mathcal F_{t_j}}g_{j+1}
 \le18(1+t_j/h_j)\,\Pi_{t_j}\Var^{P_j}g_{j+1}.
\]
For each \(z\), apply \eqref{eq:sqrt-test} of Lemma~\ref{lem:c-sqrt} with \(\mu_1=\mathcal F_{t_j}(z,\cdot)\), \(\mu_2=\mathcal F_{t_{j+1}}(z,\cdot)\), and test function \(f\). Averaging over \(z\sim\lat_{t_j}\), using Cauchy--Schwarz and then the preceding variance bound, gives
\begin{align*}
 \bigl|\lat_{t_j}(\mathcal F_{t_{j+1}}-\mathcal F_{t_j})f\bigr|
 &\le2\sqrt{b_j\,\lat_{t_j}\Var^{\mathcal F_{t_j}}f}+2b_j\\
 &\le 6\sqrt{2}\sqrt{(1+t_j/h_j)b_j\,\Pi_{t_j}\Var^{P_j}g_{j+1}}+2b_j.
\end{align*}

\end{itemize}
Combining the two bounds, and applying Young's inequality to each term containing \(\sqrt{\Pi_{t_j}\Var^{P_j}g_{j+1}}\) by allocating \(\tfrac\lambda4\Pi_{t_j}\Var^{P_j}g_{j+1}\) to each, we obtain
\begin{equation}\label{eq:split-target-movement}
 \bigl|(\Pi_{t_{j+1}}-\Pi_{t_j})f\bigr|\le \frac{\lambda}{2}\Pi_{t_j}\Var^{P_j}g_{j+1}
 +\frac{77}\lambda\big[(1+V_j/h_j)u_j+(1+t_j/h_j)b_j\big]
 +6\sqrt{2}\sqrt{\delta_j u_j}.
\end{equation}

\paragraph{Step 4: finish up.} Substituting \eqref{eq:split-target-movement} into \eqref{eq:transfer-telescope} cancels the variance terms and gives
\[
 \Pi_T\Phi_\lambda(g_0)
 \le \Pi_\tau\Phi_\lambda(g)
 +\frac{C}{\lambda}\sum_{j<N}
 \big[e_j+(1+V_j/h_j)u_j+(1+t_j/h_j)b_j\big]
 +C\sum_{j<N}\sqrt{\delta_j u_j}
\]
for some universal constant $C>0$.
Finally, apply \eqref{eq:initphi} of Lemma~\ref{lem:c-sqrt} with \(\mu_1=\Pi_T\), \(\mu_2=\mu_{\sf init}\), and test function \(g_0\) to obtain
$$ \rho_Ng=\mu_{\sf init}g_0\le\Pi_T\Phi_\lambda(g_0)+(2/\lambda)\Hel(\Pi_T,\mu_{\sf init}).$$
Using \(\Phi_\lambda(g)\le(1+\lambda)g\) and substituting the bound \eqref{eq:euler-distance} on \(e_j\) proves \eqref{eq:ov-transfer} with the errors in \eqref{eq:ov-transfer-error}, where we plug in the relation $h_j = \gamma t_j$.

\subsubsection{Proof of Lemma~\ref{lem:c-return}}\label{sec:proof-c-return}

Fix \(j<N\). Since \(P_jg\) is a Gaussian convolution of a \([0,1]\)-valued function, it is smooth with bounded gradient and values in \([0,1]\). Applying the weak variance inequality \eqref{eq:ov-wvi} to \(P_jg\), with \(\osc(P_jg)\le1\), gives
\[
 \Var_{\Pi_{t_j}}(P_jg)\le V_j\,\Pi_{t_j}\|\nabla P_jg\|^2+\delta_j.
\]
It remains to bound the gradient term by the conditional variance \(\Var^{P_j}g\). Let \(X,\widetilde X,X'\) be constructed as in Step~1. For any unit vector \(e\), differentiating the Gaussian density gives
\begin{align*}
 e^\top\nabla P_jg(x)
 &=h_j^{-1}\int e^\top(u-x)\varphi_{h_j}(u-x)\,r_u g\,du\\
 &=h_j^{-1}\E\bigl[e^\top(\widetilde X-x)\{g(X')-P_jg(x)\}\bigm|X=x\bigr].
\end{align*}
Here differentiation is justified by \(0\le r_u g\le1\) and integrability of the Gaussian density's derivatives. The second equality uses \(\E[g(X')\mid X,\widetilde X]=r_{\widetilde X}g\) and \(\E[\widetilde X-x\mid X=x]=0\). Since \(\E[(e^\top(\widetilde X-x))^2\mid X=x]=h_j\), Cauchy--Schwarz yields
\[
 |e^\top\nabla P_jg(x)|^2
 \le h_j^{-1}\Var\bigl(g(X')\bigm|X=x\bigr)
 =h_j^{-1}\Var^{P_j}g(x).
\]
Taking the supremum over unit vectors \(e\) and averaging under \(\Pi_{t_j}\) gives
\[
\Pi_{t_j}\|\nabla P_jg\|^2\le h_j^{-1}\Pi_{t_j}\Var^{P_j}g.
\]
Since \(\E[g(X')\mid X]=P_jg(X)\) and \(X,X'\sim\Pi_{t_j}\), the law of total variance gives
\begin{equation}\label{eq:proximal}
 \Pi_{t_j}\Var^{P_j}g=\E\Var\bigl(g(X')\bigm|X\bigr)=\Var_{\Pi_{t_j}} g-\Var_{\Pi_{t_j}}(P_jg).
\end{equation}
Combining \eqref{eq:proximal} with the weak variance inequality and the gradient bound above, we conclude that
\begin{align*}
 \Var_{\Pi_{t_j}}g
 &=\Pi_{t_j}\Var^{P_j}g+\Var_{\Pi_{t_j}}(P_jg)\le(1+V_j/h_j)\Pi_{t_j}\Var^{P_j}g+\delta_j.
\end{align*}

For the second inequality, use \(P_jg(X)=\E[r_{\widetilde X}g\mid X]\) and conditional Jensen's inequality to obtain
\begin{align*}
 \Pi_{t_j}(g-P_jg)^2
 &\le \E\bigl[(g(X)-r_{\widetilde X}g)^2\bigr]
 =\E\Var\bigl(g(X')\bigm|X,\widetilde X\bigr)\\*
 &\le \E\Var\bigl(g(X')\bigm|X\bigr)
 =\Pi_{t_j}\Var^{P_j}g.
\end{align*}
Here, the first equality uses the fact that \(X\) given \(\widetilde X\) and \(X'\) given \((X,\widetilde X)\) both have law \(r_{\widetilde X}\). The second inequality follows from the law of total variance.

For each \(z\), decomposing \(g=(g-P_jg)+P_jg\) gives
\[
 \bigl(\Var^{\mathcal F_{t_j}}g\bigr)(z)
 \le 2\mathcal F_{t_j}[(g-P_jg)^2](z)+2\bigl(\Var^{\mathcal F_{t_j}}(P_jg)\bigr)(z).
\]
By \eqref{eq:forward-kernel}, \(\mathcal F_{t_j}(z,\cdot)\) is Gaussian with covariance at most \(t_jI_n\), so the Gaussian Poincar\'e inequality (Lemma~\ref{lem:poincare-rules}, item~3) gives
\[
 \bigl(\Var^{\mathcal F_{t_j}}(P_jg)\bigr)(z)
 \le t_j\mathcal F_{t_j}\bigl(\|\nabla P_jg\|^2\bigr)(z).
\]
Averaging over \(z\sim\lat_{t_j}\), using \(\lat_{t_j}\mathcal F_{t_j}=\Pi_{t_j}\) and the gradient bound above, we obtain
\begin{align*}
 \lat_{t_j}\Var^{\mathcal F_{t_j}}g
 &\le 2\Pi_{t_j}(g-P_jg)^2+2t_j\Pi_{t_j}\|\nabla P_jg\|^2\\
 &\le 2(1+t_j/h_j)\Pi_{t_j}\Var^{P_j}g.
\end{align*}

\subsubsection{Proof of Lemma~\ref{lem:c-sqrt}}\label{sec:proof-c-sqrt}

Since both sides of \eqref{eq:sqrt-test} are unchanged when a constant is added to \(g\), we assume \(\mu_1g=0\). It then follows that \(|g|\le\osc(g)\) and \(\mu_1g^2=\Var_{\mu_1}g\). With \(p,q\) the densities of \(\mu_1,\mu_2\) with respect to a common dominating measure,
\begin{align*}
 \mu_2g-\mu_1g
 &=\int g\cdot2\sqrt{p}\bigl(\sqrt{q}-\sqrt{p}\bigr)
 +\int g\bigl(\sqrt{q}-\sqrt{p}\bigr)^2\\
 &\le2\sqrt{\Var_{\mu_1}g\cdot\Hel(\mu_1,\mu_2)}
 +\osc(g)\Hel(\mu_1,\mu_2),
 \end{align*}
where the second line follows from Cauchy--Schwarz and the definitions of squared Hellinger distance in \eqref{eq:Helliger}. Repeating the same bound for \(-g\) gives \eqref{eq:sqrt-test}.

Since the function \(\Phi_\lambda\) is increasing and \((1+2\lambda)\)-Lipschitz on \([0,1]\), and \(\mu_1\Phi_\lambda(g)=\Phi_\lambda(\mu_1g)+\lambda\Var_{\mu_1}g\), with \eqref{eq:sqrt-test} and \(\osc(g)\le1\),
\begin{align*}
 \Phi_\lambda(\mu_2g)&\le\Phi_\lambda(\mu_1g)+(1+2\lambda)|\mu_2g-\mu_1g|\\
 &\le\mu_1\Phi_\lambda(g)-\lambda\Var_{\mu_1}g
 +(2+4\lambda)\sqrt{\Var_{\mu_1}g\cdot\Hel(\mu_1,\mu_2)} +(1+2\lambda)\Hel(\mu_1,\mu_2)\\
 &\le\mu_1\Phi_\lambda(g)-\frac\lambda2\Var_{\mu_1}g
 +\frac{21}\lambda\Hel(\mu_1,\mu_2),
 \end{align*}
where the last inequality uses \(2ab\le a^2+b^2\) and \(2(1+2\lambda)^2+\lambda(1+2\lambda)\le21\) for \(0<\lambda\le1\). This proves \eqref{eq:Phi}.

Turning to \eqref{eq:initphi}, with \eqref{eq:sqrt-test} and \(2\sqrt{\Var_{\mu_1}g\cdot\Hel(\mu_1,\mu_2)}\le\lambda\Var_{\mu_1}g+\lambda^{-1}\Hel(\mu_1,\mu_2)\),
\begin{align*}
 \mu_2g&\le\mu_1g+\lambda\Var_{\mu_1}g+(1+\lambda^{-1})\Hel(\mu_1,\mu_2)\le\mu_1\Phi_\lambda(g)+\frac2\lambda\Hel(\mu_1,\mu_2),
 \end{align*}
since \(\mu_1\Phi_\lambda(g)=\mu_1g+\lambda\mu_1g^2\ge\mu_1g+\lambda\Var_{\mu_1}g\) and \(1\le1/\lambda\).

\subsubsection{Proof of Lemma~\ref{lem:c-euler}}\label{sec:proof-c-euler}

Assume \(\Pi_{t_j}\|\Delta_j\|^2<\infty\), since otherwise the bound is immediate. Fix \(x\). The comparison kernel \(P_j(x,\cdot)\) and the sampler update \(\Kraw_j(x,\cdot)\) are mixtures over \(u\sim\Gauss(x,h_j I_n)\), with respective conditional laws \(r_u\) and
\[
 \gamma_u\coloneqq\Gauss\bigl(u+h_j[\nabla\log\Pi_{t_j}(x)+\Delta_j(x)],h_j I_n\bigr).
\]
Thus items~1 and~2 of Lemma~\ref{lem:c-hellinger} give
\[
 \Hel(P_j(x,\cdot),\Kraw_j(x,\cdot))\le\int\varphi_{h_j}(u-x)\Hel(r_u,\gamma_u)\,du.
\]

To bound the distance between these conditional laws, fix \(u\) and let \(\psi\coloneqq\sqrt{dr_u/d\gamma_u}\). Expanding the Gaussian exponents shows that \(\psi(x')^2\) is proportional to \(\Pi_{t_j}(x')e^{-[\nabla\log\Pi_{t_j}(x)+\Delta_j(x)]^\top x'}\), and hence
\[
 \nabla\psi(x')=\tfrac12\psi(x')\bigl[\nabla\log\Pi_{t_j}(x')-\nabla\log\Pi_{t_j}(x)-\Delta_j(x)\bigr].
\]
The Gaussian factor in \(r_u\) gives \(r_u\|x'\|^2<\infty\), so Lipschitz continuity of \(\nabla\log\Pi_{t_j}\) implies \(\gamma_u\|\nabla\psi\|^2<\infty\). Together with \(\gamma_u\psi^2=1\), this verifies the hypotheses of the extended Poincar\'e inequality in Appendix~\ref{sec:general-preliminaries}. By Cauchy--Schwarz, \(0\le\gamma_u\psi\le1\); applying the Gaussian Poincar\'e inequality with constant \(h_j\), item~3 of Lemma~\ref{lem:poincare-rules}, therefore gives
\begin{align*}
 \Hel(r_u,\gamma_u)
 &=2(1-\gamma_u\psi)
 \le2\bigl(1-(\gamma_u\psi)^2\bigr)=2\Var_{\gamma_u}\psi\\
 &\le2h_j\,\gamma_u\|\nabla\psi\|^2
 =\frac{h_j}{2}\,r_u\|\nabla\log\Pi_{t_j}(x')-\nabla\log\Pi_{t_j}(x)-\Delta_j(x)\|^2.
\end{align*}

Let \(X,\widetilde X,X'\) be constructed as in Step~1. Combining the two Hellinger bounds and averaging over \(X\sim\Pi_{t_j}\) yields
\begin{align*}
 \int\Hel(P_j(x,\cdot),\Kraw_j(x,\cdot))\,\Pi_{t_j}(dx)
 &\le\frac{h_j}{2}\E\|\nabla\log\Pi_{t_j}(X')-\nabla\log\Pi_{t_j}(X)-\Delta_j(X)\|^2\\*
 &\le h_j\Pi_{t_j}\|\Delta_j\|^2+h_j\E\|\nabla\log\Pi_{t_j}(X')-\nabla\log\Pi_{t_j}(X)\|^2,
\end{align*}
where the last step uses \(\|a+b\|^2\le2\|a\|^2+2\|b\|^2\). Conditional on \(\widetilde X\), the two drift values are i.i.d. Subtracting \(\nabla\log\Pi_{t_j}(\widetilde X)\) and expanding their squared difference gives
\[
 \E\|\nabla\log\Pi_{t_j}(X')-\nabla\log\Pi_{t_j}(X)\|^2\le 2\E\|\nabla\log\Pi_{t_j}(X)-\nabla\log\Pi_{t_j}(\widetilde X)\|^2.
\]
It then follows by the \(L_j\)-Lipschitzness of the prior score and  \(\widetilde X-X=\sqrt{h_j}\,\xi\) for a standard Gaussian \(\xi\), that
\begin{align*}
 \E\|\nabla\log\Pi_{t_j}(X)-\nabla\log\Pi_{t_j}(\widetilde X)\|^2
 &\le 2L_j^2\E\|\widetilde X-X\|^2
   +2h_j\,\E\|A^\top A\xi\|^2/a_{t_j}^2\\
 &=2h_j\big[nL_j^2+\|A^\top A\|_{\rm F}^2/a_{t_j}^2\big].
\end{align*}
Plugging this into the earlier inequality gives \eqref{eq:euler-distance}.
\subsection{Weak variance inequality and containment}\label{sec:variance}\label{sec:mixture}

Recall the latent representation and the \emph{latent support} \(\mathsf S_t\coloneqq\supp z\) under \(\lat_t\) defined in Appendix~\ref{sec:direct}. We define the following large-support mass as follows.

\begin{definition}[Large-support mass]\label{def:cap-defect}
For a signal \(x^\star\) with cap \(q\) from \eqref{eq:proof-parameters}, its \emph{large-support mass} at a matrix \(A\), an observation \(y\in\R^m\) and a level \(t\) is
\[
 \delta_t\coloneqq\lat_t^{A,y}\{|\mathsf S_t|>q\}.
\]
\end{definition}
This definition measures how the latent variable is concentrated around sparse signals.

\begin{proposition}[Weak variance inequality and containment]\label{prop:weakvariance}\label{prop:containment}
Let \(x^\star\) be a signal with cap \(q\) at accuracy \(\eta\) as in \eqref{eq:proof-parameters}, and  \(\delta_t\) be the large-support mass. The following statements hold.
\begin{enumerate}
\item \emph{Weak variance inequality.} For every \(A\in\mathcal A(q)\), \(y\in\R^m\), \(\tau\le t\le T\), and bounded smooth \(g\) with bounded gradient, we have
\begin{equation}\label{eq:weakvariance}
 \Var_{\Pi_t}g\le V_t\,\Pi_t\norm{\nabla g}^2+\delta_t\osc(g)^2,
 \qquad V_t\coloneqq t+C\frac ns\sqrt{\min\{1,\sigma^2+tn/m\}} ,
\end{equation}
where one may take \(C=8\sqrt{2(K+2)}\), with \(K\) the fixed universal constant in \eqref{eq:matrix-spectral}.
\item \emph{Containment.} For every matrix \(A\) with \(\|A\|^2\le2n\) and observations \(y=Ax^\star+\sigma\zeta\), the large-support mass satisfies
\begin{equation}\label{eq:containment}
 \E_\zeta\delta_t\le e^{-14L_\eta}, \qquad t>0.
\end{equation}

\end{enumerate}
\end{proposition}

The rest of this subsection is devoted to proving Proposition~\ref{prop:weakvariance}. Steps~1--3 establish the weak variance inequality, and Step~4 bounds the defect on average over the observation noise. Fix a bounded smooth \(g\) with bounded gradient. At a fixed level \(t\), denote \(b=b_t\) and \(B=B_t\) for short.

\paragraph{Step 1: symmetrization.}\label{sec:b-proof}
 Let \(Z\) and \(Z'\) be independent with law \(\lat_t\), and draw \(X\) and \(X'\) independently from \(\mathcal F_t(Z,\cdot)\) and \(\mathcal F_t(Z',\cdot)\). By \eqref{eq:endpoint-identity}, \(X\) and \(X'\) are independent with law \(\Pi_t\). Set
\[
 \mathcal V\coloneqq\supp Z\cup\supp Z',\qquad F\coloneqq g(X)-g(X').
\]
Conditioning on the union \(\mathcal V\) preserves exchangeability of the two copies. Therefore
\[
 \E[F\mid\mathcal V]=0,\qquad
 \Var_{\Pi_t}g=\tfrac12\E F^2=\tfrac12\E\Var(F\mid\mathcal V).
\]
We call a union $ \mathcal V$ \emph{covered} if \(|\mathcal V|\le2q\). Otherwise, if \(|\mathcal V|>2q\), it requires at least one copy to have support larger than \(q\),
\begin{equation}\label{eq:uncovered}
 \PP(|\mathcal V|>2q)\le2\delta_t.
\end{equation}
It thus suffices to show that, for every covered union \(J\), the conditional law of \((X,X')\) given \(\mathcal V=J\) has Poincaré constant at most \(V_t\). Indeed, applying this bound to \(g(X)-g(X')\) on covered unions and using \(F^2\le\osc(g)^2\) on the complement would give
\[
 \begin{split}
 \Var_{\Pi_t}g
 &\le\tfrac12V_t\,\E\bigl[\|\nabla g(X)\|^2+\|\nabla g(X')\|^2\bigr]
 +\tfrac12\PP(|\mathcal V|>2q)\osc(g)^2\\
 &\le V_t\Pi_t\|\nabla g\|^2+\delta_t\osc(g)^2.
 \end{split}
\]
We establish the conditional Poincaré bound by identifying the pair law, controlling its interaction, and passing it through the forward kernel.

\paragraph{Step 2: identify and bound the pair laws.}\label{sec:matrix-event}\label{sec:geometry}
Fix a nonempty covered union \(J\) with \(\PP(\mathcal V=J)>0\). We express the conditional law of the coordinate pairs \((Z_i,Z'_i)_{i\in J}\) as a quadratic tilt of a product of pair laws. On \(\mathcal V=J\), both copies vanish off \(J\), and at least one coordinate in each pair \((Z_i,Z'_i)\), \(i\in J\), is nonzero. Thus conditioning on the union imposes a separate restriction \((Z_i,Z'_i)\ne(0,0)\) at each coordinate.

Recall the Gaussian mixture representation of the tilted target in Appendix~\ref{sec:direct}. From \eqref{eq:latent} and the product form of the prior, we have
\[
 \lat_t(dz)\propto e^{b^\top z-z^\top Bz/2}\nu(dz),\qquad
 \nu(dz)\coloneqq\bigotimes_{i=1}^n\bigl[\delta_0(dz_i)+\omega(2\pi)^{-1/2}dz_i\bigr],
\]
where  \(\omega=s/(n-s)\) is the prior odds.

We first introduce a family of reference pair laws. Fix \(B_0>0\) and define
\[
 \rho_u(dz)\coloneqq\delta_0(dz)+\omega(2\pi)^{-1/2}e^{uz-B_0z^2/2}\,dz,
 \qquad u\in\R.
\]
For \(\theta\in\R^2\), let \(\mu_\theta\) be the probability obtained by normalizing the restriction of \(\rho_{\theta_1}\otimes\rho_{\theta_2}\) to \(\R^2\setminus\{(0,0)\}\). This removes the joint atom at the origin while retaining the two axis components.

We first bound the covariance and Poincaré constant of each pair law uniformly over its linear tilts, as defined in \eqref{eq:linear_tilt}.  The proof is given in Appendix~\ref{sec:proof-b-pinning}.

\begin{lemma}[A Gaussian plane and its axes]\label{lem:pinning}
Let \(\log_+x\coloneqq\max(0,\log x)\).
For every \(\theta\in\R^2\), the pair law \(\mu_\theta\) has the following properties:
\begin{enumerate}
\item \(\mu_\theta^\eta=\mu_{\theta+\eta}\) for every \(\eta\in\R^2\),
\item the covariance of \(\mu_{\theta}\) is at most \(\frac{3}{B_0}\bigl(1+\log_+(\sqrt{B_0}/\omega)\bigr)I_2\),
\item \(\mu_\theta\) has Poincaré constant at most \(\frac{1}{B_0}+\frac{2}{\omega\sqrt{B_0}}\).
\end{enumerate}
In particular the bounds of items~2 and~3 hold for every linear tilt of \(\mu_{\theta}\).
\end{lemma}

To obtain the positive semidefinite interaction required by Lemma~\ref{lem:b-interaction} in Appendix~\ref{sec:general-preliminaries}, we slightly increase the diagonal reference precisions by setting
\[
 B_{*,J}\coloneqq\min_{j\in J}B_{jj},\qquad b_i^{\rm ref}\coloneqq B_{ii}+\matTol B_{*,J}\quad(i\in J),
\]
where \(\matTol\) is defined in Section~\ref{sec:ov-transfer}. The block condition \eqref{eq:block-condition} gives
\[
 \Xi\coloneqq\diag(b_i^{\rm ref})-B_{JJ}
 =\matTol B_{*,J}I-(B-\diag B)_{JJ},\qquad
 0\preceq\Xi\preceq2\matTol B_{*,J}I.
\]

For each \(i\in J\), take \(\mu^{(i)}=\mu_{(b_i,b_i)}\) with \(B_0=b_i^{\rm ref}\). Multiplying the two copies of the preceding factorization and imposing the pairwise restrictions shows that \(w\coloneqq((Z_i,Z'_i))_{i\in J}\) has conditional law
\begin{equation}\label{eq:pairs-law}
 \mu_H(dw)\propto e^{w^\top Hw/2}\mu(dw),\qquad \mu\coloneqq\bigotimes_{i\in J}\mu^{(i)},\qquad 0\preceq H\preceq2\matTol B_{*,J}I,
\end{equation}
where \(H\) acts as \(\Xi\) on each copy: \(w^\top Hw=Z_J^\top\Xi Z_J+(Z'_J)^\top\Xi Z'_J\). The quadratic tilt exactly compensates for the increased reference precisions: for \(z\in\R^J\),
\[
 e^{b_J^\top z-z^\top B_{JJ}z/2}
 =e^{z^\top\Xi z/2}\prod_{i\in J}e^{b_iz_i-b_i^{\rm ref}z_i^2/2}.
\]
The spectral bounds on \(H\) follow from those on \(\Xi\), and the normalizing constants of the individual pair laws are absorbed into the overall normalizer.

Lemma~\ref{lem:b-interaction} in Appendix~\ref{sec:general-preliminaries} transfers the Poincaré bound to the quadratic tilt \(\mu_H\) when the interaction is weak relative to the covariance bound.

\paragraph{Step 3: establish the conditional Poincaré inequality.} Recognizing $\mu_H$ is a quadratic tilt of \(\mu\), the plan is to verify the covariance and Poincaré conditions of Lemma~\ref{lem:b-interaction}, then pass the resulting bound through the forward kernel.
\begin{itemize}
\item For the covariance bound,
\(\|A\|^2\le Kn/m\le2n\), \(\Sigma_t\succeq\sigma^2I_m\), and \(\matTol\le1/16\) give
\[
 1\le B_{*,J}\le b_i^{\rm ref}
 \le(1+\matTol)(1+2n/\sigma^2)
 \le\frac{4n}{\min\{1,\sigma\}^2}.
\]
Since \(\omega^{-1}\le n/s\le n\), it follows that
\[
 \frac{\sqrt{b_i^{\rm ref}}}{\omega}
 \le\frac{2n^{3/2}}{\min\{1,\sigma\}},\qquad
 1+\log_+\frac{\sqrt{b_i^{\rm ref}}}{\omega}
 \le1+\log\frac{2n^{3/2}}{\min\{1,\sigma\}}
 \le2L_\eta,
\]
where the last inequality uses \(\eta\le1\) and \(\log2\le1\). Lemma~\ref{lem:pinning}(1)--(2) and the block diagonal covariance of a product therefore yield, for every linear tilt \(\mu^\theta\),
\[
 \Cov(\mu^\theta)\preceq v_0I,
 \qquad v_0\coloneqq\frac{6L_\eta}{B_{*,J}}.
\]
Together with \eqref{eq:pairs-law} and \(\matTol=c_{\rm mat}/L_\eta\), this gives
\[
 v_0\|H\|\le\frac{6L_\eta}{B_{*,J}}\,2\matTol B_{*,J}
 =12c_{\rm mat}<\tfrac{3}{4},
\]
where we used \(c_{\rm mat}<1/16\).
\item For the Poincaré bound, \eqref{eq:matrix-spectral} and \(m\le n\) give
\[
 \Sigma_t=\sigma^2I_m+t(2I_m+AA^\top)
 \preceq\bigl(\sigma^2+(K+2)tn/m\bigr)I_m
 \preceq(K+2)(\sigma^2+tn/m)I_m.
\]
Together with \(\|A_i\|^2\ge1/2\) on \(\mathcal A\), this implies
\[
 \frac1{b_i^{\rm ref}}\le\frac1{B_{ii}}
 =\bigl(1+A_i^\top\Sigma_t^{-1}A_i\bigr)^{-1}
 \le\min\{1,2(K+2)(\sigma^2+tn/m)\}.
\]
By Lemma~\ref{lem:pinning}(3), every tilted pair has Poincaré constant at most
\[
 \frac1{b_i^{\rm ref}}+\frac2{\omega\sqrt{b_i^{\rm ref}}}
 \le\frac{1+2\omega^{-1}}{\sqrt{b_i^{\rm ref}}}
 \le2\sqrt{2(K+2)}\,\frac ns\sqrt{\min\{1,\sigma^2+tn/m\}}
 \eqqcolon K_0.
\]
Here we used \(b_i^{\rm ref}\ge1\), \(1+2\omega^{-1}\le2n/s\), and \(\min\{1,cx\}\le c\min\{1,x\}\) for \(c\ge1\) and \(x\ge0\). Tensorization, Lemma~\ref{lem:poincare-rules}(1), gives the same bound \(K_0\) for every linear tilt of \(\mu\).
\end{itemize}
Lemma~\ref{lem:b-interaction} then bounds the Poincaré constant of \(\mu_H\) by
\[
 \frac{K_0}{1-v_0\|H\|}\le4K_0
 =8\sqrt{2(K+2)}\,\frac ns\sqrt{\min\{1,\sigma^2+tn/m\}}.
\]

By \eqref{eq:forward-kernel}, the conditional law of \((X,X')\) is an affine image of the latent pair, with linear part of norm at most \(\|M_t\|\le1\), plus an independent Gaussian with covariance \(\diag(tM_t,tM_t)\preceq tI_{2n}\). Lemma~\ref{lem:poincare-rules}(2)--(4) therefore gives a Poincaré constant at most \(t+4K_0=: V_t\). When \(J=\varnothing\), both latent vectors vanish and the Gaussian alone gives the bound \(t\). In view of Step~1, it now yields \eqref{eq:weakvariance}, with the explicit choice \(C=8\sqrt{2(K+2)}\) for \(c_{\rm mat}<1/16\).

\paragraph{Step 4: bound the large-support mass.}\label{sec:containment-proof}
Set \(\xi_t\coloneqq\sigma A^\top\Sigma_t^{-1}\zeta\), where \(y=Ax^\star+\sigma\zeta\). Centering the residual energy in \eqref{eq:latent} at \(x^\star\) gives
\[
\lat_t(dz)\propto e^{-\mathcal U_t(z)}p_\star(dz), \quad \mbox{where}\quad
 \mathcal U_t(z) := \tfrac12(z-x^\star)^\top G_t(z-x^\star)-\xi_t^\top(z-x^\star).
\]
With \(\mathcal T\coloneqq\{z:|\supp z|>q\}\), we have
\begin{equation}\label{eq:expression_delta_t}
 \delta_t=\frac{\int_{\mathcal T}e^{-\mathcal U_t}\,dp_\star}{\int e^{-\mathcal U_t}\,dp_\star}.
\end{equation}
We bound the denominator from below for every noise realization and the numerator from above after averaging over the noise. Recall \(S=\supp x^\star\), and first suppose \(S\neq\varnothing\).

For the denominator of \eqref{eq:expression_delta_t}, introduce \(\rho\) to be the law of a vector that vanishes off \(S\) and has distribution \(\Gauss(x^\star_S,rI_S)\) on \(S\), where \(r=\min\{1,\sigma^2/(2n)\}\). Since \(\rho\) is centered at \(x^\star\), the linear term in \(\mathcal U_t\) has zero mean under $\rho$. Moreover, \(G_t=A^\top\Sigma_t^{-1}A\preceq(2n/\sigma^2)I_n\), so for every \(\zeta\),
\[
 \rho\mathcal U_t=\tfrac r2\tr\bigl((G_t)_{SS}\bigr)\le\frac{|S|}{2}.
\]
Since the prior $p_\star$ assigns probability \((s/n)^{|S|}(1-s/n)^{n-|S|}\) to support \(S\), with standard Gaussian active coordinates, it follows \(\rho\ll p_\star\), and
\[
 \begin{split}
 \mathsf{KL}(\rho\|p_\star)
 &=|S|\log\frac ns+(n-|S|)\log(1+\omega)
 +\tfrac12\bigl(|S|r+\|x^\star\|^2-|S|-|S|\log r\bigr)\\
 & \le2|S|L_\eta+2s+\tfrac12\|x^\star\|^2,
 \end{split}
\]
where we use \(r\le1\), \(\log(n/s)\le L_\eta\), \(\log(1/r)=\log\max\{1,2n/\sigma^2\}\le2L_\eta\), and \((n-|S|)\log(1+\omega)\le n\omega\le2s\) in the last line. A change of measure followed by Jensen's inequality now gives
\[
 \int e^{-\mathcal U_t}\,dp_\star
 \ge e^{-\rho\mathcal U_t-\mathsf{KL}(\rho\|p_\star)}
\ge e^{-2|S|L_\eta-|S|/2-2s-\|x^\star\|^2/2}.
\]
When \(S=\varnothing\), we have \(x^\star=0\) and \(\mathcal U_t(0)=0\), so the prior atom at zero gives \(\int e^{-\mathcal U_t}\,dp_\star\ge p_\star(\{0\})=(1-s/n)^n\ge e^{-2s}\), which is the same denominator bound.

For the numerator of \eqref{eq:expression_delta_t}, note that \(\xi_t\) is a centered Gaussian vector whose covariance satisfies
\[
 \Cov_\zeta(\xi_t)=\sigma^2A^\top\Sigma_t^{-2}A
 \preceq A^\top\Sigma_t^{-1}A=G_t,
\]
since \(\Sigma_t\succeq\sigma^2I_m\). The Gaussian moment-generating formula therefore gives, for each \(z\),
\[
 \E_\zeta e^{-\mathcal U_t(z)}
 =\exp\left\{-\tfrac12(z-x^\star)^\top\bigl(G_t-\Cov_\zeta(\xi_t)\bigr)(z-x^\star)\right\}
 \le1.
\]
Under \(p_\star\), the support size is binomial with parameters \(n\) and \(s/n\). Thus Tonelli's theorem and Markov's inequality applied to \(e^{|\supp z|}\) yield
\[
 \begin{aligned}
 \E_\zeta\int_{\mathcal T}e^{-\mathcal U_t}\,dp_\star
 &=\int_{\mathcal T}\E_\zeta e^{-\mathcal U_t(z)}\,p_\star(dz)
 \le p_\star(\mathcal T)\le e^{-q}\Bigl(1+(e-1)\frac sn\Bigr)^n
 \le e^{(e-1)s-q},
 \end{aligned}
\]
where the last inequality uses \(1+u\le e^u\).

Combining the uniform denominator bound with the expected numerator bound in \eqref{eq:expression_delta_t}, we obtain
\[
 \E_\zeta\delta_t
 \le\exp\Bigl\{2|S|L_\eta+\tfrac{|S|}2+(e+1)s+\tfrac12\|x^\star\|^2-q\Bigr\}
 \le e^{-14L_\eta}.
\]
Here, the last inequality follows from \(L_\eta\ge1\), \(e+1\le4\), and \(q\ge14(1+s+|S|+\|x^\star\|^2)L_\eta\). This proves \eqref{eq:containment} for every \(t>0\) and completes the proof of Proposition~\ref{prop:weakvariance}.

\subsubsection{Proof of Lemma~\ref{lem:pinning}}\label{sec:proof-b-pinning}

\paragraph{Proof of item 1.} Completing the square gives, for \(u\in\R\),
\[
 \omega(2\pi)^{-1/2}e^{uz-B_0z^2/2}dz=k_u\Gauss(u/B_0,B_0^{-1})(dz),\qquad
 k_u\coloneqq\frac{\omega}{\sqrt{B_0}}e^{u^2/(2B_0)}.
\]
Let \(\gamma\coloneqq\Gauss(\theta/B_0,B_0^{-1}I_2)\), and let \(\gamma_i\) be the law of \(X\sim\gamma\) with coordinate \(i\) replaced by zero. Expanding \(\rho_{\theta_1}\otimes\rho_{\theta_2}\), removing the atom at the origin, and dividing by \(k_{\theta_1}k_{\theta_2}\) yields
\begin{equation}\label{eq:mixture_dis}
 \mu_\theta=\frac{1}{1+\lambda_1+\lambda_2}\bigl(\gamma+\lambda_1\gamma_1+\lambda_2\gamma_2\bigr),\qquad
 \lambda_i\coloneqq k_{\theta_i}^{-1}=\frac{\sqrt{B_0}}{\omega}e^{-\theta_i^2/(2B_0)}.
\end{equation}
Multiplication by \(e^{\eta^\top z}\) changes \(\theta\) to \(\theta+\eta\) and preserves the restriction to \(\R^2\setminus\{0\}\), proving item~1.

\paragraph{Proof of item 2.} Following the mixture distribution \eqref{eq:mixture_dis}, under \(\gamma\) the variance is \(B_0^{-1}I_2\), while under \(\gamma_i\) the variance is bounded by \(B_0^{-1}(I_2-e_ie_i^\top)+(\theta_i^2/B_0^2)e_ie_i^\top\), where \(e_i\) is the \(i\)th coordinate vector. Thus
\[
 \Cov(\mu_\theta)\preceq B_0^{-1}I_2+\sum_{i=1}^2\frac{\lambda_i}{1+\lambda_1+\lambda_2}\frac{\theta_i^2}{B_0^2}e_ie_i^\top.
\]
Using \(\lambda_i/(1+\lambda_1+\lambda_2)\le\min\{1,\lambda_i\}\), we obtain
\[
 \frac{\lambda_i}{1+\lambda_1+\lambda_2}\frac{\theta_i^2}{B_0^2}
 \le\frac{\theta_i^2}{B_0^2}\min\left\{1,\frac{\sqrt{B_0}}{\omega}e^{-\theta_i^2/(2B_0)}\right\}
 \le\frac{2}{B_0}\left(1+\log_+\frac{\sqrt{B_0}}{\omega}\right).
\]
The last inequality uses \(r\min\{1,ae^{-r}\}\le1+\log_+a\) for \(r\ge0\) and \(a>0\). Indeed, the case \(r\le\log_+a\) is immediate; otherwise write \(r=\log_+a+y\) and use \(ye^{-y}\le1\). Since the matrix sum above is diagonal,
\[
 \Cov(\mu_\theta)
 \preceq\frac{3+2\log_+(\sqrt{B_0}/\omega)}{B_0}I_2
 \preceq\frac{3}{B_0}\left(1+\log_+\frac{\sqrt{B_0}}{\omega}\right)I_2,
\]
which proves item~2.

\paragraph{Proof of item 3.} We first control the difference between a Gaussian average and the value at the origin. For \(X\sim\Gauss(b,B_0^{-1})\) and a smooth compactly supported \(g\), we claim that
\begin{equation}\label{eq:ald-trace}
 |\E g(X)-g(0)|^2\le\frac{2}{B_0}e^{B_0b^2/2}\E|g'(X)|^2.
\end{equation}
The proof of the above statement can be found at the end of the proof.
For a smooth compactly supported \(g\) on \(\R^2\), apply \eqref{eq:ald-trace} in coordinate \(i\) with \(b=\theta_i/B_0\) and average over the other coordinate. Multiplying by \(\lambda_i=(\sqrt{B_0}/\omega)e^{-\theta_i^2/(2B_0)}\) cancels the exponential factor, so Jensen's inequality gives
\[
 \lambda_i|\gamma_i g-\gamma g|^2
 \le\frac{2}{\omega\sqrt{B_0}}\,\gamma|\partial_i g|^2.
\]
Centering at the plane mean \(\gamma g\) gives
\[
 \Var_{\mu_\theta}g\le
 \frac{\Var_\gamma g+\sum_{i=1}^2\lambda_i\bigl(\Var_{\gamma_i}g+|\gamma_i g-\gamma g|^2\bigr)}{1+\lambda_1+\lambda_2}.
\]
Each Gaussian component has Poincaré constant at most \(B_0^{-1}\) by Lemma~\ref{lem:poincare-rules}(3). Combining this bound with the two coordinate estimates and \(\gamma/(1+\lambda_1+\lambda_2)\le\mu_\theta\) gives
\[
 \Var_{\mu_\theta}g
 \le\left(\frac{1}{B_0}+\frac{2}{\omega\sqrt{B_0}}\right)
 \mu_\theta\|\nabla g\|^2,
\]
proving item~3.

\begin{proof}[Proof of \eqref{eq:ald-trace}]
Let \(\varphi,\Phi\) denote the standard Gaussian density and distribution function, and set \(a=-b\sqrt{B_0}\) and \(f(z)=g(b+z/\sqrt{B_0})\). Integrating \(f'\) between \(a\) and a standard Gaussian \(G\) gives
\[
 \E f(G)-f(a)
 =\int_a^\infty f'(z)(1-\Phi(z))\,dz-\int_{-\infty}^a f'(z)\Phi(z)\,dz .
\]
By Cauchy--Schwarz, the square is at most \(\E|f'(G)|^2\) times
\(\int_a^\infty (1-\Phi)^2/\varphi+\int_{-\infty}^a\Phi^2/\varphi\).
For \(z\ge0\),
\[
 1-\Phi(z)=e^{-z^2/2}\int_0^\infty e^{-zu}\varphi(u)\,du
 \le\tfrac12e^{-z^2/2}.
\]
Consequently, the sum of the two integrals at \(a=0\) is at most \(\pi/2\). By symmetry, moving the split point from zero to \(a\) increases their sum by
\[
 \int_0^{|a|}\frac{2\Phi(z)-1}{\varphi(z)}\,dz
 \le\int_0^{|a|}2z e^{z^2/2}\,dz
 =2\bigl(e^{a^2/2}-1\bigr),
\]
where we used \(2\Phi(z)-1\le2z\varphi(0)\) for \(z\ge0\). The sum is therefore at most \(\pi/2+2(e^{a^2/2}-1)\le2e^{a^2/2}\). Since \(f(a)=g(0)\) and \(\E|f'(G)|^2=B_0^{-1}\E|g'(X)|^2\), this proves \eqref{eq:ald-trace}.
\end{proof}
\subsection{Proof of Lemma~\ref{lem:ov-transfer}}\label{sec:transfer-overview}\label{sec:finite-model}

We prove Lemma~\ref{lem:ov-transfer} by verifying the hypotheses of Proposition~\ref{prop:ov-abstract}, then applying it with averaged errors
to establish the bounds
\[
 \E_\zeta\mathcal E\le C\bigl(\eta+\Esc(A,x^\star)\bigr),
 \qquad
 \E_\zeta\mathcal E_{\sf def}\le C\eta
\]
for its error terms in \eqref{eq:ov-transfer-error}.

Throughout this proof, we denote \(\ell=\log(T/\tau)\). Since \(m\le n\), \(\eta\le1\), and \(\ell=\log(n/\tau^2)\le2\log(en/\tau)\), the iteration bound \eqref{eq:ald-transfer-count} implies
\[
 N\ge\max\left\{\ell,\sqrt{\frac mn\left(1+\frac{n}{s\sigma\max\{1,\sigma\}}\right)}\right\}.
\]
Together with the definition of \(\gamma\), this gives \(0<\gamma\le1\) and \(\ell/N\le1\), which we use in the error estimates below.

\paragraph{Step 1: verify the assumptions of Proposition~\ref{prop:ov-abstract}.}\label{sec:variance-consequence}
We start by establishing a Lipschitz bound for the prior score. Since the score of the smoothed prior acts coordinatewise, we first bound the curvature of a single coordinate.

For \(t>0\), let \(s_t\) be the scalar score of \((1-s/n)\Gauss(0,t)+(s/n)\Gauss(0,1+t)\). Differentiating the scalar mixture density gives
\[
 s_t(x)=-\frac xt+\frac{xp}{t(1+t)},\qquad
 s_t'(x)=-\frac1t+\frac{p+2v p(1-p)}{t(1+t)},
\]
where \(v=x^2/[2t(1+t)]\), \(r=(s/(n-s))\sqrt{t/(1+t)}\) and \(p=re^v/(1+re^v)\).
Since \(s\le n/2\), we have \(r\le1\). For \(v\le\log(1/r)\), \(v p(1-p)\le\log(1/r)\). For \(v=\log(1/r)+w\) with \(w\ge0\), it is at most \((\log(1/r)+w)e^{-w}\le\log(1/r)+1\). It follows that
\begin{equation}\label{eq:ald-score-curvature}
 \sup_x |s_t'(x)|\le \frac Ct\left[1+\log\left(\frac{n-s}{s}\sqrt{1+1/t}\right)\right].
\end{equation}
for some constant $C$. For \(t\ge\tau\), \eqref{eq:ald-score-curvature} gives a Lipschitz constant \(C\log(en/\tau)/t\) for \(\nabla\log p_t\). Moreover, since \(A\in\mathcal A(q)\), Proposition~\ref{prop:weakvariance}(1) gives the weak variance inequality \eqref{eq:weakvariance} for \(\Pi_{t_j}\), with constant \(V_{t_j}\) and defect \(\delta_{t_j}\). Thus Proposition~\ref{prop:ov-abstract} applies with
\[
 L_j=C\log(en/\tau)/t_j,\qquad V_j=V_{t_j},\qquad \delta_j=\delta_{t_j}.
\]

\paragraph{Step 2: bound the target movement and the latent defect.}
Let \(u_j\) and \(b_j\) denote the squared Hellinger movements of the latent law and the conditional Gaussian kernel in Proposition~\ref{prop:ov-abstract}. The following lemma bounds both movement terms and the latent Hellinger length. Its proof is given in Section~\ref{sec:ald-sparse-motion}.

\begin{lemma}[Target movement and latent Hellinger length]\label{lem:ald-latent-estimates}
Use the cap \(q\) and matrix tolerance \(\matTol\) from \eqref{eq:proof-parameters}. Suppose \(A\in\mathcal A(q)\), \(q\le m\), and \(y=Ax^\star+\sigma\zeta\). Let \(\ell=\log(T/\tau)\) and suppose \(N\ge\ell\). With \(h_j=\gamma t_j\) and \(0<\gamma\le1\), we have
\begin{align}
 \E_\zeta\sum_{j<N}\left(1+\frac{V_{t_j}}{h_j}\right)u_j
 & \le \frac{C m\ell}{N\gamma}\left(1+\frac{n^2}{ms\sigma\max\{1,\sigma\}}\right), \label{eq:ald-weighted-movement}\\
 \E_\zeta b_j & \le \frac{Cn\ell^2}{N^2},\qquad 0\le j<N, \label{eq:ald-gaussian-motion} \\
 \sum_{j<N}\sqrt{\E_\zeta u_j}& \le C\left(\frac{n^3}{\sigma^2}+nm\right). \label{eq:ald-latent-length}
\end{align}
\end{lemma}

Combining \eqref{eq:ald-weighted-movement} and \eqref{eq:ald-gaussian-motion}, and using \(m\le n\) yields
\[
 \E_\zeta\sum_{j<N}\left[\left(1+\frac{V_{t_j}}{h_j}\right)u_j+
 \left(1+\frac{1}{\gamma}\right)b_j\right]
 \le \frac{C \log^2(en/\tau)}{N\gamma}
 \left(n+\frac{n^2}{s\sigma\max\{1,\sigma\}}\right).
\]

By the containment bound \eqref{eq:containment} and Cauchy--Schwarz at each level, we have
\begin{align*}
 \E_\zeta\mathcal E_{\sf def}
 \le\sum_{j<N}\sqrt{\E_\zeta\delta_{t_j}}\sqrt{\E_\zeta u_j}
&  \le e^{-7L_\eta}\sum_{j<N}\sqrt{\E_\zeta u_j} \\
 &  \le C e^{-7L_\eta} n^3\max\{1,\sigma^{-2}\}\le C\eta ,
\end{align*}
where the second line applies the Hellinger-length bound \eqref{eq:ald-latent-length}, $m\leq n$, and  \(e^{-L_\eta}=\eta\min\{1,\sigma\}/(en)\) by \eqref{eq:proof-parameters}.

\paragraph{Step 3: combine the estimates and conclude.} For the initialization error, since \eqref{eq:cap-facts} gives \(\|x^\star\|^2\le q\le m\le n\), we have the following estimate; its proof is given in Section~\ref{sec:proof-c-init}.
\begin{lemma}[Initialization]\label{lem:c-init}
For a signal \(x^\star\) with \(\|x^\star\|^2\le n\) and \(y=Ax^\star+\sigma\zeta\),
\begin{equation}\label{eq:init-error}
 \E_\zeta\Hel(\Pi_T,\mu_{\sf init})\le n/T=\tau.
\end{equation}
\end{lemma}

For the discretization error, use the bound on \(L_j\) from Step~1 and note that the likelihood drift is affine with linear part \(-A^\top A/a_t\). Using \(h_j=\gamma t_j\), \(a_{t_j}\ge2t_j\) and \(\|A^\top A\|_{\mathrm F}^2\le m\|A\|^4\), the discretization term in \eqref{eq:ov-transfer-error} is bounded by
\begin{equation}\label{eq:ald-euler-cost}
 C N\gamma^2\left[n\log^2(en/\tau)+m\|A\|^4\right]
 \le C_K N\gamma^2\frac{n^2}{m}\log^2(en/\tau),
\end{equation}
where the last inequality uses \(m\le n\) and \(m\|A\|^4\le K^2n^2/m\) on \(\mathcal A(q)\).

Combining the error bounds, and averaging the score estimation error to \(\Esc(A,x^\star)\) by \eqref{eq:score-error-def}, we have
\[
 \E_\zeta\mathcal E\le \tau+\Esc(A,x^\star)+
 C_K\log^2(en/\tau)\left[\frac{n^2}{m}N\gamma^2+
 \frac{n}{N\gamma}\left(1+\frac{n}{s\sigma\max\{1,\sigma\}}\right)\right].
\]
Choosing \(\gamma\) using the choice of \eqref{eq:h} balances both terms in the brackets to
\[
 \left[\frac{n^4}{mN}\left(1+\frac{n}{s\sigma\max\{1,\sigma\}}\right)^2\right]^{1/3}.
\]
The iteration bound of $N$ in \eqref{eq:ald-transfer-count} makes this quantity at most \(C\eta/\log^2(en/\tau)\), so the total contribution of the two terms is at most \(C\eta\). Since \(\tau\le\eta\), this gives the claimed bound on \(\E_\zeta\mathcal E\), while Step~2 gives \(\E_\zeta\mathcal E_{\sf def}\le C\eta\).

Averaging \eqref{eq:ov-transfer} over the noise and using \(\lambda\le1\) proves \eqref{eq:ov-averaged-eta}. For each fixed \((A,y)\), applying \eqref{eq:ov-transfer} to an indicator and its complement gives
\[
 \TV(\rho_N,\Pi_\tau)\le\lambda+C\mathcal E/\lambda+C\mathcal E_{\sf def}.
\]
Choose \(\lambda=\sqrt{\mathcal E}\) when \(0<\mathcal E\le1\), take \(\lambda\downarrow0\) when \(\mathcal E=0\), and use \(\TV\le1\) when \(\mathcal E>1\). Averaging and applying Jensen's inequality gives
\[
 \E_\zeta\TV(\rho_N,\Pi_\tau)
 \le C\sqrt{\E_\zeta\mathcal E}+C\E_\zeta\mathcal E_{\sf def}
 \le C\sqrt{\eta+\Esc(A,x^\star)}+C\eta.
\]
Since \(0<\eta\le1\), the last term is absorbed into the square-root term, proving \eqref{eq:ov-averaged-tv}.

\subsubsection{Proof of Lemma~\ref{lem:ald-latent-estimates}}\label{sec:ald-sparse-motion}

 We first bound the expected residual energy, then use this bound to control the movements of the latent law and the conditional Gaussian. Finally, we bound the total latent Hellinger length.  We begin with a density identity that will be used in both latent estimates. For fixed \(y\) and every \(t>0\), set
\[
 E_t(z)=\frac12\|\Sigma_t^{-1/2}(y-Az)\|^2,\qquad
 Z_t=\int e^{-E_t}\,dp_\star,\qquad
 f_t=\sqrt{\frac{d\lat_t}{dp_\star}}=\frac{e^{-E_t/2}}{\sqrt{Z_t}},
\]
where the last equality follows from \eqref{eq:latent}. Since \(\partial_t\log Z_t=-\lat_t(\partial_tE_t)\), differentiation gives
\[
 \partial_t f_t=-\frac{f_t}{2}\bigl(\partial_t E_t-\lat_t(\partial_t E_t)\bigr),
 \qquad
 \|\partial_t f_t\|_{L^2(p_\star)}^2=\tfrac14\Var_{\lat_t}(\partial_t E_t).
\]
These differentiations hold in \(L^2(p_\star)\) on compact intervals of positive levels by Gaussian moment domination on each of the finitely many supports.

\paragraph{Step 1: bound the residual energy.}
The lower bound on the normalizing constant in Step~4 of the proof of Proposition~\ref{prop:weakvariance} gives
\[
 Z_t e^{E_t(x^\star)}
 =\int e^{-(E_t(z)-E_t(x^\star))}\,p_\star(dz)\ge e^{-Cq},
\]
including when the support of \(x^\star\) is empty. Since \(\Sigma_t\succeq\sigma^2 I_m\), we have \(E_t(x^\star)\le\|\zeta\|^2/2\). On the other hand, nonnegativity of relative entropy gives
\[
 0\le\mathsf{KL}(\lat_t\|p_\star)=-\lat_t E_t-\log Z_t.
\]
Combining these bounds, then using \(q\le m\) and \(\E\|\zeta\|^2=m\), yields, for every \(t\in[\tau,T]\),
\begin{equation}\label{eq:ald-residual-energy}
 \lat_t E_t\le-\log Z_t\le Cq+\tfrac12\|\zeta\|^2,\qquad
 \E_\zeta[-\log Z_t]\le Cm .
\end{equation}

\paragraph{Step 2: bound the weighted latent movement.}
To prove \eqref{eq:ald-weighted-movement}, we control the density derivative in logarithmic time. Use a dot to denote differentiation in \(\log t\), and set
\[
 \psi_t=-\dot E_t,\qquad
 \theta_t=\frac{(K+2)tn/m}{\sigma^2+(K+2)tn/m}.
\]
The factor \(\theta_t\) bounds the relative rate of decrease of the residual energy, as shown below. We first establish, for each fixed \(y\), the integrated variance bound
\begin{equation}\label{eq:ald-weighted-fisher}
 \int_{\log\tau}^{\log T}\frac{\Var_{\lat_t}\psi_t}{\theta_t}\,d\log t
 \le -\log Z_\tau .
\end{equation}
Since \(m\le n\), the matrix condition gives \(2+\|A\|^2\le(K+2)n/m\). In an eigenbasis of \(2I_m+AA^\top\), let \(d_i\) denote its eigenvalues and \(b_i\) the coordinates of \(y-Az\). With
\(E_{t,i}=b_i^2/[2(\sigma^2+td_i)]\) and \(\theta_{t,i}=td_i/(\sigma^2+td_i)\), the \(i\)-th contribution to \(\psi_t\) is \(\psi_{t,i}=E_{t,i}\theta_{t,i}\). Since \(d_i\le(K+2)n/m\), we have \(0\le\theta_{t,i}\le\theta_t\) and hence \(0\le\psi_t/\theta_t\le E_t\). Differentiation gives
\[
 \frac{d}{d\log t}\left(\frac{\psi_{t,i}}{\theta_t}\right)
 =(\theta_t-2\theta_{t,i})\frac{\psi_{t,i}}{\theta_t}.
\]
Because \(|\theta_t-2\theta_{t,i}|\le\theta_t\), summing over \(i\) yields
\[
 \left|\frac{d}{d\log t}\left(\frac{\psi_t}{\theta_t}\right)\right|\le\psi_t.
\]
The logarithmic derivative of the normalized latent density is \(\psi_t-\lat_t\psi_t\), and \(\dot{\log Z_t}=\lat_t\psi_t\). Differentiating \(\lat_t(\psi_t/\theta_t)\) thus yields
\[
 \frac{\Var_{\lat_t}\psi_t}{\theta_t}
 =\frac{d}{d\log t}\lat_t\left(\frac{\psi_t}{\theta_t}\right)
 -\lat_t\frac{d}{d\log t}\left(\frac{\psi_t}{\theta_t}\right)
 \le\frac{d}{d\log t}\left[\lat_t\left(\frac{\psi_t}{\theta_t}\right)+\log Z_t\right].
\]
Differentiation under the integral is justified by the same Gaussian moment domination as above. To bound the endpoint terms, use \(0\le\psi_t/\theta_t\le E_t\) and \(\lat_t E_t+\log Z_t\le0\) from Step~1. The bracket is at most zero at \(T\) and at least \(\log Z_\tau\) at \(\tau\). Thus
\[
 \int_{\log\tau}^{\log T}\frac{\Var_{\lat_t}\psi_t}{\theta_t}\,d\log t
 \le\left[\lat_t\left(\frac{\psi_t}{\theta_t}\right)+\log Z_t\right]_{t=\tau}^{t=T}
 \le-\log Z_\tau,
\]
which proves \eqref{eq:ald-weighted-fisher}.

We next pass from this integral estimate to movement along the geometric grid. The density identity above gives \(\|\dot f_t\|_{L^2(p_\star)}^2=\Var_{\lat_t}\psi_t/4\). Since each grid interval has logarithmic length \(\ell/N\), Cauchy--Schwarz yields
\[
 u_j=\|f_{t_j}-f_{t_{j+1}}\|_{L^2(p_\star)}^2
 \le\frac{\ell}{4N}\int_{\log t_{j+1}}^{\log t_j}\Var_{\lat_t}\psi_t\,d\log t .
\]
For the weighted sum, we use the monotonicity of \(\theta_t\), together with the bound on \(V_t\). Indeed, \(\theta_t/t\le(K+2)(n/m)/(\sigma^2+tn/m)\), so \eqref{eq:weakvariance} and \(\gamma\le1\) give
\[
 \theta_{t_j}\left(1+\frac{V_{t_j}}{\gamma t_j}\right)
 \le \frac{C_K}\gamma\left(1+\frac{n^2}{ms}
 \frac{\sqrt{\min\{1,\sigma^2+t_jn/m\}}}{\sigma^2+t_jn/m}\right)
 \le\frac{C_K}\gamma\left(1+\frac{n^2}{ms\sigma\max\{1,\sigma\}}\right).
\]
The last step uses that \(x\mapsto\sqrt{\min\{1,x\}}/x\) is decreasing and \(\sigma^2+t_jn/m\ge\sigma^2\). Since \(\theta_t\le\theta_{t_j}\) for \(t\in[t_{j+1},t_j]\), it follows that
\begin{align*}
 \sum_{j<N}\left(1+\frac{V_{t_j}}{h_j}\right)u_j
 &\le\frac{\ell}{4N}\sum_{j<N}\int_{\log t_{j+1}}^{\log t_j}
 \left(1+\frac{V_{t_j}}{\gamma t_j}\right)\theta_t
 \frac{\Var_{\lat_t}\psi_t}{\theta_t}\,d\log t\\
 &\le\frac{C_K\ell}{N\gamma}
 \left(1+\frac{n^2}{ms\sigma\max\{1,\sigma\}}\right)(-\log Z_\tau),
\end{align*}
where the last inequality uses \eqref{eq:ald-weighted-fisher}. Averaging over \(\zeta\) and applying \eqref{eq:ald-residual-energy} proves \eqref{eq:ald-weighted-movement}.

\paragraph{Step 3: bound the movement of the conditional Gaussian.}
To prove \eqref{eq:ald-gaussian-motion}, fix \(z\) and \(y\), and write the covariance and mean in \eqref{eq:forward-kernel} as
\[
 C_t=tM_t,\qquad \mu_t=z+tA^\top\Sigma_t^{-1}(y-Az).
\]
We bound the contributions of the covariance and mean to the Fisher information for \(\log t\). In a singular direction of \(A\) with squared singular value \(a\), set \(S_t=a_t+ta\). The corresponding eigenvalue of \(C_t\) is \(c_t=ta_t/S_t\), and
\[
 \frac{\dot c_t}{c_t}=1-\frac{ta\sigma^2}{a_t S_t}\in[0,1].
\]
In a right null direction this logarithmic derivative is one, so \(\tr[(C_t^{-1}\dot C_t)^2]\le n\). For the mean, differentiation gives \(\dot\mu_t=t\sigma^2 A^\top\Sigma_t^{-2}(y-Az)\). In the singular direction considered above, the coefficient of the squared residual divided by \(S_t\) in \(\dot\mu_t^\top C_t^{-1}\dot\mu_t\) is
\[
 \frac{ta}{a_t}\left(\frac{\sigma^2}{S_t}\right)^2
 \le \frac{ta/a_t}{(1+ta/a_t)^2}\le1.
\]
Summing over the singular directions and adding the covariance contribution therefore bounds the Fisher information by
\[
 \dot\mu_t^\top C_t^{-1}\dot\mu_t
 +\tfrac12\tr\bigl[(C_t^{-1}\dot C_t)^2\bigr]
 \le (y-Az)^\top\Sigma_t^{-1}(y-Az)+n/2 .
\]
To control movement on one grid interval, note that
\(\Sigma_t\succeq(t/t_j)\Sigma_{t_j}\succeq e^{-\ell/N}\Sigma_{t_j}\) for \(t_{j+1}\le t\le t_j\). As with the latent density, the squared \(L^2\) norm of the logarithmic-time derivative of the square-root Gaussian density is one quarter of its Fisher information. Integrating over the interval and applying Cauchy--Schwarz gives
\[
 \Hel\bigl(\mathcal F_{t_j}(z,\cdot),\mathcal F_{t_{j+1}}(z,\cdot)\bigr)
 \le\frac{\ell^2}{4N^2}\left[e^{\ell/N}(y-Az)^\top\Sigma_{t_j}^{-1}(y-Az)+n/2\right].
\]
Averaging first over \(z\sim\lat_{t_j}\), as in the definition of \(b_j\), turns the residual quadratic form into \(2\lat_{t_j}E_{t_j}\). Averaging next over \(\zeta\) and applying \eqref{eq:ald-residual-energy}, with \(m\le n\) and \(\ell/N\le1\), yields
\[
 \E_\zeta b_j
 \le\frac{\ell^2}{4N^2}\left[2e^{\ell/N}\E_\zeta\lat_{t_j}E_{t_j}+n/2\right]
 \le\frac{Cn\ell^2}{N^2},
\]
proving \eqref{eq:ald-gaussian-motion}.

\paragraph{Step 4: bound the total latent Hellinger length.}
For \eqref{eq:ald-latent-length}, we return to the density derivative in ordinary time \(t\). We will bound its norm by an integrable function on \((0,\infty)\), giving a length estimate independent of the grid endpoints. Since \(\partial_t\Sigma_t=2I_m+AA^\top\preceq(K+2)(n/m)I_m\) and \(\Sigma_t\succeq(\sigma^2+2t)I_m\),
\[
 0\le-\partial_t E_t\le \frac{(K+2)n}{m(\sigma^2+2t)}\,E_t .
\]
To bound the second moment of \(E_t\), observe that \(u^2\) is increasing and \(e^{-u}\) is decreasing for \(u\ge0\). Their covariance under the prior law of \(E_t(Z)\) is therefore nonpositive, so, for each fixed \(y\),
\[
 \lat_t E_t^2
 =\frac{\int E_t^2e^{-E_t}\,dp_\star}{\int e^{-E_t}\,dp_\star}
 \le\int E_t^2\,dp_\star.
\]
For \(Z\sim p_\star\) independent of \(\zeta\), we have \(\E\|Z\|^4\le s^2+3s\) and \(\E\|\zeta\|^4=m(m+2)\). Using \(s\le n\), \(\|x^\star\|^2\le q\le m\le n\), and \(\|A\|^2\le2n\) gives
\[
 \bigl(\E_{\zeta,Z}\|A(x^\star-Z)+\sigma\zeta\|^4\bigr)^{1/2}
 \le C(n^2+\sigma^2 m).
\]
Combining these estimates with the density identity and
\(E_t(z)\le\|y-Az\|^2/[2(\sigma^2+2t)]\) bounds the speed of the square-root density:
\[
 \begin{aligned}
 \bigl(\E_\zeta\|\partial_t f_t\|_{L^2(p_\star)}^2\bigr)^{1/2}
 &=\tfrac12\bigl(\E_\zeta\Var_{\lat_t}(\partial_tE_t)\bigr)^{1/2}\\
 &\le\tfrac12\bigl(\E_\zeta\lat_t[(\partial_tE_t)^2]\bigr)^{1/2}
 \le\frac{C(K+2)n(n^2+\sigma^2m)}{m(\sigma^2+2t)^2}.
 \end{aligned}
\]
Integrating \(\partial_t f_t\) on each grid interval and applying Minkowski's inequality in \(L^2\) over the product of \(p_\star\) and the noise law now gives
\begin{align*}
 \sum_{j<N}\sqrt{\E_\zeta u_j}
 &\le\int_\tau^T\!\bigl(\E_\zeta\|\partial_t f_t\|_{L^2(p_\star)}^2\bigr)^{1/2}\,dt\\
 &\le C(K+2)\frac nm(n^2+\sigma^2m)\int_0^\infty\frac{dt}{(\sigma^2+2t)^2}
 \le C(K+2)\left(\frac{n^3}{m\sigma^2}+n\right)\\
 &\le C\left(\frac{n^3}{\sigma^2}+nm\right).
\end{align*}
Here, the integral equals \(1/(2\sigma^2)\), and the last inequality uses \(K+2\le4m\), since \(K\le2m\) and \(m\ge1\). This proves \eqref{eq:ald-latent-length} and completes the proof.

\subsubsection{Proof of Lemma~\ref{lem:c-init}}\label{sec:proof-c-init}
The forward-kernel representation reduces the initialization error to a second moment of the latent law. Indeed, by \eqref{eq:forward-kernel},
\[
 \gamma_z\coloneqq\mathcal F_T(z,\cdot)=\Gauss\bigl(M_Tz+Tb_T,\,TM_T\bigr).
\]
The target \(\Pi_T\) is the mixture of \(\gamma_z\) over \(z\sim\lat_T\) by \eqref{eq:endpoint-identity}, whereas \(\mu_{\sf init}=\gamma_0\) by \eqref{eq:init}. These Gaussians have common covariance \(TM_T\) and mean difference \(M_Tz\). Since \(0\prec M_T\preceq I_n\), the mixture inequality and Gaussian formula in Lemma~\ref{lem:c-hellinger} give
\[
 \begin{aligned}
 \Hel(\Pi_T,\mu_{\sf init})
 &\le\int\Hel(\gamma_z,\gamma_0)\,\lat_T(dz)\\
 &\le\frac14\int(M_Tz)^\top(TM_T)^{-1}(M_Tz)\,\lat_T(dz)
 =\frac{\E_{\lat_T}[Z^\top M_TZ]}{4T}
 \le\frac{\E_{\lat_T}\|Z\|^2}{4T}.
 \end{aligned}
\]

To bound this second moment, condition on the latent support \(\mathsf S_T=W\). By \eqref{eq:latent-support-law} and \(B_{WW}=(B_T)_{WW}\succeq I_W\),
\[
 \E_{\lat_T}\bigl[\|Z\|^2\bigm|\mathsf S_T=W\bigr]
 =\tr B_{WW}^{-1}+\bigl\|B_{WW}^{-1}(b_T)_W\bigr\|^2
 \le |W|+\|b_T\|^2\le n+\|b_T\|^2.
\]
For \(W=\varnothing\), the conditional second moment is zero, so the same bound holds. Averaging over \(W\) yields
\[
 \Hel(\Pi_T,\mu_{\sf init})\le\frac{n+\|b_T\|^2}{4T}.
\]

It remains to average \(\|b_T\|^2\) over the observation noise. Substituting \(y=Ax^\star+\sigma\zeta\) gives \(b_T=G_Tx^\star+\sigma A^\top\Sigma_T^{-1}\zeta\). The identities for \(M_T\) imply \(0\preceq G_T=T^{-1}(I_n-M_T)\preceq T^{-1}I_n\), while \(\Sigma_T\succeq\sigma^2I_m\) gives \(\sigma^2A^\top\Sigma_T^{-2}A\preceq G_T\). Consequently, using \(\|x^\star\|^2\le n\) and \(T=n/\tau\ge1\),
\[
 \E_\zeta\|b_T\|^2
 =\|G_Tx^\star\|^2+\sigma^2\tr(A^\top\Sigma_T^{-2}A)
 \le\frac{\|x^\star\|^2}{T^2}+\tr G_T
 \le\frac n{T^2}+\frac nT\le2n,
\]
and hence \(\E_\zeta\Hel(\Pi_T,\mu_{\sf init})\le3n/(4T)\le n/T\), as claimed.

\section{Terminal guarantees}\label{sec:terminal-proofs}
This appendix proves the accuracy bound of Lemma~\ref{prop:ov-terminal} and the posterior-comparison bound of Lemma~\ref{lem:ov-covariance} for the terminal target.

\subsection{Proof of Lemma~\ref{prop:ov-terminal}}\label{sec:recovery-overview}
Fix \(A\in\mathcal A(q)\), with \(q\) and \(\matTol\) chosen at \(\eta=\varepsilon\) in \eqref{eq:proof-parameters}. Throughout this subsection, write \(\mathcal A=\mathcal A(q)\), \(\ell=\log(en/\varepsilon)\), and \(H=k+\ell^2\), where \(|S|\le k\). Then \(\ell\ge1\) and \(\sqrt H\le\sqrt k+\ell\le\sqrt{2H}\). We bound the mass that \(\Pi_\tau\) assigns to the failure set \(\mathcal B(x^\star)\) in \eqref{eq:ov-failure-set} by decomposing it over the latent supports.

A support can omit true coordinates and still give accurate recovery if their total energy is small. We use this fact to decompose the terminal target.
\begin{definition}[Classes of supports]\label{def:recovery-sets}
For a sufficiently large universal constant \(K_{\rm rec}\), set \(k_{\rm ex}=\lceil K_{\rm rec}H/\ell\rceil\). A support \(W\subseteq[n]\) is \emph{large} if \(|W|>q\). Among supports of size at most \(q\), call \(W\) \emph{good} if
\[
 |W\setminus S|<k_{\rm ex},\qquad \|x^\star_{S\setminus W}\|^2\le K_{\rm rec}\sigma^2 H,
\]
and \emph{bad} otherwise.
\end{definition}

For each support \(W\), let \(\Gamma_W=\lat_\tau^W\mathcal F_\tau\) be its Gaussian component in \eqref{eq:target-components}. Bounding the failure mass by one on large and bad supports gives
\begin{equation}\label{eq:terminal-support-split}
 \Pi_\tau(\mathcal B)=\sum_W\lat_\tau\{\mathsf S_\tau=W\}\,\Gamma_W(\mathcal B)
 \le\delta_\tau+\lat_\tau\{\mathsf S_\tau\text{ is bad}\}
 +\sum_{W\text{ good}}\lat_\tau\{\mathsf S_\tau=W\}\,\Gamma_W(\mathcal B).
\end{equation}
Proposition~\ref{prop:containment} controls the large-support mass \(\delta_\tau\). Lemma~\ref{lem:bad} bounds the mass of bad supports, and Lemma~\ref{lem:good} controls the failure mass within every good component.

The two cutoffs serve both the support-weight and component estimates. The choice \(k_{\rm ex}\asymp H/\ell\) makes the false-addition tail dominate the factor \(e^{CH/\ell}\) in \eqref{eq:wrong-product}, while keeping the logarithm of the number of good supports at \(O(k+k_{\rm ex}\log n)=O(H)\). The omitted-energy cutoff supplies an exponentially small weight for bad supports and bounds \(\|x^\star_{S\setminus W}\|\) by \(C\sigma\sqrt H\) on good supports.

\paragraph{Bounds at the terminal level.}
Since \(s\ge1\), \eqref{eq:param} gives \(\sigma\le1\) for \(c\le1\). The choice \(\eta=\varepsilon\) in \eqref{eq:proof-parameters} therefore gives
\begin{equation}\label{eq:Ebounds}
 L_\varepsilon=\log\frac{en}{\varepsilon\sigma}\ge\ell,\qquad
 \matTol=\frac{c_{\rm mat}}{L_\varepsilon}\le\frac{c_{\rm mat}}\ell.
\end{equation}
Also \(\tau\le\sigma^2/(2n)\le\sigma^2\) for \(c\le1/2\). On \(\mathcal A\), \(\|A\|^2\le Kn/m\le2n\) since \(K\le2m\). Hence
\[
 \Sigma_\tau=a_\tau I_m+\tau AA^\top\preceq(\sigma^2+2\tau+2n\tau)I_m\preceq4\sigma^2I_m.
\]
Consequently, \(G_\tau\succeq A^\top A/(4\sigma^2)\), and \(\|A_i\|^2\ge1/2\) gives
\begin{equation}\label{eq:endpoint-geometry}
 \tau\le\sigma^2,\qquad B_{\tau,ii}=1+G_{\tau,ii}\ge1+\frac{\|A_i\|^2}{4\sigma^2}\ge1+\frac1{8\sigma^2}\qquad(i\in[n]).
\end{equation}

\begin{lemma}[Bad supports]\label{lem:bad}
Suppose \eqref{eq:param} and \eqref{eq:signal-admission} hold. For the bad supports of Definition~\ref{def:recovery-sets}, every \(A\in\mathcal A\) satisfies
\[
 \E_\zeta\lat_\tau\{\mathsf S_\tau\text{ is bad}\}\le C\varepsilon.
\]
\end{lemma}

The proof of this lemma can be found in Appendix~\ref{sec:proof-bad}.

\begin{lemma}[Good supports]\label{lem:good}
Suppose \eqref{eq:param}, \eqref{eq:signal-admission}, and \(A\in\mathcal A\) hold. There is a universal constant \(C_{\sf rad}\) such that, with \(\mathcal B(x^\star)\) defined by \eqref{eq:ov-failure-set} using this constant, there is an event of the observation noise of probability at least \(1-\varepsilon\) on which every good support \(W\) of Definition~\ref{def:recovery-sets} satisfies \(\Gamma_W(\mathcal B(x^\star))\le\varepsilon\).
\end{lemma}
The proof of this lemma can be found in Appendix~\ref{sec:proof-good}.

Let \(\mathcal E\) be the noise event of Lemma~\ref{lem:good}, with \(\PP_\zeta(\mathcal E^c)\le\varepsilon\). Averaging \eqref{eq:terminal-support-split} over the noise gives
\begin{align}\label{eq:terminal-assembly}
 \E_\zeta\Pi_\tau(\mathcal B)
 &\le\E_\zeta\delta_\tau+\E_\zeta\lat_\tau\{\mathsf S_\tau\text{ is bad}\}
 +\varepsilon\,\PP_\zeta(\mathcal E)+\PP_\zeta(\mathcal E^c) \nonumber \\
 &\le\E_\zeta\delta_\tau+\E_\zeta\lat_\tau\{\mathsf S_\tau\text{ is bad}\}+2\varepsilon\le C\varepsilon.
\end{align}
The last inequality follows from \(\E_\zeta\delta_\tau\le e^{-14L_\varepsilon}\le\varepsilon\) by Proposition~\ref{prop:containment} and from the bound on bad supports in Lemma~\ref{lem:bad}.

\subsubsection{Proof of Lemma~\ref{lem:bad}}\label{sec:proof-bad}\label{sec:retained}
We compare the weight of each bad support with that of the true support \(S\). We then sum the resulting fractional moments over supports with too many false additions or excessive omitted signal energy.

\paragraph{Step 1: reduce the posterior mass to fractional moments.}
For any \(0<\iota\le1\), the latent mass of \(W\) is at most \(\min\{1,w_W/w_S\}\le(w_W/w_S)^\iota\). Hence
\[
 \lat_\tau\{\mathsf S_\tau\text{ is bad}\}
 \le\sum_{W\text{ bad}}\min\Bigl\{1,\frac{w_W}{w_S}\Bigr\}
 \le\sum_{W\text{ bad}}\Bigl(\frac{w_W}{w_S}\Bigr)^\iota.
\]
The true support \(S\) is good, so every bad support differs from \(S\) and has size at most \(q\). The next lemma uses \(\iota\) close to one to keep \(n^{1-\iota}\) bounded in the sum over false additions. It also retains an exponential penalty for the total energy of the omitted coordinates.

\begin{lemma}[A fractional moment of one weight ratio]\label{lem:a-small}
There are universal constants \(c\in(0,1/2]\) and \(c_0,C>0\) with the following property. Suppose \(0<\sigma\le1\) and \(0<\tau\le c\min\{\sigma^2/n,\varepsilon\}\). Let \(x^\star\) have support \(S\), and choose \(q\) and \(\matTol\) at \(\eta=\varepsilon\) in \eqref{eq:proof-parameters}. For \(A\in\mathcal A(q)\) and \(y=Ax^\star+\sigma\zeta\), let \(w_W\) be the weights in \eqref{eq:weights} at level \(\tau\). Write \(B=B_\tau\), \(\omega=s/(n-s)\), and set
\[
 \iota\coloneqq1-\frac1{2\ell},\qquad
 \iota_1\coloneqq\frac{\iota(1+\matTol)}{1-\matTol},
\]
so that \(1/2\le\iota<1\) and \(1-\iota_1\ge1/(4\ell)\). Define
\[
 \begin{aligned}
 \chi_i&\coloneqq\frac{\omega^\iota((1-\matTol)B_{ii})^{-\iota/2}}{\sqrt{1-\iota_1}}
 &&(i\notin S),\\
 \psi_i&\coloneqq
 \frac{\omega^{-\iota} B_{ii}^{\iota/2}}{\sqrt{1-\iota_1}}\,
 e^{-c_0 B_{ii}(x_i^\star)^2}
 &&(i\in S).
 \end{aligned}
\]
For each support \(W\ne S\) with \(|W|\le q\), let \(F=W\setminus S\) and \(S_0=S\setminus W\). Then
\begin{equation}\label{eq:terminal-weight-product}
 \E_\zeta\Bigl(\frac{w_W}{w_S}\Bigr)^\iota
 \le \exp\!\left(C\ell\matTol^2\sigma^2\|x^\star\|^2
        -c_0\sum_{i\in S_0} B_{ii}(x_i^\star)^2\right)
       \prod_{i\in F}\chi_i\prod_{i\in S_0}\psi_i .
\end{equation}
\end{lemma}
The factors \(\chi_i\) and \(\psi_i\) account for adding and omitting individual coordinates. The separate exponential in \eqref{eq:terminal-weight-product} controls the total energy of the omitted coordinates. Appendix~\ref{sec:proof-a-small} proves the estimate.

\paragraph{Step 2: bound the coordinate factors.}
Write \(\chi_{\rm tot}=\sum_{i\notin S}\chi_i\) and \(\psi_{\rm tot}=\sum_{i\in S}\psi_i\). Using the lower bound on \(B_{ii}\) in \eqref{eq:endpoint-geometry}, the bound \(1-\iota_1\ge1/(4\ell)\) from Lemma~\ref{lem:a-small}, and \(n\omega\le2s\), we obtain
\[
 \chi_{\rm tot}\le 2n^{1-\iota}(Cs\sigma)^\iota\sqrt\ell.
\]
Here \(n^{1-\iota}\le e^{1/2}\). By \eqref{eq:param}, \(s\sigma\sqrt\ell\le c\), and \(\ell^{(1-\iota)/2}=\ell^{1/(4\ell)}\le2\). Since \(\iota\ge1/2\), taking \(c\) sufficiently small yields \(\chi_{\rm tot}\le1\).

The minimum-amplitude condition in \eqref{eq:signal-admission} controls \(\psi_{\rm tot}\). For every \(i\), the matrix event and \(\Sigma_\tau\succeq\sigma^2 I_m\) give
\[
 \frac{1}{8\sigma^2}\le B_{ii}
 =1+A_i^\top\Sigma_\tau^{-1}A_i
 \le 1+\frac{2n}{\sigma^2}.
\]
Let \(C_{\rm det}\) denote its universal constant. For \(i\in S\), we have \(B_{ii}(x_i^\star)^2\ge C_{\rm det}^2 L_\varepsilon/8\). Using \(\omega^{-\iota}\le\omega^{-1}\le n\), \(B_{ii}^{\iota/2}\le\sqrt{B_{ii}}\), and \(|S|\le n\), we obtain
\[
 \psi_{\rm tot}\le 2 n^2\sqrt{\ell}\sqrt{1+2n/\sigma^2}\,
            \exp(-c_0 C_{\rm det}^2 L_\varepsilon/8)\le1.
\]
The last inequality holds for a sufficiently large universal \(C_{\rm det}\), because \(\log n\), \(\log(1/\sigma)\), and \(\log\ell\) are all bounded by \(L_\varepsilon\).

\paragraph{Step 3: sum over bad supports.}
Let \(F=W\setminus S\) and \(S_0=S\setminus W\). The signal-norm condition in \eqref{eq:signal-admission}, together with \eqref{eq:Ebounds}, gives
\[
 \ell\matTol^2\sigma^2\|x^\star\|^2
 \le C\ell\frac{(\sqrt k+\ell)^2}{\ell^2}
 \le\frac{CH}{\ell}.
\]
Thus the common factor in \eqref{eq:terminal-weight-product} is at most \(e^{CH/\ell}\). Apply that estimate to the bad supports, all of which have size at most \(q\), and then extend the resulting nonnegative product sums to unrestricted subsets \(F\subseteq S^c\) and \(S_0\subseteq S\). Each \(W\) is determined by \((F,S_0)\), and the unrestricted sum is bounded by
\(\prod_{i\notin S}(1+\chi_i)\prod_{i\in S}(1+\psi_i)\le e^{\chi_{\rm tot}+\psi_{\rm tot}}\le e^2\).
For supports with too many false additions, \(\1\{|F|\ge k_{\rm ex}\}\le e^{|F|-k_{\rm ex}}\) gives
\[
 \sum_{\substack{F\subseteq S^c\\|F|\ge k_{\rm ex}}}\prod_{i\in F}\chi_i
 \le e^{-k_{\rm ex}}\prod_{i\notin S}(1+e\chi_i)
 \le e^{-k_{\rm ex}+e\chi_{\rm tot}}.
\]
For supports with \(\|x^\star_{S_0}\|^2>K_{\rm rec}\sigma^2H\), \eqref{eq:endpoint-geometry} gives \(\sum_{i\in S_0}B_{ii}(x_i^\star)^2>K_{\rm rec}H/8\). Every bad support satisfies at least one of these two conditions. Adding their contributions yields
\begin{equation}\label{eq:wrong-product}
 \E_\zeta\lat_\tau\{\mathsf S_\tau\text{ is bad}\}
 \le C e^{CH/\ell}\left(e^{-k_{\rm ex}}+e^{-c_0 K_{\rm rec}H/8}\right).
\end{equation}
Since \(k_{\rm ex}\ge K_{\rm rec}H/\ell\) and \(H\ge H/\ell\ge\ell\), choosing the universal \(K_{\rm rec}\) sufficiently large makes the right-hand side at most \(C e^{-\ell}\le C\varepsilon\).

\subsubsection{Proof of Lemma~\ref{lem:good}}\label{sec:proof-good}\label{sec:correct}
We first bound each good component's failure mass after averaging over the observation noise. Since the class of good supports is fixed by \(x^\star\), summing these bounds and applying Markov's inequality will give one noise event on which every good component has small failure mass.

\paragraph{Step 1: identify the component law and covariance.}
Write \(G=G_\tau\), \(M=M_\tau\), \(B=B_\tau\) and \({b}=b_\tau\). For any support \(W\), \eqref{eq:latent-support-law} gives \(z_W\sim\Gauss(B_{WW}^{-1}b_W,B_{WW}^{-1})\) under \(\lat_\tau^W\), with \(z_{W^c}=0\). By \eqref{eq:forward-kernel}, a draw from \(\Gamma_W\) is \(M_{[n]W}z_W+\tau b\) plus an independent \(\Gauss(0,\tau M)\) vector. Thus \(\Gamma_W=\Gauss(L_Wy,C_W)\), where
\begin{equation}\label{eq:correct-law}
 L_W\coloneqq M_{[n]W}B_{WW}^{-1}A_W^\top\Sigma_\tau^{-1}+\tau A^\top\Sigma_\tau^{-1},\qquad
 C_W\coloneqq M_{[n]W}B_{WW}^{-1}M_{W[n]}+\tau M.
\end{equation}
For \(W=\varnothing\), this gives \(L_Wy=\tau b\) and \(C_W=\tau M\).

For every nonempty good support \(W\), we have \(|W|\le q\), so the block condition \eqref{eq:block-condition} gives
\[
 B_{WW}\succeq(1-\matTol)\min_{i\in W}B_{ii}\,I_W.
\]
Together with \(0\prec M\preceq I_n\), \(\matTol\le1/2\), \eqref{eq:endpoint-geometry}, and \(\tau\le\sigma^2/n\), this yields
\begin{equation}\label{eq:good-scales}
 \|B_{WW}^{-1}\|\le C\sigma^2,\qquad
 \tr{C}_W\le\tr B_{WW}^{-1}+\tau n\le C\sigma^2(|W|+1),\qquad
 \|{C}_W\|\le C\sigma^2.
\end{equation}
These bounds also hold for \(W=\varnothing\), when \({C}_W=\tau M\).

\paragraph{Step 2: bound the bias and observation noise.}
Let \(S_0=S\setminus W\) and \(\xi_\tau=\sigma A^\top\Sigma_\tau^{-1}\zeta\), so \(b=Gx^\star+\xi_\tau\). By \(G=B-I\) and \(W\cap S_0=\varnothing\),
\[
 (Gx^\star)_W=(B_{WW}-I)x^\star_W+B_{W S_0}x^\star_{S_0}.
\]
Substitution in \eqref{eq:correct-law} and the identity \(M+\tau G=I\) give
\[
 \begin{aligned}
 L_W y-x^\star
 ={}&-M_{[n]W}B_{WW}^{-1}x^\star_W
   +\bigl(M_{[n]W}B_{WW}^{-1}B_{W S_0}-M_{[n]S_0}\bigr)x^\star_{S_0}\\
 &+V_W+\tau\xi_\tau,\qquad V_W=M_{[n]W}B_{WW}^{-1}(\xi_\tau)_W.
 \end{aligned}
\]
For nonempty \(W,S_0\), apply \eqref{eq:block-condition} on \(U=W\cup S\), where \(|U|\le 2q\). It gives \(\|B_{W S_0}\|\le\matTol\min_{i\in U}B_{ii}\), while \(\lambda_{\min}(B_{WW})\ge(1-\matTol)\min_{i\in U}B_{ii}\). Thus \(\|B_{WW}^{-1}B_{W S_0}\|\le\matTol/(1-\matTol)\). With empty blocks contributing zero, \eqref{eq:good-scales} and \(\|M\|\le1\) bound the deterministic terms by
\[
 \|L_WAx^\star-x^\star\|
 \le C\sigma^2\|x^\star\|+C\|x^\star_{S_0}\|\le C\sigma\sqrt H.
\]
Here we used the signal-norm bound in \eqref{eq:signal-admission} and the omitted-energy bound for good supports.

Since \(\Sigma_\tau\succeq a_\tau I_m\succeq\sigma^2I_m\), we have \(\Cov\xi_\tau=\sigma^2A^\top\Sigma_\tau^{-2}A\preceq G\). Together with \(G_{WW}\preceq B_{WW}\), this gives \(\Cov V_W\preceq M_{[n]W}B_{WW}^{-1}M_{W[n]}\), whose trace and norm are at most \(C\sigma^2|W|\) and \(C\sigma^2\). Also \(G=\tau^{-1}(I_n-M)\preceq\tau^{-1}I_n\), so \(\Cov(\tau\xi_\tau)\preceq\tau^2G\preceq\tau I_n\), with trace at most \(\tau n\le\sigma^2\) and norm at most \(\tau\le\sigma^2\). The observation-noise contribution is \(\sigma L_W\zeta=V_W+\tau\xi_\tau\). Using
\[
 \sigma^2L_WL_W^\top
 =\Cov(V_W+\tau\xi_\tau)
 \preceq2\Cov V_W+2\Cov(\tau\xi_\tau),
\]
we conclude that
\[
 \tr(\sigma^2L_WL_W^\top)\le C\sigma^2(|W|+1),
 \qquad \|\sigma^2L_WL_W^\top\|\le C\sigma^2.
\]

\paragraph{Step 3: bound the failure mass simultaneously over good supports.}
For each good support \(W\), let \(g\sim\Gauss(0,I_n)\) be independent of \(\zeta\) and set
\[
 X_W=L_W(Ax^\star+\sigma\zeta)+C_W^{1/2}g.
\]
Conditional on \(\zeta\), this vector has law \(\Gamma_W\). After averaging over \(\zeta\), it is Gaussian with mean \(L_WAx^\star\) and covariance \(C_W+\sigma^2L_WL_W^\top\). By \eqref{eq:good-scales} and Step~2, this covariance has trace at most \(C\sigma^2(|W|+1)\) and norm at most \(C\sigma^2\).

We use the Gaussian concentration inequality \citep[Theorem~5.6]{boucheron2013concentration} for the \(\|\Sigma\|^{1/2}\)-Lipschitz function \(g\mapsto\|\Sigma^{1/2}g\|\). Together with \(\E\|\Sigma^{1/2}g\|\le\sqrt{\tr\Sigma}\), it gives, for a centered Gaussian vector \(V\) with covariance \(\Sigma\) and \(0<\delta<1\),
\begin{equation}\label{eq:gauss-conc}
 \PP\Bigl(\|V\|>\sqrt{\tr\Sigma}+\sqrt{2\|\Sigma\|\log(1/\delta)}\Bigr)\le\delta.
\end{equation}

A good \(W\) chooses a subset of \(S\) and fewer than \(k_{\rm ex}\) coordinates outside \(S\). Hence the number of good supports is at most
\[
 J=2^{|S|}\sum_{j=0}^{k_{\rm ex}-1}\binom{n-|S|}{j}\le 2^{k+1}n^{k_{\rm ex}-1},
 \qquad \log(J/\varepsilon^2)\le C H.
\]
The last bound follows from \(k_{\rm ex}-1<K_{\rm rec}H/\ell\), \(\log n\le\ell\), and \(\log(1/\varepsilon)\le\ell\le H\). Also \(|W|\le k+k_{\rm ex}-1\le C H\). Apply \eqref{eq:gauss-conc} to \(X_W-L_WAx^\star\) with failure probability \(\delta=\varepsilon^2/J\), and combine it with the bias bound in Step~2. Since \(\sqrt H\le\sqrt k+\ell\), choosing the universal constant \(C_{\sf rad}\) in \eqref{eq:ov-failure-set} sufficiently large after fixing \(K_{\rm rec}\) gives
\[
 \E_\zeta\Gamma_W(\mathcal B)
 =\PP(X_W\in\mathcal B)\le\frac{\varepsilon^2}{J}.
\]
Finally, Markov's inequality and the bound on the number of good supports give
\[
 \PP_\zeta\!\left(\max_{W\text{ good}}\Gamma_W(\mathcal B)>\varepsilon\right)
 \le\frac1\varepsilon\sum_{W\text{ good}}\E_\zeta\Gamma_W(\mathcal B)
 \le\varepsilon.
\]
The complementary noise event proves the lemma.

\subsubsection{Proof of Lemma~\ref{lem:a-small}}\label{sec:proof-a-small}
We derive the product estimate \eqref{eq:terminal-weight-product} by writing the exact weight ratio through Schur complements, integrating the Gaussian noise, and bounding the deterministic shifts.
First, the terminal bounds \eqref{eq:Ebounds} give \(\matTol\le c_{\rm mat}/\ell\), and \(c_{\rm mat}\le1/16\). Since \(\matTol\le1/2\),
\[
 \iota_1=\iota\Bigl(1+\frac{2\matTol}{1-\matTol}\Bigr)
 \le\Bigl(1-\frac1{2\ell}\Bigr)\Bigl(1+\frac1{4\ell}\Bigr)
 \le1-\frac1{4\ell}.
\]
Thus \(1-\iota_1\ge1/(4\ell)\), as claimed. Fix \(W\ne S\) with \(|W|\le q\), and write
\[
 \bar W\coloneqq S\cup W,\qquad F\coloneqq W\setminus S,\qquad
 S_0\coloneqq S\setminus W.
\]
Empty determinants and products are understood to equal one, and quadratic forms on empty coordinate sets equal zero.

\paragraph{Step 1: express the weight ratio through Schur complements.}
Set \(G=G_\tau\), \({b}=b_\tau\), \({D}\coloneqq(\diag B)_{\bar W\bar W}\) and \(\check\beta\coloneqq B_{\bar W\bar W}^{-1}{b}_{\bar W}\). Since \(W\ne S\) and \(|\bar W|\le q+|S|\le2q\) by \eqref{eq:cap-facts}, the block condition \eqref{eq:block-condition} gives \(\|B_{\bar W\bar W}-{D}\|\le\matTol\min_{i\in\bar W}B_{ii}\). As \({D}\succeq(\min_{i\in\bar W}B_{ii})I\), this implies \(-\matTol{D}\preceq B_{\bar W\bar W}-{D}\preceq\matTol{D}\), hence \((1-\matTol){D}\preceq B_{\bar W\bar W}\preceq(1+\matTol){D}\).

For \(J\in\{S,W\}\) and \(K=\bar W\setminus J\), set \(\Upsilon_K=B_{KK}-B_{KJ}B_{JJ}^{-1}B_{JK}\). Completing the square in \(v_J\) gives
\[
 v_K^\top\Upsilon_Kv_K=\min_{v_J\in\R^J}v^\top B_{\bar W\bar W}v
 \qquad\text{for each fixed }v_K\in\R^K.
\]
The comparison with \(D\), together with the choice \(v_J=0\) for the upper bound, gives \((1-\matTol)D_{KK}\preceq\Upsilon_K\preceq B_{KK}\preceq(1+\matTol)D_{KK}\). Hadamard's inequality and monotonicity of the determinant therefore yield
\begin{equation}\label{eq:schur-bounds}
 (1-\matTol){D}_{KK}\preceq\Upsilon_K\preceq(1+\matTol){D}_{KK},\qquad
 \det\Upsilon_{S_0}\le\prod_{i\in S_0}B_{ii},\qquad
 \det\Upsilon_F\ge\prod_{i\in F}(1-\matTol)B_{ii}.
\end{equation}

The block determinant identity and completion of the square also give, for the same \((J,K)\),
\[
 \begin{aligned}
 \det B_{\bar W\bar W}&=\det B_{JJ}\det\Upsilon_K,\\
 b_J^\top B_{JJ}^{-1}b_J
 &=\check\beta^\top B_{\bar W\bar W}\check\beta-\check\beta_K^\top\Upsilon_K\check\beta_K.
 \end{aligned}
\]
For the second identity, maximize \(2b_J^\top v-v^\top B_{JJ}v\) over vectors supported on \(J\), and use the variational formula above for the remaining quadratic form on \(K\). Substituting these identities into \eqref{eq:weights} gives
\begin{equation}\label{eq:ratio}
 \frac{w_W}{w_S}=\omega^{|F|-|S_0|}\sqrt{\frac{\det\Upsilon_{S_0}}{\det\Upsilon_F}}\,
 \exp\Bigl\{\tfrac12\check\beta_F^\top\Upsilon_F\check\beta_F
            -\tfrac12\check\beta_{S_0}^\top\Upsilon_{S_0}\check\beta_{S_0}\Bigr\}.
\end{equation}

\paragraph{Step 2: average over the Gaussian noise.}
Set \(a=B_{\bar W\bar W}^{-1}x^\star_{\bar W}\). Substituting \(y=Ax^\star+\sigma\zeta\) into \({b}=A^\top\Sigma_\tau^{-1}y\) gives \({b}=Gx^\star+\xi_\tau\), where \(\xi_\tau\coloneqq\sigma A^\top\Sigma_\tau^{-1}\zeta\) is centered Gaussian. Since \(\Sigma_\tau\succeq a_\tau I_m\succeq\sigma^2I_m\), its covariance satisfies \(\Cov\xi_\tau=\sigma^2A^\top\Sigma_\tau^{-2}A\preceq G\). As \(\supp x^\star\subseteq\bar W\) and \(G_{\bar W\bar W}=B_{\bar W\bar W}-I\preceq B_{\bar W\bar W}\),
\[
 \check\beta=x^\star_{\bar W}-{a}+B_{\bar W\bar W}^{-1}(\xi_\tau)_{\bar W},\qquad
 \Cov\check\beta=B_{\bar W\bar W}^{-1}\Cov\bigl((\xi_\tau)_{\bar W}\bigr)B_{\bar W\bar W}^{-1}
 \preceq B_{\bar W\bar W}^{-1}\preceq(1-\matTol)^{-1}{D}^{-1}.
\]

To bound the exponent, set \({r}\coloneqq{D}^{1/2}\check\beta\), \(\bar{{r}}\coloneqq\E_\zeta {r}={D}^{1/2}(x^\star_{\bar W}-{a})\) and \({z}\coloneqq{r}-\bar{{r}}\), so that \({z}\) is centered Gaussian with \(\Cov{z}\preceq(1-\matTol)^{-1}I\) by the last step. By \eqref{eq:schur-bounds}, the exponent in \eqref{eq:ratio} is at most \(\tfrac12(1+\matTol)\|{r}_F\|^2-\tfrac12(1-\matTol)\|{r}_{S_0}\|^2\), and \(\|{r}_{S_0}\|^2\ge\tfrac12\|\bar{{r}}_{S_0}\|^2-\|{z}_{S_0}\|^2\). Raising \eqref{eq:ratio} to the power \(\iota\) therefore gives
\begin{equation}\label{eq:ratio-iota}
 \Bigl(\frac{w_W}{w_S}\Bigr)^\iota\le\omega^{\iota(|F|-|S_0|)}\Bigl(\frac{\det\Upsilon_{S_0}}{\det\Upsilon_F}\Bigr)^{\iota/2}
 e^{f({z})}\,e^{-\iota(1-\matTol)\|\bar{{r}}_{S_0}\|^2/4},
\end{equation}
with the convex function
\[
 f({z})\coloneqq\tfrac\iota2\bigl((1+\matTol)\|\bar{{r}}_F+{z}_F\|^2+(1-\matTol)\|{z}_{S_0}\|^2\bigr).
\]

We now enlarge the covariance in this convex upper bound. Let \({w}\sim\Gauss(0,(1-\matTol)^{-1}I-\Cov{z})\) be independent of \({z}\). Jensen's inequality for \(e^f\), conditionally on \({z}\), gives \(\E e^{f({z})}\le\E e^{f({z}+{w})}\), and the coordinates of \({z}+{w}\) are independent with variance \({v}\coloneqq(1-\matTol)^{-1}\). For real \(\mu\) and \(\gamma\), a scalar \({g}_0\sim\Gauss(0,{v})\) with \(\gamma{v}<1\) satisfies
\[
 \E e^{\gamma(\mu+{g}_0)^2/2}=(1-\gamma{v})^{-1/2}e^{\gamma\mu^2/(2(1-\gamma{v}))}.
\]
On \(F\) we take \(\gamma=\iota(1+\matTol)\) and \(\mu=\bar{{r}}_i\), so that \(\gamma{v}=\iota_1\). On \(S_0\) we take \(\gamma=\iota(1-\matTol)\) and \(\mu=0\), so that \(\gamma{v}=\iota\le\iota_1\). Since \(1-\iota_1\ge1/(4\ell)\), the Gaussian integral gives
\[
 \E_\zeta e^{f(z)}
 \le (1-\iota_1)^{-(|F|+|S_0|)/2}
     \exp\!\left(C\ell\|\bar r_F\|^2\right).
\]
Taking expectations in \eqref{eq:ratio-iota} and applying the determinant bounds in \eqref{eq:schur-bounds}, we obtain
\begin{equation}\label{eq:weight-ratio-shifts}
 \E_\zeta\Bigl(\frac{w_W}{w_S}\Bigr)^\iota
 \le \exp\!\left(C\ell\|\bar r_F\|^2
          -\frac{\iota(1-\matTol)}4\|\bar r_{S_0}\|^2\right)
     \prod_{i\in F}\chi_i
     \prod_{i\in S_0}\frac{\omega^{-\iota}B_{ii}^{\iota/2}}{\sqrt{1-\iota_1}} .
\end{equation}

\paragraph{Step 3: retain the penalty for omitted signal energy.}
Let \(R=D^{-1/2}B_{\bar W\bar W}D^{-1/2}\). Step~1 gives \(\|R-I\|\le\matTol\) and hence \(\|R^{-1}-I\|\le\matTol/(1-\matTol)\). Since \(D^{1/2}a=R^{-1}D^{-1/2}x^\star_{\bar W}\), the deterministic mean has the decomposition
\[
 \bar r=(I-D^{-1})D^{1/2}x^\star_{\bar W}+e,
 \qquad e=(I-R^{-1})D^{-1/2}x^\star_{\bar W},
 \qquad \|e\|\le C\matTol\sigma\|x^\star\| .
\]
The last inequality follows from \(B_{ii}\ge 1/(8\sigma^2)\) in \eqref{eq:endpoint-geometry}. On \(F\), the signal vanishes, so \(\|\bar r_F\|\le\|e\|\). On \(S_0\), use \(B_{ii}=1+G_{ii}\ge1+1/(8\sigma^2)\) and \(\sigma\le1\) to get \(1-B_{ii}^{-1}\ge1/9\). Thus \(\|u+v\|^2\ge\|u\|^2/2-\|v\|^2\) gives
\[
 \|\bar r_{S_0}\|^2
 \ge \frac{1}{162}\sum_{i\in S_0}B_{ii}(x_i^\star)^2
       -C\matTol^2\sigma^2\|x^\star\|^2 .
\]
Substitute both bounds into \eqref{eq:weight-ratio-shifts}. Since \(\ell\ge1\) and \(\iota(1-\matTol)/4\ge1/16\), the exponent there is at most
\[
 C\ell\matTol^2\sigma^2\|x^\star\|^2
 -2c_0\sum_{i\in S_0}B_{ii}(x_i^\star)^2
\]
for a universal \(c_0>0\). Each factor \(\psi_i\) includes \(\exp(-c_0B_{ii}(x_i^\star)^2)\). The other half of the exponent gives the omitted-energy penalty in \eqref{eq:terminal-weight-product}.

\subsection{Proof of Lemma~\ref{lem:ov-covariance}}\label{sec:covariance-proof}
\paragraph{The modified observation covariance.}
For a matrix \(A\), an observation \(y\in\R^m\) and any positive definite \(\Sigma\in\R^{m\times m}\), define
\begin{equation}\label{eq:ov-nu-def}
 \nu_\Sigma^{A,y}(dz)\propto e^{-(y-Az)^\top\Sigma^{-1}(y-Az)/2}\,p_\star(dz).
\end{equation}
This is the conditional law of \(Z\sim p_\star\) given \(AZ+w=y\), where \(w\sim\Gauss(0,\Sigma)\) is independent of \(Z\).

Fix \(A,y\) and recall the auxiliary model of Section~\ref{sec:latent}. Given \(X_\tau\), the observation \(\widetilde Y_\tau\) is independent of \(Z\). Hence the endpoint kernel \(\mathcal D_\tau(x,\cdot)=\mathcal L(Z\mid X_\tau=x)\) is also the law of \(Z\) given \(X_\tau=x\) and \(\widetilde Y_\tau=y\). Integrating against \(\Pi_\tau=\mathcal L(X_\tau\mid\widetilde Y_\tau=y)\) gives \(\Pi_\tau\mathcal D_\tau=\lat_\tau\). By \eqref{eq:latent} and the true observation model, the two posteriors are
\begin{equation}\label{eq:ov-latent-nu}
 \Pi_\tau\mathcal D_\tau=\nu^{A,y}_{\sigma^2I_m+\Xi},\qquad
 p_\star^{A,y}=\nu^{A,y}_{\sigma^2I_m},
\end{equation}
where \(\Xi\coloneqq\tau(2I_m+AA^\top)\succeq0\).

\paragraph{Comparison and averaging.}
We first bound the effect of the covariance perturbation at fixed \(A,y\).

\begin{lemma}[Comparison at fixed data]\label{lem:covariance}\label{lem:e-endpoint}
Suppose \(p_\star=\mathsf{BG}(s,n)\). For every \(A\in\R^{m\times n}\), \(y\in\R^m\), and symmetric \(\Xi\in\R^{m\times m}\) with \(\Xi\succeq0\), we have
\begin{equation}\label{eq:covariance-comparison}
 \TV\bigl(\nu^{A,y}_{\sigma^2I_m+\Xi},\, p_\star^{A,y}\bigr)
 \le\frac{\norm{\Xi}}{2\sigma^2}\Bigl(m+\frac{\norm y^2}{\sigma^2}\Bigr).
\end{equation}
\end{lemma}

Lemma~\ref{lem:covariance} is proved in Appendix~\ref{sec:proof-covariance}.
To deduce Lemma~\ref{lem:ov-covariance}, we average over the true observation law.
Fix \(A\) satisfying \(\|A\|^2\le2n\). The covariance perturbation in \eqref{eq:ov-latent-nu} satisfies \(\|\Xi\|\le\tau(2+2n)\le4n\tau\). Since \(\Xs\) is independent of \(A\) and has independent, centered coordinates with second moment \(s/n\),
\[
 \E\bigl[\|y\|^2\mid A\bigr]
 =\frac sn\|A\|_F^2+\sigma^2m\le2sm+\sigma^2m,
\]
where we used \(\|A\|_F^2\le m\|A\|^2\le2nm\). Applying \eqref{eq:covariance-comparison} and averaging over the observation at this fixed \(A\) gives
\[
 \E\bigl[\TV(\Pi_\tau\mathcal D_\tau,p_\star^{A,y})\mid A\bigr]
 \le\frac{2n\tau}{\sigma^2}\Bigl(m+\frac{\E[\|y\|^2\mid A]}{\sigma^2}\Bigr)\\
 \le\frac{4nm\tau}{\sigma^2}\Bigl(1+\frac s{\sigma^2}\Bigr).
\]
Multiplying by \(\1_{\{\|A\|^2\le2n\}}\), taking expectation over \(A\), and using \(m\le n\) proves Lemma~\ref{lem:ov-covariance}.

\subsubsection{Proof of Lemma~\ref{lem:covariance}}\label{sec:proof-covariance}

\paragraph{Step 1: compare the likelihoods.}
Set \(\Sigma=\sigma^2I_m+\Xi\), \(P=p_\star^{A,y}\), and \(Q=\nu_\Sigma^{A,y}\). In \eqref{eq:ov-nu-def}, the unnormalized density of \(P\) differs from that of \(Q\) by a factor \(e^{-\Delta(z)}\), where, writing \(r=y-Az\),
\[
 \Delta(z)\coloneqq\tfrac12 r^\top(\sigma^{-2}I_m-\Sigma^{-1})r
 =\frac{1}{2\sigma^2}(\Sigma^{-1/2}r)^\top\Xi(\Sigma^{-1/2}r)\ge0.
\]
Normalizing gives \(\alpha P(dz)=e^{-\Delta(z)}Q(dz)\), where \(\alpha=\E_Q e^{-\Delta(Z)}\). Since \(e^{-\Delta(z)}\le1\), both \(P\) and \(Q\) dominate \(\alpha P\), a measure of mass \(\alpha\). Thus
\[
 \TV(P,Q)\le 1-\alpha
 =\E_Q\bigl[1-e^{-\Delta(Z)}\bigr]
 \le\E_Q\Delta(Z)
 \le\frac{\|\Xi\|}{2\sigma^2}\,\E_Q\|\Sigma^{-1/2}(y-AZ)\|^2.
\]

\paragraph{Step 2: bound the residual uniformly over supports.}
Condition under \(Q\) on \(\supp Z=W\), and set \(F=\Sigma^{-1/2}A_W\), \(\widetilde y=\Sigma^{-1/2}y\), and \(H=(I_m+FF^\top)^{-1}\). On this support the active coordinates have a standard Gaussian prior. Completing the square shows that \(Z_W\) has conditional mean \((I+F^\top F)^{-1}F^\top\widetilde y\) and covariance \((I+F^\top F)^{-1}\). Since \(F(I+F^\top F)^{-1}F^\top=I_m-H\), the residual \(\widetilde y-FZ_W\) has mean \(H\widetilde y\) and covariance \(I_m-H\). Hence
\[
 \E_Q\!\left[\|\Sigma^{-1/2}(y-AZ)\|^2\mid\supp Z=W\right]
 =\|H\widetilde y\|^2+\tr(I_m-H)
 \le \frac{\|y\|^2}{\sigma^2}+m,
\]
because \(0\preceq H\preceq I_m\) and \(\|\widetilde y\|\le\|y\|/\sigma\). This also covers \(W=\varnothing\), for which \(H=I_m\) and the residual equals \(\widetilde y\). Averaging over \(W\) under \(Q\) and applying Step~1 proves \eqref{eq:covariance-comparison}.

\section{Verification of the matrix condition}\label{sec:matrix-proof}

This appendix verifies the matrix condition of Definition~\ref{def:truth} for the three sensing families. The following proposition supplies the probability bounds used in Sections~\ref{sec:ov-recovery} and~\ref{sec:ov-sampling}.

\begin{proposition}[The matrix event]\label{prop:ov-sensing}
There is a universal constant \(C>0\) such that the following holds. Fix \(0<\varepsilon<1\) and \(1\le s\le n/2\), and suppose that the measurement matrix comes from one of the families in Definition~\ref{def:families}. In the matrix event \(\mathcal A(q)\) of Definition~\ref{def:truth}, take \(K=1\) for the two tight-frame families and a sufficiently large universal constant \(K\) for independent sub-Gaussian entries.
\begin{enumerate}
\item Let \(\xs\) be a fixed $k$-sparse signal with support \(S\). Suppose \(0<\sigma\le1\), and take \(\eta=\varepsilon\) in \eqref{eq:proof-parameters}. If \(m\) satisfies \eqref{eq:sample-size}, then \(q\le m\) and \(\PP(A\notin\mathcal A(q))\le\varepsilon\).
\item Let \(\Xs\sim\mathsf{BG}(s,n)\) be independent of the measurement matrix, and take \(\eta=\varepsilon^2\) in \eqref{eq:proof-parameters}. For every \(\sigma>0\), if \(m\) satisfies \eqref{eq:sampling-sample-size}, then \(\PP\bigl(q>m\text{ or }A\notin\mathcal A(q)\bigr)\le2\varepsilon\), where \(q\) is the cap of \(\Xs\).
\end{enumerate}
\end{proposition}

In Theorems~\ref{thm:application} and~\ref{thm:sampling}, the constants \(c\) are taken small enough and the constants \(C\) large enough for this proposition to apply. We first verify the matrix event under the following sufficient measurement conditions, expressed in terms of the cap \(q\), the tolerance \(\matTol\), and a failure probability \(0<\varepsilon_0<1\):
\begin{align}
 m&\ge C\kappa^8\matTol^{-2}
 \{2q\log(en)+\log(en/\varepsilon_0)\}
 &&\text{(independent sub-Gaussian entries)},\label{eq:direct-iid-cost}\\*
 m&\ge C\kappa_b^2\matTol^{-2}q\{\log^4(en)+\log(e/\varepsilon_0)\}
 &&\text{(sampled bases)},\label{eq:d-basis-cost}\\*
 m&\ge C\kappa_f^2\matTol^{-2}q^2
 &&\text{(deterministic frames)}.\label{eq:d-frame-cost}
\end{align}
Here, the value of the constant $C$ can change from line to line.

We begin with sampled bases and frames, for which tightness gives \(B_t=I+\beta_tA^\top A\), so restricted isometry and coherence bounds suffice. For independent entries, we prove one uniform resolvent estimate and use a push-through identity to obtain the block condition. Finally, we substitute the prescribed \(q\) and \(\matTol\) and pass from fixed to random signals to prove the two parts of Proposition~\ref{prop:ov-sensing}.

The matrix quantities \(B_t\), \(q\) and \(\matTol\) are defined in Section~\ref{sec:ov-transfer}, and \(\matTol\le1/2\) by \eqref{eq:cap-facts}. For a fixed signal, the event \(\mathcal A(q)\) depends only on the measurement matrix. Since \(\|\diag R\|\le\|R\|\), every square matrix \(R\) satisfies
\begin{equation}\label{eq:offdiag}
 \|R-\diag R\|\le2\|R\|.
\end{equation}

\subsection{Sampled orthonormal bases and frames}

\paragraph{Reduction to restricted isometry.}
Both families satisfy \(AA^\top=(n/m)I_m\). We show that the matrix event holds whenever
\begin{equation}\label{eq:d-tight-rip}
 \|(A^\top A-I)_{JJ}\|\le\matTol/3
 \qquad\text{for every }J\subseteq[n]\text{ with }1\le|J|\le2q.
\end{equation}

Indeed, tightness gives \(\|A\|^2=n/m\), so \(K=1\) satisfies the spectral condition, and the singletons give \(\|A_i\|^2\ge1-\matTol/3\ge2/3\). Since \(\Sigma_t=(a_t+tn/m)I_m\),
\[
 B_t=I+\beta_tA^\top A,\qquad \beta_t\coloneqq\{\sigma^2+t(2+n/m)\}^{-1}.
\]
The off-diagonal part of \((B_t)_{JJ}\) is \(\beta_t\) times that of \((A^\top A-I)_{JJ}\), so by \eqref{eq:offdiag},
\[
 \|(B_t-\diag B_t)_{JJ}\|\le\frac{2\matTol\beta_t}3\le\matTol B_{t,ii}\qquad(i\in J),
\]
since \(B_{t,ii}=1+\beta_t\|A_i\|^2\ge2\beta_t/3\).

\paragraph{Sampled orthonormal bases.}
Let \(F\) consist of the rows of \(U\) retained independently with probability \(m/n\), each scaled by \(\sqrt{n/m}\), and let \(M\) be the number retained.
By \citet[Theorem~4.1, Eq.~(4.3)]{dirksen2015tail}, the order-\(r\) restricted isometry constant \(\delta_r(F)\) satisfies \(\PP(\delta_r(F)\ge\delta)\le\eta\) for \(0<\delta,\eta<1\) whenever
\[
 m\ge r\kappa_b^2\delta^{-2}
 \max\bigl\{d_1\log^2(r)\log m\log n,\ d_2\log(1/\eta)\bigr\},
\]
where \(d_1,d_2>0\) are universal. Take \(r=2q\), \(\delta=\matTol/3\), and \(\eta=\varepsilon_0/(n+1)\), so \(r\delta^{-2}=18q\matTol^{-2}\). Since \(2q\le m\le n\) under \eqref{eq:d-basis-cost} for \(C\) large enough,
\[
 \log^2(2q)\log m\log n\le\log^4(en),\qquad
 \log\frac{n+1}{\varepsilon_0}\le\log(en)+\log(1/\varepsilon_0).
\]
Thus \eqref{eq:d-basis-cost} implies Dirksen's bound for \(C\ge C_{\rm b}\), where \(C_{\rm b}\) is universal.
Conditional on \(M=m\), the selected rows form a uniform \(m\)-subset, so \(F\) has the law of \(A\).
Since \(m\) is a mode of \(M\sim\operatorname{Bin}(n,m/n)\), we have \(\PP(M=m)\ge1/(n+1)\).
Thus \(A\) violates \eqref{eq:d-tight-rip} with probability at most \(\frac{\varepsilon_0}{n+1}/\PP(M=m)\le\varepsilon_0\), and the reduction above gives \(\PP(A\notin\mathcal A(q))\le\varepsilon_0\).

\paragraph{Deterministic frames.}
The symmetric matrix \((A^\top A-I)_{JJ}\) has zero diagonal and off-diagonal entries of magnitude at most \(\kappa_f/\sqrt m\). By Gershgorin's bound and \eqref{eq:d-frame-cost} with $C$ large enough,
\[
 \|(A^\top A-I)_{JJ}\|\le\frac{2q\kappa_f}{\sqrt m}\le\frac\matTol3\qquad(|J|\le2q).
\]
As \(AA^\top=(n/m)I_m\), the reduction above gives \(A\in\mathcal A(q)\).

\subsection{Independent sub-Gaussian entries}

The block condition~\eqref{eq:block-condition} requires the off-diagonal part of each block \((B_t)_{JJ}\) to be small relative to its smallest diagonal entry, uniformly over \(J\) and \(t\). By~\eqref{eq:Sigma_Bt}, this block involves \(A_J^\top(I_m+vAA^\top)^{-1}A_J\) with \(v=t/a_t\). The resolvent depends on the same columns \(A_J\) that appear on either side of it. To decouple the dependency, we form the resolvent with \(A_{J^c}A_{J^c}^\top\), which is independent of \(A_J\).
For \(J\subseteq[n]\) with \(1\le|J|\le2q\) and \(0\le v\le1/2\), set
\[
 \begin{gathered}
 \Theta_J\coloneqq A_{J^c}A_{J^c}^\top=\sum_{i\notin J}A_iA_i^\top,\qquad
 \gamma_J\coloneqq\frac{n-|J|}m,\\
 M_J(v)\coloneqq(I_m+v\Theta_J)^{-1},\qquad
 \iota_J(v)\coloneqq m^{-1}\tr M_J(v),\qquad
 W_J(v)\coloneqq A_J^\top M_J(v)A_J.
 \end{gathered}
\]

Conditional on \(A_{J^c}\), the outside resolvent \(M_J(v)\) is fixed and the weighted Gram matrix \(W_J(v)\) has mean \(\iota_J(v)I_J\). Lemma~\ref{lem:d-ridge} bounds the relative deviation of \(W_J(v)\) from this scalar matrix uniformly over \(J\) and \(v\).

\begin{lemma}[Uniform resolvent concentration]\label{lem:d-ridge}
For independent sub-Gaussian entries, suppose \eqref{eq:direct-iid-cost} holds with a sufficiently large universal constant \(C\). Then, with probability at least \(1-\varepsilon_0\), \(\|A\|^2\le K n/m\) and
\begin{equation}\label{eq:d-ridge}
 \|W_J(v)-\iota_J(v)I_J\|\le\frac\matTol4\,\iota_J(v)
 \qquad\text{for every }J\subseteq[n],\ 1\le|J|\le2q,\ \text{and }0\le v\le1/2.
\end{equation}
Here, \(K\ge1\) is universal and the measurement condition ensures \(K\le2m\).
\end{lemma}

The proof is deferred to Section~\ref{sec:proof-d-ridge}.
Lemma~\ref{lem:d-ridge} gives the condition in Definition~\ref{def:truth}. At \(v=0\), \(W_J(0)=A_J^\top A_J\) and \(\iota_J(0)=1\), so \eqref{eq:d-ridge} on singletons gives \(\min_i\|A_i\|^2\ge1-\matTol/4\ge1/2\).

Fix \(\tau\le t\le T\), so that \(0<v\le1/2\) and \(\Sigma_t=a_t(I_m+vAA^\top)\). Fix \(J\) with \(1\le|J|\le2q\) and write \(M\), \(W\) and \(\iota\) for \(M_J(v)\), \(W_J(v)\) and \(\iota_J(v)\). Since \(I_m+vAA^\top=M^{-1}+vA_JA_J^\top\),
\[
 (I_m+vAA^\top)\,MA_J=A_J(I_J+vW),
\]
so \((I_m+vAA^\top)^{-1}A_J=MA_J(I_J+vW)^{-1}\) and
\[
 (B_t)_{JJ}=I_J+a_t^{-1}A_J^\top(I_m+vAA^\top)^{-1}A_J
 =I_J+a_t^{-1}f(W),\qquad f(w)\coloneqq\frac w{1+vw}.
\]

Each eigenvector of \(W\) with eigenvalue \(w\ge0\) is therefore an eigenvector of \((B_t)_{JJ}\) with eigenvalue \(1+a_t^{-1}f(w)\). Set \(b\coloneqq1+a_t^{-1}f(\iota)\). By \eqref{eq:d-ridge}, the relative deviation of this eigenvalue satisfies
\[
 \frac{|1+a_t^{-1}f(w)-b|}{b}
 =\frac{a_t^{-1}f(\iota)}{b}\,\frac{|w-\iota|}{\iota(1+vw)}
 \le\frac{|w-\iota|}{\iota}\le\frac{\matTol}{4}.
\]
Hence \(\|(B_t)_{JJ}-bI_J\|\le\matTol b/4\). This gives \(B_{t,ii}\ge(1-\matTol/4)b\) for \(i\in J\), while \eqref{eq:offdiag} bounds the off-diagonal norm by \(\matTol b/2\). We obtain the block condition~\eqref{eq:block-condition} since
\[
 \frac{\|(B_t-\diag B_t)_{JJ}\|}{\min_{i\in J}B_{t,ii}}
 \le\frac{\matTol/2}{1-\matTol/4}\le\matTol .
\]
Thus under \eqref{eq:direct-iid-cost}, Lemma~\ref{lem:d-ridge} gives \(\PP(A\notin\mathcal A(q))\le\varepsilon_0\).

\subsubsection{Proof of Lemma~\ref{lem:d-ridge}}\label{sec:proof-d-ridge}

\begin{proof}
Fix \(J\subseteq[n]\) with \(1\le |J|\le2q\).

\paragraph{Step 1: bound the outside resolvent.}
Let \(\lambda_{\min}\) and \(\lambda_{\max}\) be the extreme eigenvalues of \(\Theta_J\).
For \(J^c\ne\varnothing\), the rows of \(\sqrt m\,A_{J^c}^{\top}\) are independent, centered, and isotropic. Their independent coordinates have \(\psi_2\) norm at most \(\kappa\), so each row has sub-Gaussian norm at most \(C\kappa\) by \citet[Lemma~3.4.2]{vershynin2018high}.
Applying \citet[Theorem~4.6.1, Eq.~(4.21)]{vershynin2018high} with tail parameter \(\sqrt{3m}\) gives, with probability at least \(1-2e^{-3m}\),
\[
 \sqrt{\gamma_J}-C\kappa^2\le\sqrt{\lambda_{\min}}
 \le\sqrt{\lambda_{\max}}\le\sqrt{\gamma_J}+C\kappa^2 .
\]
Substituting the leftmost bound into the rightmost bound, we have \(\sqrt{\lambda_{\max}}\le\sqrt{\lambda_{\min}}+2C\kappa^2\), which gives \(\lambda_{\max}\le2\lambda_{\min}+C'\kappa^4\).
Since \(\iota_J(v)\ge(1+v\lambda_{\max})^{-1}\), there is a universal \(C_0\ge1\) such that, with \(D\coloneqq C_0\kappa^4\),
\begin{equation}\label{eq:direct-outside}
 \frac{\|M_J(v)\|}{\iota_J(v)}
 \le\frac{1+v\lambda_{\max}}{1+v\lambda_{\min}}\le D
 \qquad(0\le v\le1/2).
\end{equation}
If \(J^c=\varnothing\), then \(M_J(v)=I_m\) and \eqref{eq:direct-outside} holds deterministically.

\paragraph{Step 2: prove the resolvent estimate uniformly over \(J\) and \(v\).}
Conditioning on \(A_{J^c}\), let \(\lambda_1<\cdots<\lambda_l\) be the distinct eigenvalues of \(\Theta_J\), and let \(\bar E_j\) project onto the sum of its eigenspaces with eigenvalues at most \(\lambda_j\). These \(l\le m\) projections are fixed under the conditioning.

Fix one such projection \(P\) of rank \(r\) and a unit vector \(x\in\R^J\).
The entries of \(Y=\sqrt m\,A_Jx\) are independent, centered, and have variance one and \(\psi_2\) norm at most \(C\kappa\). Thus \(x^\top A_J^\top P A_Jx=Y^\top(P/m)Y\) has conditional mean \(r/m\). Since \(\|P/m\|=1/m\) and \(\|P/m\|_F^2=r/m^2\), the Hanson--Wright inequality \citep[Theorem~1.1]{rudelson2013hanson} gives
\[
 \PP\!\left(\left.|x^\top A_J^\top P A_Jx-r/m|>z\,\right|\, A_{J^c}\right)
 \le 2\exp\!\left[-c\min\!\left\{\frac{m^2z^2}{\kappa^4r},\frac{mz}{\kappa^2}\right\}\right].
\]
Take \(z=\matTol(r+m/D)/(16m)\). Since \((r+m/D)^2/r\ge4m/D\), \(\matTol\le1\), and \(\kappa\ge1\), the failure probability is at most \(2\exp\{-c m\matTol^2/(\kappa^4D)\}\).
A \(1/4\)-net of the unit sphere in \(\R^J\) has at most \(9^{|J|}\) points, and twice the largest absolute quadratic form on the net bounds the operator norm of a symmetric matrix \citep[Corollary~4.2.13 and Exercise~4.4.3(b)]{vershynin2018high}. Hence the net gives
\begin{equation}\label{eq:direct-projections}
 \left\|A_J^\top P A_J-\frac rm I_J\right\|
 \le\frac{\matTol}{8}\left(\frac rm+\frac1D\right).
\end{equation}
For each \(J\), take a conditional union bound over the net and the \(l\le m\) projections, then average over \(A_{J^c}\). Since \(\sum_{j=1}^{\min(2q,n)}\binom nj9^j\le(9en)^{2q}\), a union bound over supports, including the failures of \eqref{eq:direct-outside}, gives total failure probability at most
\[
 2(9en)^{2q}\left(e^{-3m}+m e^{-c m\matTol^2/(\kappa^4 D)}\right)
 \le\varepsilon_0/2.
\]
The last inequality follows from \eqref{eq:direct-iid-cost} with \(C\) sufficiently large, since \(D=C_0\kappa^4\) and \(m\le n\).

On this event, fix \(J\) and \(v\in[0,1/2]\). Set \(f_j=(1+v\lambda_j)^{-1}\) for \(1\le j\le l\) and \(f_{l+1}=0\).
The weights \(\varpi_j=f_j-f_{j+1}\) are nonnegative, and summation by parts gives
\[
 M_J(v)=\sum_{j=1}^l\varpi_j\bar E_j,\qquad
 \sum_{j=1}^l\varpi_j=\|M_J(v)\|,\qquad
 \sum_{j=1}^l\varpi_j\,\frac{\tr\bar E_j}{m}=\iota_J(v).
\]
Multiplying \eqref{eq:direct-projections} by \(\varpi_j\) and summing yields
\[
 \|W_J(v)-\iota_J(v)I_J\|
 \le\frac{\matTol}{8}\left(\iota_J(v)+\frac{\|M_J(v)\|}{D}\right)
 \le\frac{\matTol}{4}\,\iota_J(v)
\]
by \eqref{eq:direct-outside}. This proves \eqref{eq:d-ridge} for all \(J\) and \(v\) on the same event.

\paragraph{Step 3: bound the spectral norm of the full matrix.}
Write \(X=\sqrt m A\), set \(b=\kappa\sqrt{6\log(2n)}\), and let \(R_{ij}=X_{ij}\1\{|X_{ij}|>b\}\). The centered matrix \(H=X-R-\E(X-R)\) has independent entries bounded by \(2b\), each of variance at most \(\E X_{ij}^2=1\). Its symmetric dilation
\(\left(\begin{smallmatrix}0&H\\ H^\top&0\end{smallmatrix}\right)\)
has norm \(\|H\|\), dimension \(m+n\le2n\), and row sums of entry variances at most \(n\). Applying \citet[Corollary~3.12 and Remark~3.13]{bandeira2016sharp} to this dilation gives, with probability at least \(1-\varepsilon_0/4\),
\[
 \|H\|\le C\bigl(\sqrt n+b\sqrt{\log(8n/\varepsilon_0)}\bigr).
\]
For the discarded entries, the bound \(\E e^{X_{ij}^2/\kappa^2}\le2\) gives
\[
 \E e^{R_{ij}^2/(2\kappa^2)}
 \le 1+e^{-b^2/(2\kappa^2)}\E e^{X_{ij}^2/\kappa^2}
 \le 1+2(2n)^{-3}.
\]
By independence and \(mn\le n^2\), \(\E e^{\|R\|_{\rm F}^2/(2\kappa^2)}\le(1+2(2n)^{-3})^{mn}\le e\). Jensen's inequality gives \(\|\E R\|_{\rm F}\le\sqrt{2}\kappa\), and Markov's inequality gives \(\|R\|_{\rm F}\le\kappa\sqrt{2\log(4e/\varepsilon_0)}\) with probability at least \(1-\varepsilon_0/4\). Since \(\E X=0\), combining these bounds in \(X=H+R-\E R\) yields
\[
 \sqrt m\,\|A\|\le C\left(\sqrt n+
 \kappa\sqrt{\log(2n)\log(8n/\varepsilon_0)}\right).
\]
By \eqref{eq:proof-parameters}, \(\matTol^{-2}\ge\log(2n)\). Hence \eqref{eq:direct-iid-cost} and \(\kappa\ge1\) give
\(\kappa^2\log(2n)\log(8n/\varepsilon_0)\le Cm\le Cn\).
Thus \(\|A\|^2\le K n/m\) for a universal \(K\ge1\). Taking \(C\ge K/2\) in \eqref{eq:direct-iid-cost} ensures \(K\le2m\). The failures in Steps~2 and~3 total at most \(\varepsilon_0\).
\end{proof}

\subsection{Proof of Proposition~\ref{prop:ov-sensing}}

With \(C\) sufficiently large in \eqref{eq:direct-iid-cost}--\eqref{eq:d-frame-cost}, each condition gives \(q\le m\), and the preceding arguments give \(\PP(A\notin\mathcal A(q))\le\varepsilon_0\) for the corresponding family. It remains to derive these conditions from the sample bounds in Proposition~\ref{prop:ov-sensing}.

\paragraph{Step 1: fixed signals.}
Fix \(x^\star\), set \(k_0=1+s+|S|+\|x^\star\|^2\), and take \(\varepsilon_0=\varepsilon\) in the three measurement conditions above. For either \(\eta=\varepsilon\) or \(\eta=\varepsilon^2\), the logarithm \(L_\eta\) in \eqref{eq:proof-parameters} satisfies \(L_\eta\ge\log(en/\varepsilon)\). Hence
\[
 q\le 15k_0 L_\eta,\qquad
 \matTol^{-2}=c_{\rm mat}^{-2}L_\eta^2,\qquad
 \log^4(en)+\log(e/\varepsilon)\le2L_\eta\log^3(en).
\]
Substitution in \eqref{eq:direct-iid-cost}--\eqref{eq:d-frame-cost} bounds their right-hand sides, respectively, by
\[
 C\kappa^8 k_0 L_\eta^3\log(en),\qquad
 C\kappa_b^2 k_0 L_\eta^4\log^3(en),\qquad
 C\kappa_f^2 k_0^2 L_\eta^4 .
\]
For \(\eta=\varepsilon\) and \(0<\sigma\le1\), we have \(L_\eta=\log(en/(\varepsilon\sigma))\). Thus \eqref{eq:sample-size} with \(k=|S|\) gives the required measurement bound for each family, proving item~1.

\paragraph{Step 2: random signals.}\label{sec:matrix-event-prob}
Take \(\eta=\varepsilon^2\). For every \(\sigma>0\), \eqref{eq:proof-parameters} gives
\[
 L_{\varepsilon^2}=L_\varepsilon+\log(1/\varepsilon)\le2L_\varepsilon,
 \qquad L_\varepsilon=\log\frac{en}{\varepsilon\min\{1,\sigma\}}.
\]
Set \(k_\varepsilon=4(s+\log(1/\varepsilon))\). For each fixed signal with \(|S|+\|\xs\|^2\le k_\varepsilon\), we have \(k_0\le 6(s+\log(1/\varepsilon))\). With these two bounds, \eqref{eq:sampling-sample-size} implies the corresponding measurement condition in Step~1, so \(q\le m\) and \(\PP(A\notin\mathcal A(q))\le\varepsilon\) for each such signal.

The coordinates of \(\Xs\) are independent, and \(|\supp\Xs|+\|\Xs\|^2=Y\coloneqq\sum_i\1\{X_i^\star\ne0\}(1+(X_i^\star)^2)\). For \(g\sim\Gauss(0,1)\), \(e^{1/4}\E e^{g^2/4}=e^{1/4}\sqrt2\le2\), so
\[
 \E e^{Y/4}=\Bigl(1-\frac sn+\frac sn\,e^{1/4}\E e^{g^2/4}\Bigr)^n\le\Bigl(1+\frac sn\Bigr)^n\le e^s.
\]
Markov's inequality applied to \(e^{Y/4}\) gives \(\PP(Y>k_\varepsilon)\le e^{s-k_\varepsilon/4}=\varepsilon\).

On \(Y\le k_\varepsilon\), the cap of \(\Xs\) satisfies \(q\le m\). By independence of \(A\) and \(\Xs\), the fixed-signal bound then gives
\[
 \PP\bigl(q>m\text{ or }A\notin\mathcal A(q)\bigr)
 \le\PP(Y>k_\varepsilon)+\E\bigl[\1\{Y\le k_\varepsilon\}\,\PP(A\notin\mathcal A(q)\mid\Xs)\bigr]\le2\varepsilon .
\]

\end{document}